\documentclass[twoside]{article}

\usepackage[accepted]{aistats2024}

\usepackage{appendix}
\usepackage{amsthm}
\usepackage{amsmath}

\usepackage{amsfonts}
\usepackage{multirow}
\usepackage{multicol}

\usepackage{color}
\usepackage{algorithm}
\usepackage{algorithmic}

\usepackage{xcolor,colortbl}
\definecolor{LightCyan}{rgb}{0.88,1,1}

\usepackage{graphicx}
\usepackage{subfigure}
\usepackage{tcolorbox}

\newtheorem{theorem}{Theorem}
\newtheorem{lemma}{Lemma}

\newtheorem{assumption}{Assumption}
\newtheorem{remark}{Remark}

\begin{document}

%

%

\twocolumn[

\aistatstitle{Adaptive Federated Minimax Optimization with Lower Complexities}

\aistatsauthor{ Feihu Huang$^{1,2,*}$ \And Xinrui Wang$^{1,2}$ \And  Junyi Li$^3$ \And  Songcan Chen$^{1,2}$}

\aistatsaddress{ 1. College of Computer Science and Technology, Nanjing University of Aeronautics and Astronautics, China; \\ 2. MIIT Key Laboratory of Pattern Analysis and Machine Intelligence, China; *huangfeihu2018@gmail.com;
\\  3. Department of Electrical and Computer Engineering, University of Pittsburgh, Pittsburgh, USA. } ]

\begin{abstract}
  Federated learning is a popular distributed and privacy-preserving learning paradigm in machine learning. Recently, some federated learning algorithms have been proposed to solve the distributed minimax problems. However, these federated minimax algorithms still suffer from high gradient or communication complexity. Meanwhile, few algorithm focuses on using adaptive learning rate to accelerate these algorithms. To fill this gap, in the paper, we study a class of nonconvex minimax optimization, and propose an efficient adaptive federated minimax optimization algorithm (i.e.,  AdaFGDA) to solve these distributed minimax problems. Specifically, our AdaFGDA builds on the momentum-based variance reduced and local-SGD techniques, and it can flexibly incorporate various adaptive learning rates by using the unified adaptive matrices. Theoretically, we provide a solid convergence analysis framework for our AdaFGDA algorithm under non-i.i.d. setting. Moreover, we prove our AdaFGDA algorithm obtains a lower gradient (i.e., stochastic first-order oracle, SFO) complexity of $\tilde{O}(\epsilon^{-3})$ with lower communication complexity of $\tilde{O}(\epsilon^{-2})$ in finding $\epsilon$-stationary point of the nonconvex minimax problems. Experimentally, we conduct some experiments on the deep AUC maximization and robust neural network training tasks to verify efficiency of our algorithms.
\end{abstract}

\section{Introduction}
Minimax optimization, due to its hierarchical structure, is widely used in machine learning tasks such as adversarial training of Deep Neural Networks (DNNs)~\cite{tramer2018ensemble}, Generative Adversarial Networks (GANs)~\cite{goodfellow2014generative}, distributional robust learning~\cite{reisizadeh2020robust,deng2021distributionally} and reinforcement learning~\cite{wai2019variance}.
In the paper, we study a class of nonconvex distributed minimax optimization problems based on
the data distributed in multiple clients (such as mobile
devices, institutions, organizations, etc.),
defined as
\begin{align} \label{eq:1}
 \min_{x \in \mathbb{R}^d}\max_{y\in \mathbb{R}^p}  \ f(x,y)\equiv\frac{1}{K}\sum_{k=1}^Kf^k(x,y),
\end{align}
where $f^k(x,y)=\mathbb{E}_{\xi^k \sim \mathcal{D}^k}\big[f^k(x,y;\xi^k)\big]$ denotes the local objective function at $k$-th client for any $k\in[K]=\{1,2,\cdots,K\}$. The global objective function $f(x,y)$ is possibly nonconvex on the variable $x\in \mathbb{R}^d$, while it is
Strongly-Concave (SC) on the variable $y\in \mathbb{R}^p$, or is still nonconvex on the variable $y\in \mathbb{R}^p$ but satisfies Polyak-{\L}ojasiewicz (PL) condition~\cite{polyak1963gradient}.
Here $\xi^k$ for any $k\in[K]$ are independent random variables following unknown distributions $\mathcal{D}^k$, and for any $k,j\in [K]$ possibly $\mathcal{D}^k \neq \mathcal{D}^j$. Let
$y^*(x)\in \arg\max_{y\in \mathbb{R}^p} f(x,y)$ and $F(x)=f(x,y^*(x))$. In solving the above minimax problem (\ref{eq:1}), our goal is to search for an $\epsilon$-stationary solution, i.e.,
$\|\nabla F(x)\| \leq \epsilon \ (\epsilon\geq 0)$ as in~\cite{deng2021local,sharma2022federated}.
\begin{table*}
  \centering
  \caption{ \textbf{Gradient (i.e., SFO)} and \textbf{Communication} complexities comparison of the representative \textbf{federated minimax optimization} algorithms in searching for
  an $\epsilon$-stationary point of the NC-SC or NC-PL minimax problem \eqref{eq:1}, i.e., $\mathbb{E}\|\nabla F(x)\|\leq \epsilon$
  or its equivalent variants. \textbf{ALR} denotes adaptive learning rate.  }
  \label{tab:1}
   \resizebox{0.95\textwidth}{!}{
\begin{tabular}{c|c|c|c|c|c|c}
  \hline
   \textbf{Algorithm} & \textbf{Reference} & \textbf{Gradient} & \textbf{Communication}& \textbf{NC-SC} & \textbf{NC-PL}& \textbf{ALR}   \\ \hline
  Local-SGDA  & \cite{deng2021local}  & $O(\epsilon^{-6})$ & $O(\epsilon^{-4})$ & $\surd$ & $\surd$ &  \\  \hline
  FEDNEST  & \cite{tarzanagh2022fednest}  & $\tilde{O}(\epsilon^{-4})$ & $\tilde{O}(\epsilon^{-4})$ & $\surd$ & &\\  \hline
  Momentum-Local-SGDA & \cite{sharma2022federated} & $\tilde{O}(\epsilon^{-4})$ & $\tilde{O}(\epsilon^{-3})$ & $\surd$ & $\surd$ & \\  \hline
  SAGDA & \cite{yang2022sagda} & $O(\epsilon^{-4})$ & \textcolor{red}{$O(\epsilon^{-2})$} & $\surd$ & $\surd$ & \\  \hline
  CDMA & \cite{xie2023cdma} & \textcolor{red}{$\tilde{O}(\epsilon^{-3})$ } & $\tilde{O}(\epsilon^{-3})$ & $\surd$ & $\surd$ &  \\  \hline
  FGDA & Ours & \textcolor{red}{$\tilde{O}(\epsilon^{-3})$} & \textcolor{red}{$\tilde{O}(\epsilon^{-2})$} &$\surd$ & $\surd$ &  \\  \hline
  AdaFGDA & Ours & \textcolor{red}{$\tilde{O}(\epsilon^{-3})$} & \textcolor{red}{$\tilde{O}(\epsilon^{-2})$} &$\surd$ & $\surd$ & $\surd$ \\  \hline
\end{tabular}
 }
\end{table*}

When $K=1$ in Problem \eqref{eq:1}, i.e., non-distributed minimax optimization, \cite{lin2019gradient,lin2020near}
proposed the stochastic gradient descent ascent (SGDA) method, which is a simple generalization of stochastic gradient descent (SGD)~\cite{bottou2018optimization}. Specifically, it alternately conducts SGD for updating the variable $x$ and stochastic gradient ascent (SGA) for updating the variable $y$. Subsequently, some accelerated SGDA methods \cite{luo2020stochastic,yang2020catalyst,huang2022accelerated,huang2023adagda,yang2022nest} have been developed to solve the NonConvex-Strongly-Convex (NC-SC) minimax optimization at a single client. Meanwhile, \cite{nouiehed2019solving,yang2020global,chen2022faster,huang2023enhanced} studied the NonConvex-PL (NC-PL) minimax optimization. For example, \cite{luo2020stochastic} proposed an accelerated SGDA method (i.e.,SREDA) for NC-SC minimax optimization based on the variance reduced technique of SARAH~\cite{nguyen2017sarah}/SPIDER~\cite{fang2018spider}.
 An accelerated momentum-based SGDA method (i.e.,Acc-MDA)~\cite{huang2022accelerated} for NC-SC minimax optimization has been proposed by using the variance reduced technique of STORM~\cite{cutkosky2019momentum} without relying on the large batches. Meanwhile, \cite{chen2022faster} proposed an faster
AccSPIDER method to solve NC-PL minimax problems based on the SPIDER.
 \cite{huang2023adagda,yang2022nest,huang2023enhanced}
 studied the adaptive SGDA methods to solve the NC-SC or NC-PL minimax problems.
 \cite{chen2021proximal,huang2021efficient} studied the NC-SC minimax optimization with nonsmooth regularization.

The above proposed methods
mainly focus on solving the minimax optimization problems at a single client.
Recently, big data applications often rely on multiple
sources or clients for data collection. Clearly,
transferring all local data to a single server is undesirable, and
the data privacy is not be protected. Thus, recently some distributed optimization methods~\cite{tsaknakis2020decentralized,xian2021faster,deng2021local,tarzanagh2022fednest,sharma2022federated}
have been developed to solve the distributed NC-SC minimax problem \eqref{eq:1} with $K>1$.
For example, \cite{tsaknakis2020decentralized,xian2021faster} proposed
some effective decentralized methods to solve the distributed minimax optimization over decentralized networks.
In parallel, \cite{deng2021local} studied the federated learning methods for distributed minimax
 optimization over centralized networks with a server, and proposed an effective local-SGDA method. Subsequently,
 \cite{tarzanagh2022fednest,sharma2022federated} proposed some accelerated local-SGDA methods. More recently,  \cite{yang2022sagda} presented a class of new stochastic sampling averaging gradient descent ascent algorithms (i.e, SAGDA) for nonconvex-PL minimax optimization and obtain a lower communication complexity.
\cite{xie2023cdma} proposed an efficient momentum-based federated algorithm for NC-PL minimax optimization.
 Meanwhile, \cite{huang2023near} proposed a near-optimal momentum-based decentralized algorithm for NC-PL minimax optimization
 over a decentralized network.

Federated Learning (FL)~\cite{mcmahan2017communication} is an effective distributed and privacy-preserving learning paradigm in machine learning. 
In the paper, thus, we focus on the federated learning algorithms
for minimax optimization. From Table \ref{tab:1}, the existing FL methods for the NC-SC and NC-PL minimax problem
\eqref{eq:1} still suffer high gradient (i.e., stochastic first-order oracle, SFO) or communication complexities in searching for
an $\epsilon$-stationary point of the minimax problem \eqref{eq:1} (i.e., $\mathbb{E}\|\nabla F(x)\|\leq \epsilon$).
Thus there exists an open question:
\begin{center}
\begin{tcolorbox}
\textbf{ Could we develop federated algorithms with lower gradient and communication complexities simultaneously in finding
an $\epsilon$-stationary point of Problem \eqref{eq:1} ? }
\end{tcolorbox}
\end{center}
In the paper, we affirmatively answer to the above question and propose
a class of accelerated federated minimax optimization methods (i.e., FGDA and AdaFGDA) to solve
the NC-SC or NC-PL minimax problem \eqref{eq:1}, which build on the momentum-based variance reduced ~\cite{cutkosky2019momentum} and local-SGD~\cite{stich2019local} techniques.
In particular, our adaptive algorithm (i.e., AdaFGDA) can flexibly incorporate various adaptive learning rates by using the unified adaptive matrices.
Moreover, our FL methods obtain lower sample and communication complexities simultaneously.
In summary, our main contributions are:
\begin{itemize}
\item[(1)] We propose a class of accelerated federated minimax optimization methods (i.e., FGDA and AdaFGDA) to solve the minimax Problem \eqref{eq:1}. In particular, our AdaFGDA can use various adaptive learning rates.
\item[(2)] We provide a solid convergence analysis framework for our algorithms, and prove that they obtain lower gradient complexity of $\tilde{O}(\epsilon^{-3})$ with lower communication complexity of $\tilde{O}(\epsilon^{-2})$ in finding an $\epsilon$-stationary point of Problem~(\ref{eq:1}). From \cite{arjevani2019lower}, the optimal gradient complexity is $O(\epsilon^{-3})$ in finding
an $\epsilon$-stationary point of nonconvex smooth problem $\min_{x\in \mathbb{R}^d}f(x)$. Thus, our algorithms obtain the optimal gradient complexity with lower communication complexity.
\item[(3)] Experimental results demonstrate efficiency of our algorithms on the deep AUC maximization and robust neural network training tasks.
\end{itemize}

\begin{figure*}[ht]
\centering
  \subfigure{\includegraphics[width=0.45\textwidth]{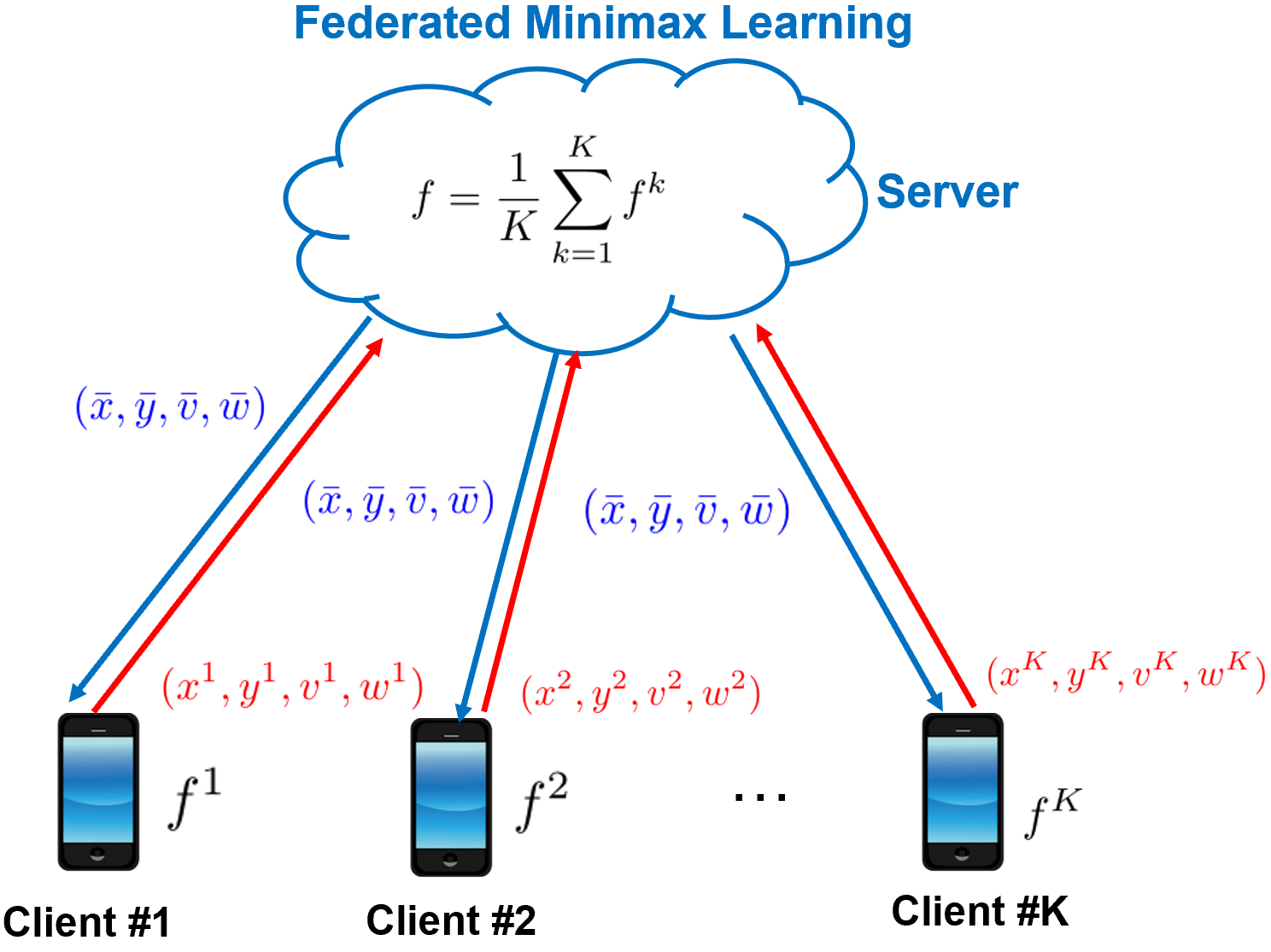}}
  \qquad
  \subfigure{\includegraphics[width=0.45\textwidth]{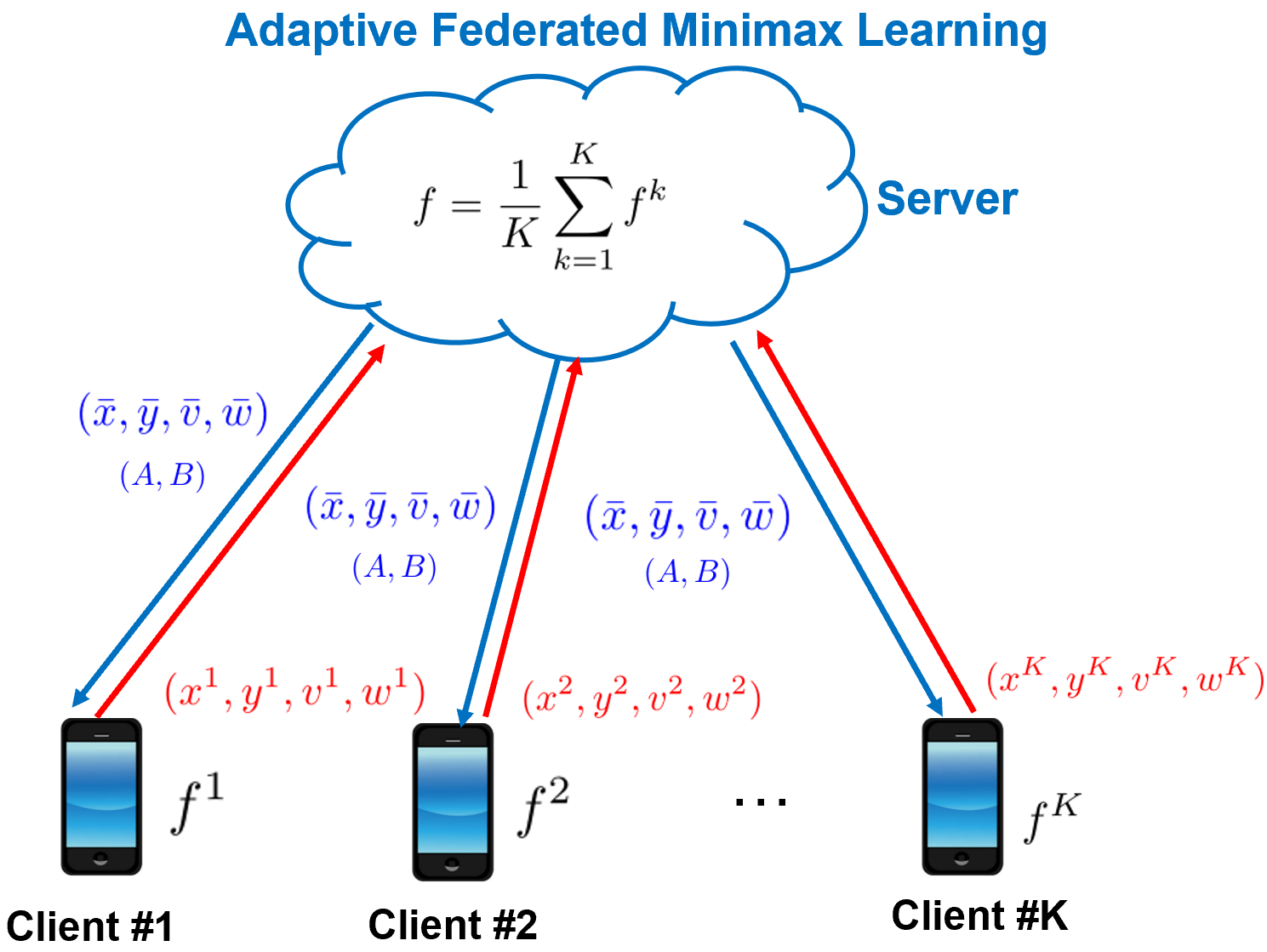}}
  \hfill
\caption{Depiction of our federated minimax algorithms, i.e., our \textbf{FGDA} (left) and \textbf{AdaFGDA} (right), $A$ and $B$ denote the adaptive diagonal matrices (or vectors).}
\label{fml}
\end{figure*}
\section{Related Works}
\vspace*{-6pt}
In this section, we overview some representative federated learning algorithms and
distributed minimax optimization, respectively.
\vspace*{-4pt}
\subsection{ Federated Learning Algorithms }
\vspace*{-6pt}
Federated Learning (FL)~\cite{mcmahan2017communication} is an effective distributed and privacy-preserving learning paradigm, which learns a global model from a set of located clients under the coordination of a server. In FL, the edge clients do not send
their data to the server to improve the privacy afforded to
the clients. Meanwhile, FL applies the local-SGD technique to reduce the cost of communication. 
FedAvg~\cite{mcmahan2017communication}/Local-SGD~\cite{stich2019local} algorithm is one of the earliest
FL algorithms, where each client takes multiple steps of SGD with its local data
and then sends the learned parameter to the server for averaging.
Recently, the convergence properties of local-SGD and FedAvg algorithms have been studied in \cite{li2019convergencefedvrg,khaled2020tighter,deng2021local,glasgow2022sharp}.
For example, \cite{li2019convergencefedvrg} provided the convergence analysis of FedAvg/local-SGD algorithms for strongly-convex optimization.
\cite{khaled2020tighter} studied the tight convergence rates of local-SGD for both convex and nonconvex optimizations.  Due to lacking of solution personalization, the basic FL methods often show poor
performances in the presence of local data heterogeneity deteriorating the performance of the global FL model
on individual clients. Thus, some personalized FL methods~\cite{t2020personalized,fallah2020personalized} recently have been studied. Meanwhile,
to accelerate the basic local-SGD and FedAvg, some accelerated FL algorithms~\cite{karimireddy2020scaffold,yuan2020federated,khanduri2021stem,das2022faster} are developed. For example, \cite{khanduri2021stem} proposed a faster FL algorithm for nonconvex optimization with
simultaneously near-optimal sample and communication complexities.
More recently, \cite{das2022faster} proposed a faster federated learning for nonconvex optimization via global and local momentums.
In parallel, some adaptive FL methods \cite{reddi2020adaptive,chen2020toward,li2022federated} have been developed to
accelerate the basic local-SGD and FedAvg algorithms. For example, \cite{reddi2020adaptive} proposed a class of adaptive FL algorithms via using adaptive learning rates at the server side. Meanwhile, an efficient local-AMSGrad algorithm~\cite{chen2020toward} has been proposed, where clients locally update variables by using
adaptive learning rates shared with all clients.
\vspace*{-4pt}
\subsection{ Distributed Minimax Optimization }
\vspace*{-6pt}
Minimax optimization is widely applied in many machine learning problems such as robust learning, fair learning
 and reinforcement learning. For the big data applications, recently, there exists an increasing interest in distributed minimax optimization, e.g., training robust Deep Neural Networks (DNNs) over multiple clients
 and policy evaluation over multi-agents.
Recently, decentralized optimization methods \cite{liu2020decentralized,beznosikov2021near,
rogozin2021decentralized,tsaknakis2020decentralized,zhang2021taming,xian2021faster} for distributed minimax optimization have been developed.
For example, \cite{tsaknakis2020decentralized} studied the decentralized optimization methods for the nonconvex-(strongly)-concave minimax optimization. Subsequently, \cite{xian2021faster} proposed a
faster decentralized minimax optimization method for NC-SC minimax optimization.
In parallel, some federated minimax optimization methods~\cite{reisizadeh2020robust,hou2021efficient,liao2021local,deng2021local,
tarzanagh2022fednest,sharma2022federated} have been developed to solve the distributed minimax problems.
For example, \cite{reisizadeh2020robust} studied the federated learning methods for NC-PL minimax optimization.
\cite{deng2021local} proposed a class of effective Local-SGDA methods for minimax optimization, and provide the convergence analysis for the general minimax optimization.
 \cite{yang2022sagda,sharma2023federated} proposed some communication-efficient federated algorithms for
 NC-SC/NC-PL minimax optimization.
Subsequently, \cite{tarzanagh2022fednest,sharma2022federated,xie2023cdma} proposed some accelerated Local-SGDA methods based on the variance reduced techniques.
\vspace*{-4pt}
\section{Preliminaries}
\vspace*{-6pt}
\subsection{Notations}
$[K]$ denotes the set $\{1,2,\cdots,K\}$.
$\|\cdot\|$ denotes the $\ell_2$ norm for vectors and spectral norm for matrices.
$\langle x,y\rangle$ denotes the inner product of two vectors $x$ and $y$. For vectors $x$ and $y$, $x^r \ (r>0)$ denotes the element-wise
power operation, $x/y$ denotes the element-wise division and $\max(x,y)$ denotes the element-wise maximum. $I_{d}$ denotes a $d$-dimensional identity matrix. Matrix $A \succ 0$ is positive definite.
Given function $f(x,y)$, $f(x,\cdot)$ denotes  function \emph{w.r.t.} the second variable with fixing $x$,
and $f(\cdot,y)$ denotes function \emph{w.r.t.} the first variable
with fixing $y$.
$a_m=O(b_m)$ denotes that $a_m \leq c b_m$ for some constant $c>0$. The notation $\tilde{O}(\cdot)$ hides logarithmic terms.

\begin{algorithm*}[t]
\caption{ \textbf{FGDA} and \textbf{AdaFGDA} Algorithms }
\label{alg:1}
\begin{algorithmic}[1]
\STATE {\bfseries Input:} $T, q$, tuning parameters $\{\gamma, \lambda, \eta_t, \alpha_t, \beta_t\}$, initial inputs $x_1\in \mathbb{R}^d$, $y_1\in \mathbb{R}^p$; \\
\STATE {\bfseries initialize:} Set $x^k_1=x_1$ and $y^k_1=y_1$ for $k \in [K]$, and draw $q$ samples $\{\xi^k_{1,j}\}_{j=1}^q$,
and then compute $v^k_1 = \frac{1}{q}\sum_{j=1}^q\nabla_y f^k(x^k_1,y^k_1;\xi^k_{1,j})$, and $w^k_1 = \frac{1}{q}\sum_{j=1}^q\nabla_xf^k(x^k_1,y^k_1;\xi^k_{1,j})$ for all $k \in [K]$,
and generate adaptive matrices $A_1 \in \mathbb{R}^{d \times d}$ and $B_1 \in \mathbb{R}^{p \times p}$. \\
\FOR{$t=1$ \textbf{to} $T$}
\IF {$\mod(t,q)=0$}
\STATE $\bar{v}_t = \frac{1}{K} \sum_{k=1}^{K} v^k_t$, $\bar{w}_t = \frac{1}{K} \sum_{k=1}^{K} w^k_t$, $\bar{y}_t = \frac{1}{K} \sum_{k=1}^{K} y^k_t$, $\bar{x}_t = \frac{1}{K} \sum_{k=1}^{K} x^k_t$; \\
\STATE Generate the adaptive matrices $A_t \in \mathbb{R}^{d \times d}$ and $B_t \in \mathbb{R}^{p \times p}$;\\
\textcolor{blue}{One example of $A_t$ and $B_t$ by using update rule ($a_0 = 0$, $b_0 = 0$, $ 0 < \varrho < 1$, $\rho>0$.) } \\
\textcolor{blue}{ Compute $ a_t = \varrho a_{t-1} + (1 - \varrho)\bar{w}_t^2$, $A_t = \mbox{diag}(\sqrt{a_t} + \rho)$}; \\
\textcolor{blue}{ Compute $ b_t = \varrho b_{t-1} + (1 - \varrho)\bar{v}_{t}^2$, $B_t = \mbox{diag}(\sqrt{b_t} + \rho)$}; \\
\STATE $\hat{y}^k_{t+1}=\hat{y}_{t+1} = \bar{y}_t + \lambda B_t^{-1}\bar{v}_t$, $\hat{x}^k_{t+1}=\hat{x}_{t+1} = \bar{x}_t - \gamma A_{t}^{-1} \bar{w}_t$; \\
\STATE $y^k_{t+1}=\bar{y}_{t+1} = \bar{y}_t + \eta_t(\hat{y}_{t+1}-\bar{y}_t)$,
$x^k_{t+1}=\bar{x}_{t+1} = \bar{x}_t + \eta_t(\hat{x}_{t+1}-\bar{x}_t)$; (Send them to Clients) \\
\ELSE
\FOR{each client $k\in [K]$ \ (\textbf{in parallel})}
\STATE $\hat{y}^k_{t+1} = y^k_t + \lambda B_t^{-1}v^k_t$, $\hat{x}^k_{t+1} = x^k_t - \gamma A_t^{-1} w^k_t$; \\
\STATE $y^k_{t+1} = y^k_t + \eta_t(\hat{y}^k_{t+1}-y^k_t)$, $x^k_{t+1} = x^k_t + \eta_t(\hat{x}^k_{t+1}-x^k_t)$; \\
\STATE $A_{t+1} = A_t$, $B_{t+1} = B_t$; \\
\STATE Draw one sample $\xi^k_{t+1}$ for any $k\in [K]$;
\STATE $v^k_{t+1} = \nabla_y f^k(x^k_{t+1},y^k_{t+1};\xi^k_{t+1}) + (1-\alpha_{t+1})\big[v^k_t - \nabla_y f^k(x^k_t,y^k_t;\xi^k_{t+1})\big]$; \\
\STATE $w^k_{t+1} = \nabla_x f^k(x^k_{t+1},y^k_{t+1};\xi^k_{t+1}) + (1-\beta_{t+1})\big[w^k_t - \nabla_x f^k(x^k_t,y^k_t;\xi^k_{t+1})\big] $; \\
\ENDFOR
\ENDIF
\ENDFOR
\STATE {\bfseries Output:} Chosen uniformly random from $\{\bar{x}_t,\bar{y}_t\}_{t=1}^{T}$.
\end{algorithmic}
\end{algorithm*}

\subsection{Some Assumptions}
\begin{assumption} \label{ass:1}
For any $k\in [K]$, the \textbf{local function} $f^k(x,y;\xi^k)$ has a $L_f$-Lipschitz gradient, e.g.,
 for all $x,x_1,x_2\in \mathbb{R}^d$ and $y, y_1,y_2 \in \mathbb{R}^p$, we have
\begin{align}
& \|\nabla_x f^k(x_1,y;\xi^k) - \nabla_x f^k(x_2,y;\xi^k)\| \leq L_f \|x_1 - x_2\|, \nonumber \\
& \|\nabla_y f^k(x,y_1;\xi^k) - \nabla_y f^k(x,y_2;\xi^k)\| \leq L_f \|y_1 - y_2\|. \nonumber
\end{align}
\end{assumption}

Assumption \ref{ass:1} imposes the smoothness of stochastic function $f(x,y;\xi)$ as in the variance reduced federated algorithm~\cite{khanduri2021stem}.
\begin{assumption} \label{ass:2a}
For $x\in \mathbb{R}^d$, the \textbf{global function} $f(x,y)=\frac{1}{K}\sum_{k=1}^Kf^k(x,y)$ is $\mu$-strongly concave on variable $y \in \mathbb{R}^p$, i.e., for all $x\in \mathbb{R}^d$ and $y, y' \in \mathbb{R}^p$,
we have
\begin{align}
 f(x,y) \leq f(x,y') + \langle\nabla_y f(x,y'), y-y'\rangle - \frac{\mu}{2}\|y-y'\|^2. \label{eq:3.2}
\end{align}
\end{assumption}

\begin{assumption} \label{ass:2}
For $x\in \mathbb{R}^d$, the \textbf{global function} $f(x,y)=\frac{1}{K}\sum_{k=1}^Kf^k(x,y)$ satisfies $\mu$-PL condition in variable $y \in \mathbb{R}^p$ for some $\mu>0$ if for any given $x\in \mathcal{X}$, it holds that
\begin{align}
 \|\nabla_y f(x,y')\|^2 \geq 2\mu \big(\max_y f(x,y)-f(x,y')\big), \ \forall y' \in \mathbb{R}^p. \label{eq:3.3}
\end{align}
\end{assumption}

By maximizing the inequality~(\ref{eq:3.2}) with respect to $y$, we have
\begin{align}
 & \max_y f(x,y) \nonumber \\
 & \leq \max_y \big\{ f(x,y') + \langle\nabla_y f(x,y'), y-y'\rangle - \frac{\mu}{2}\|y-y'\|^2 \big\}. \label{eq:3.4}
\end{align}
For its right hand side, we have
\begin{align}
 \nabla f(x,y') - \mu (y-y') = 0  \Rightarrow y = y' + \frac{1}{\mu}\nabla f(x,y').
\end{align}
Then putting $y = y' + \frac{1}{\mu}\nabla f(x,y')$ into the right hand side of the above inequality~(\ref{eq:3.4}), we have
\begin{align}
 \max_y f(x,y) \leq f(x,y') +  \frac{1}{2\mu}\|\nabla f(x,y')\|^2. \label{eq:3.5}
\end{align}
Then we can get
\begin{align}
 \|\nabla_y f(x,y')\|^2 \geq 2\mu \big(\max_y f(x,y)-f(x,y')\big), \ \forall y \in \mathbb{R}^p.
\end{align}
Thus, the strong concavity implies Polyak-Lojasiewicz inequality is satisfied. In other words, Assumption~\ref{ass:2} implies that Assumption~\ref{ass:2a} holds. In the following our convergence analysis, thus, we only use the above Assumption~\ref{ass:2}, i.e., satisfying PL condition.

\subsection{ Distributed Minimax Optimization }
In this subsection, we review the first-order method to solve the following distributed minimax optimization problem,
\begin{align} \label{eq:2}
 \min_{x \in \mathbb{R}^d} \max_{y \in \mathbb{R}^p} \ f(x,y)\equiv \frac{1}{K}\sum_{k=1}^K f^k(x,y).
\end{align}
For solving Problem \eqref{eq:2}, we can iteratively conduct the gradient descent for the variables $x$
and the gradient ascent for the variables $y$: at the $t$-th step
\begin{align}
 x_{t+1} = x_t - \gamma \nabla_x f(x_t,y_t), \quad y_{t+1} = y_t + \lambda \nabla_y f(x_t,y_t), \nonumber
\end{align}
where $\lambda>0$ and $\gamma>0$ denote the learning rates.
Based on the above Assumption \ref{ass:2}, the function $f(x,y) = \frac{1}{K}\sum_{k=1}^K f^k(x,y)$  satisfies PL condition in $y\in \mathbb{R}^p$. Thus, there exists a unique solution to
the problem $\max_{y\in \mathbb{R}^p} f(x,y)$ for any $x$.
Here we let $y^*(x)=\arg\max_{y \in \mathbb{R}^p} f(x,y)=\arg\max_{y \in \mathbb{R}^p} \frac{1}{K}\sum_{k=1}^K f^k(x,y)$, and
$F(x)=f(x,y^*(x))=\max_{y \in \mathbb{R}^p} \frac{1}{K}\sum_{k=1}^Kf^k(x,y)$.
In the paper, we mainly focus on the distributed stochastic minimax problem  (\ref{eq:1}). For any $k\in[K]$, $f^k(x,y)=\mathbb{E}_{\xi^k}\big[f^k(x,y;\xi^k)\big]$. 
Next, we review a useful lemma in \cite{nouiehed2019solving}.

\begin{lemma} \label{lem:1}
(Lemma A.5 of \cite{nouiehed2019solving})
Let $F(x)= f(x,y^*(x))$ with $y^*(x) \in \
\arg\max_y f(x,y)$. Under the above Assumptions~\ref{ass:1}, \ref{ass:2}, we have
$\nabla F(x)=\nabla_x f(x,y^*(x))$ and $F(x)$ is $L$-smooth, i.e.,
\begin{align}
\|\nabla F(x_1) - \nabla F(x_2)\| \leq L\|x_1-x_2\|, \quad \forall x_1,x_2
\end{align}
where $L=L_f(1+\frac{\kappa}{2})$ with $\kappa=\frac{L_f}{\mu}$.
\end{lemma}

\section{ Faster Federated Minimax Optimization Algorithms }
In this section, we propose a class of accelerated federated minimax optimization methods (i.e., FGDA and AdaFGDA) to solve Problem \eqref{eq:1}, based on the momentum-based variance reduced and
local-SGD techniques.
In particular, our AdaFGDA algorithm uses the unified adaptive matrices to flexibly incorporate various adaptive learning rates to update variables $x$ and $y$. 
Figure~\ref{fml} shows the basic idea of our federated minimax optimization algorithms.
Meanwhile, Algorithm \ref{alg:1} shows a procedure framework of our FGDA and AdaFGDA algorithms.

In Algorithm \ref{alg:1}, when $\mbox{mod}(t,q)= 0$ (i.e., synchronization step), the server receives the updated variables $\{x^k_t,y^k_t\}_{k=1}^K$ and
estimated stochastic gradients $\{w_t^k,v_t^k\}_{k=1}^K$ from the clients, and then averages them to obtain
the averaged variables $\{\bar{x}_t,\bar{y}_t\}$ and averaged gradients $\{\bar{w}_t,\bar{v}_t\}$. Based on
these averaged gradients, we can generate some adaptive matrices (i.e., adaptive learning rates). Besides one example given at the line 6 of Algorithm \ref{alg:1}, we can also generate many other adaptive matrices. 
For example, we can generate
adaptive matrices as in AdaBelief~\cite{zhuang2020adabelief} algorithm, defined as
\begin{align}
 & a_t = \varrho a_{t-1} + (1 - \varrho)\big(\bar{w}_t - \bar{w}_{t_0}\big)^2, \ A_t = \mbox{diag}(\sqrt{a_t} + \rho), \nonumber \\
 & b_t = \varrho b_{t-1} + (1 - \varrho)\big(\bar{v}_t - \bar{v}_{t_0}\big)^2, \ B_t = \mbox{diag}(\sqrt{b_t} + \rho), \nonumber
\end{align}
where $t_0=t-q$. 
We update the variables $x$ and $y$ in the server by using these adaptive matrices,
then sent the updated variables to each client.

When $\mbox{mod}(t,q) \neq 0$ (i.e., asynchronization step), the clients receive the updated variables $\{\bar{x}_t,\bar{y}_t\}$
and the generated adaptive matrices $\{A_t,B_t\}$ from the server. Then the clients use the momentum-based variance reduced technique of STORM~\cite{cutkosky2019momentum}/
ProxHSGD~\cite{tran2022hybrid}
to update the stochastic gradients based on local data: for $k\in [K]$
\begin{align}
v^k_{t+1} & = \nabla_y f^k(x^k_{t+1},y^k_{t+1};\xi^k_{t+1}) \nonumber \\
& \quad + (1-\alpha_{t+1})\big[v^k_t - \nabla_y f^k(x^k_t,y^k_t;\xi^k_{t+1})\big],  \nonumber
\end{align}
where $\alpha_{t+1}\in (0,1)$, and it is similar for $w^k_{t+1}$. Based on the estimated stochastic gradients and adaptive matrices, the clients
update the variables $\{x^k_t,y^k_t\}_{k=1}^K$, defined as
\begin{align}
 & \hat{y}^k_{t+1} = y^k_t - \lambda B_t^{-1} v^k_t, \ y^k_{t+1} = y^k_t + \eta_t(\hat{y}^k_{t+1}-y^k_t), \nonumber  \\
 & \hat{x}^k_{t+1} = x^k_t - \gamma A_t^{-1} w^k_t, \ x^k_{t+1} = x^k_t + \eta_t(\hat{x}^k_{t+1}-x^k_t). \nonumber
\end{align}
In our algorithms, all clients use the same adaptive matrices generated from the server to avoid model divergence.
\textbf{Note that} for our non-adaptive \textbf{FGDA} algorithm, we only set $A_t=I_d$ and $B_t=I_p$ for all $t\geq 1$ in Algorithm \ref{alg:1}.

\section{Convergence Analysis}
In this section, we study the convergence properties of our \textbf{FGDA} and \textbf{AdaFGDA} algorithms
under some mild assumptions.
All related proofs are provided in the Appendix.
We first review some useful lemmas and assumptions.

\begin{assumption} \label{ass:3}
For any $k\in [K]$, each component function $f^k(x,y;\xi^k)$ has an unbiased stochastic gradient with
bounded variance $\sigma^2$, i.e., for all $\xi^k\sim \mathcal{D}^k, x \in \mathbb{R}^d, y \in \mathbb{R}^p$
\begin{align}
& \mathbb{E}[\nabla f^k(x,y;\xi^k)] = \nabla f^k(x,y), \nonumber \\
& \mathbb{E}\|\nabla f^k(x,y)-\nabla f^k(x,y;\xi^k)\|^2 \leq \sigma^2. \nonumber
\end{align}
\end{assumption}

\begin{assumption} \label{ass:4}
 For any $k, j \in [K]$, $x\in \mathbb{R}^d$ and $y\in \mathbb{R}^p$, we have $\|\nabla_x f^{k}(x, y) -  \nabla_x f^{j}(x, y) \| \leq \delta_x$, $\|\nabla_y f^{k} (x, y) -  \nabla_y f^{j}(x, y)\| \leq \delta_y$,
 where $\delta_x>0$ and $\delta_y>0$ are constants.
\end{assumption}

\begin{assumption} \label{ass:5}
The function $F(x)=\arg\max_{y\in \mathbb{R}^p} f(x,y)$ is bounded below, \emph{i.e.,} $F^* = \inf_{x\in \mathbb{R}^d}F(x) > -\infty$.
\end{assumption}

\begin{assumption} \label{ass:6}
In our AdaFGDA algorithm, the adaptive matrices $A_t$ and $B_t$ for all $t\geq 1$ satisfy
$A_t \succeq \rho I_d \succ 0 $ and $ \rho_u I_p \succeq  B_t \succeq \rho_l I_p \succ 0 $, where $\rho_u\geq\rho_l=\rho>0$ is an appropriate positive number.
\end{assumption}

Assumption \ref{ass:3} shows that the stochastic gradients in each client are unbiased, and their variances are bounded, which is very common in the stochastic optimization~\cite{ghadimi2016mini,fang2018spider,cutkosky2019momentum}. Assumption \ref{ass:4} shows that
under non-i.i.d. setting, the data heterogeneity is bounded, which is very common in the federated optimization~\cite{khanduri2021stem,sharma2022federated}. Assumption~\ref{ass:5} guarantees the feasibility of Problem~\eqref{eq:1}. Assumption~\ref{ass:6} ensures that the adaptive matrices $A_t$ for all $t\geq 1$ are positive definite as in \cite{huang2021super,huang2023enhanced}.

Next, based on the above assumptions, 
we give the convergence properties of our \textbf{FGDA} and \textbf{AdaFGDA} algorithms.
\subsection{ Convergence Properties of AdaFGDA Algorithm }

\begin{theorem} \label{th:1}
Assume the sequence $\{\bar{x}_t,\bar{y}_t\}_{t=1}^T$ be generated from Algorithm \ref{alg:1}.
 Under the above Assumptions~\ref{ass:1},\ref{ass:2}-\ref{ass:6}, and let $\eta_t=\frac{nK^{1/3}}{(m+t)^{1/3}}$ for all $t\geq 0$, $\alpha_{t+1}=c_1\eta_t^2$, $\beta_{t+1}=c_2\eta_t^2$, $m \geq \max\Big(2,n^3, (c_1n)^3K, (c_2n)^3K,  \frac{K\big(12\sqrt{2}n\lambda qL_f\big)^3}{\rho^3}\Big)$, $n>0$, $c^2_1+c^2_2 \leq \frac{12^4\lambda^4q^2L^2_f}{\rho^4}$,
 $ c_1 \geq \frac{2}{3n^3K} + \frac{9\rho_uL_f^2}{2\mu^2\rho}$, $c_2 \geq \frac{2}{3n^3K} + \frac{9}{2}$, $\gamma=\tau\lambda$,
 $\tau \leq \min\big(\frac{\sqrt{5K}}{4\sqrt{2\Lambda}},1\big)$, $\gamma \leq  \min\big( \frac{m^{1/3}\rho}{4Ln}, \frac{\lambda\mu}{16\rho_uL},\frac{\rho_l\mu}{16\rho_uL^2_f}, \frac{2\lambda\mu^2\rho}{27L^2_f\rho_u}, \frac{\sqrt{K}\rho}{8\sqrt{3}L_f}\big)$,  $\lambda \leq \min\big( \frac{m^{1/3}}{4L_fn\rho_u},\frac{3\sqrt{5K}}{32\sqrt{2}\mu}\big)$, $0<\rho\leq 1$ and $0<\rho_u\leq \frac{135}{64\rho^2}$, we have
\begin{align}
 & \frac{1}{T}\sum_{t=1}^T\mathbb{E}\|\nabla F(\bar{x}_t)\| \nonumber \\
 & \leq \sqrt{\frac{1}{T}\sum_{t=1}^T\mathbb{E}\|A_t\|^2}\Big( \frac{\sqrt{3G}m^{1/6}}{K^{1/6}T^{1/2}} +
 \frac{\sqrt{3G}}{K^{1/6}T^{1/3}}\Big),
\end{align}
where $G = \frac{4(F(\bar{x}_1) - F^*)}{\rho\gamma n} + \frac{36\rho_u L^2_f}{\rho\lambda\mu^2n}\big(F(\bar{x}_1) -f(\bar{x}_1,\bar{y}_1) + \frac{8m^{1/3}\sigma^2}{qK^{4/3}n^2\rho} +  8Kn^2\Big( \frac{(c_1^2+c_2^2)\sigma^2}{\rho^2K} + \frac{\Lambda\Delta }{15K\lambda^2L^2_f}\Big)\ln(m+t)$,
 $\Delta = c^2_2\sigma^2 + c^2_1\sigma^2 + 3c^2_2\delta_x^2 + 3c^2_1\delta_y^2$ and $\Lambda= \frac{1}{16} + \frac{L^2_f\rho_u}{4\mu^2}+ \frac{16\lambda^2L^2_f}{K\rho^2}$.
\end{theorem}

\begin{remark}
Assume the bounded stochastic gradient $\|\nabla_x f^k(x^k_t,y^k_t;\xi^k_t)\| \leq C_{fx}$ for all $k\in [K]$, we have
$\|\frac{1}{K}\sum_{k=1}^K\nabla_x f^k(x^k_t,y^k_t;\xi^k_t)\| \leq C_{fx}$
As the existing adaptive algorithms such as Adam, the adaptive matrix $A_t$ generated from Algorithm \ref{alg:1}, we have $\sqrt{\frac{1}{T}\sum_{t=1}^T\mathbb{E}\|A_t\|^2} \leq \sqrt{2(C^2_{fx}+\rho^2)}$.
Similarly, assume the bounded stochastic gradient $\|\nabla_y f^k(x^k_t,y^k_t;\xi^k_t)\| \leq C_{fy}$ for all $k\in [K]$, we can obtain $\rho_u=O(1)$.
\end{remark}

\begin{remark}
Without loss of generality, let $k=O(1)$, $\rho=\rho_l=O(1)$, $c_1=O(1)$, $c_2=O(1)$ and $m=O(q^3)$,
we have $G=\tilde{O}(1)$ and $\sqrt{\frac{1}{T}\sum_{t=1}^T\mathbb{E}\|A_t\|^2}=O(1)$. Based on the above Theorem \ref{th:1}, let $q=T^{1/3}$,
we have
\begin{align}
\frac{1}{T}\sum_{t=1}^T\mathbb{E}\|\nabla F(\bar{x}_t)\| \leq \tilde{O}\Big(\frac{\sqrt{q}}{\sqrt{T}}+\frac{1}{T^{1/3}}\Big)
= \tilde{O}\Big(\frac{1}{T^{1/3}}\Big) \leq \epsilon, \nonumber
\end{align}
then we can obtain $T=O(\epsilon^{-3})$. Our AdaFGDA algorithm needs to compute two stochastic gradients at each iteration
except for the first iteration requires $2q$ stochastic gradients,
so it has a gradient (i.e., SFO) complexity of $2q+2T = \tilde{O}(\epsilon^{-3})$. Thus, our AdaFGDA algorithm requires $\tilde{O}(\epsilon^{-3})$
gradient complexity
and $\frac{T}{q} = T^{2/3}=\tilde{O}(\epsilon^{-2})$ communication complexity in searching for an $\epsilon$-stationary point of Problem~(\ref{eq:1}), which improves the existing federated minimax optimization methods by a factor of $O(\epsilon^{-1})$ in gradient or communication complexities (Please see Table~\ref{tab:1}).
\end{remark}

\subsection{ Convergence Properties of FGDA Algorithm }
\begin{theorem} \label{th:2}
Assume the sequence $\{\bar{x}_t,\bar{y}_t\}_{t=1}^T$ be generated from Algorithm \ref{alg:1} when
$A_t=I_d$ and $B_t = I_p$ for all $t\geq1$.
 Under the above Assumptions~\ref{ass:1},\ref{ass:2}-\ref{ass:5}, and let $\eta_t=\frac{nK^{1/3}}{(m+t)^{1/3}}$ for all $t\geq 0$, $\alpha_{t+1}=c_1\eta_t^2$, $\beta_{t+1}=c_2\eta_t^2$, $m \geq \max\Big(2,n^3, (c_1n)^3K, (c_2n)^3K, K\big(12\sqrt{2}n\lambda qL_f\big)^3 \Big)$, $n>0$, $c^2_1+c^2_2 \leq 12^4\lambda^4q^2L^2_f$,
 $ c_1 \geq \frac{2}{3n^3K} + \frac{9 L_f^2}{2\mu^2 }$, $c_2 \geq \frac{2}{3n^3K} + \frac{9}{2}$, $\gamma=\tau\lambda$,
 $\tau \leq \min\big(\frac{\sqrt{5K}}{4\sqrt{2\Lambda}},1\big)$, $\gamma \leq  \min\big( \frac{m^{1/3} }{4Ln}, \frac{\lambda\mu}{16 L},\frac{ \mu}{16 L^2_f}, \frac{2\lambda\mu^2 }{27L^2_f }, \frac{\sqrt{K} }{8\sqrt{3}L_f}\big)$, and $\lambda \leq \min\big( \frac{m^{1/3}}{4L_fn },\frac{3\sqrt{5K}}{32\sqrt{2}\mu}\big)$, we have
\begin{align}
 \frac{1}{T}\sum_{t=1}^T\mathbb{E}\|\nabla F(\bar{x}_t)\| \leq \frac{\sqrt{3G}m^{1/6}}{K^{1/6}T^{1/2}} +
 \frac{\sqrt{3G}}{K^{1/6}T^{1/3}},
\end{align}
where $G = \frac{4(F(\bar{x}_1) - F^*)}{ \gamma n} + \frac{36  L^2_f}{ \lambda\mu^2n}\big(F(\bar{x}_1) -f(\bar{x}_1,\bar{y}_1) + \frac{8m^{1/3}\sigma^2}{qK^{4/3}n^2 } +  8Kn^2\Big( \frac{(c_1^2+c_2^2)\sigma^2}{ K} + \frac{\Lambda\Delta }{15K\lambda^2L^2_f}\Big)\ln(m+t)$,
 $\Delta = c^2_2\sigma^2 + c^2_1\sigma^2 + 3c^2_2\delta_x^2 + 3c^2_1\delta_y^2$ and $\Lambda= \frac{1}{16} + \frac{L^2_f }{4\mu^2}+ \frac{16\lambda^2L^2_f}{K }$.
\end{theorem}

\begin{remark}
The proof of Theorem \ref{th:2} can follow the proofs of the above Theorem \ref{th:1} with $A_t=I_d$ and $B_t=I_p$ for all $t\geq 1$, and $\rho=\rho_u=\rho_l=1$.
Since the conditions of Theorem \ref{th:2} are similar to these of of Theorem \ref{th:1}, clearly, our FGDA algorithm still can obtain a lower gradient complexity of $\tilde{O}(\epsilon^{-3})$ and  lower communication complexity of $\tilde{O}(\epsilon^{-2})$ for finding an $\epsilon$-stationary point of Problem \eqref{eq:1}.
\end{remark}

\section{Numerical Experiments}
In this section, we perform numerical experiments on some federated minimax optimization problems to demonstrate the efficiency of our FGDA and AdaFGDA algorithms.
We compared our FGDA and AdaFGDA algorithms with state-of-the-art federated minimax optimization algorithms, including Local-SGDA~\cite{deng2021local}, Momentum-Local-SGDA~\cite{sharma2022federated}, CDMA~\cite{xie2023cdma} and FEDNEST~\cite{tarzanagh2022fednest}. All experiments are run over machine with Intel(R) Xeon(R) W-2255 CPU and Nvidia RTX2080ti(s).

\begin{figure}[ht]
\centering
  \subfigure{\includegraphics[width=0.235\textwidth]{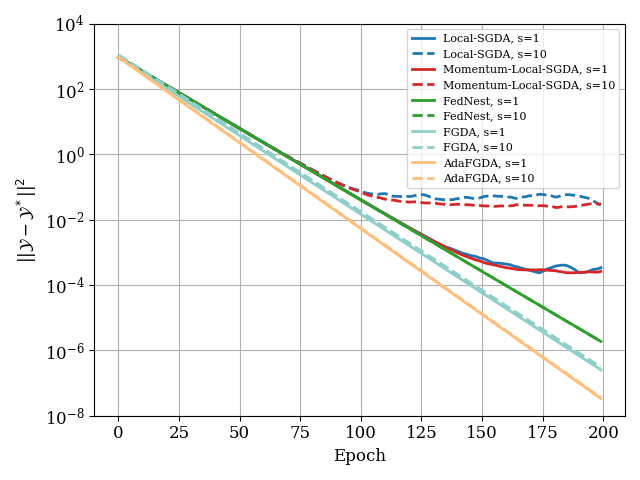}}
  \hfill
  \subfigure{\includegraphics[width=0.235\textwidth]{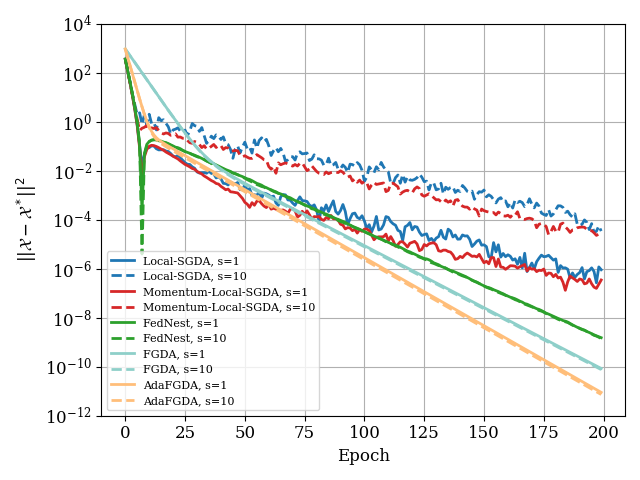}}
  \hfill
\caption{$L_2$ distance from the saddle-point $(x^*, y^*)$ with varying $s$.}
\label{synthetic}
\end{figure}

\subsection{Synthetic Federated Minimax Problem}
In the subsection, we conduct a synthetic federated minimax optimization problem as in \cite{tarzanagh2022fednest} formulated as:
\begin{align} \label{eq:11}
 \min_{x \in \mathbb{R}^d}\max_{y\in \mathbb{R}^p} \frac{1}{K}\sum_{k=1}^Kf^k(x,y),
\end{align}
where $f^k(x,y)=\frac{\tau}{2}{||x||}^2-\big(\frac{1}{2}{||y||}^2-b_k^Ty+y^TA_kx\big)$.
In fact, this minimax problem~(\ref{eq:11}) is expected to find a saddle point of the following problem:
\begin{align}
    & \min_{x \in \mathbb{R}^d} \ \frac{1}{2}{\Big|\Big|\frac{1}{K}\sum_{k=1}^K A_kx-b_k\Big|\Big|}^2, \nonumber \\
    & \mbox{s.t.} \ b_k=\Acute{b_k}-\frac{1}{K}\sum_{k=1}^K\Acute{b_k}, \ A_k=t_kI_d. \nonumber
\end{align}
 Here we set $\tau$ to 10 and sampled $\Acute{b_k}$ and $t_k$ from $\Acute{b_k}\sim \mathcal{N}(0, s^2I_d)$ and $t_k\sim U(0,0.1)$, respectively. In the experiment, we train the model for $200$ epochs as~\cite{tarzanagh2022fednest}.

Figure~\ref{synthetic} shows that our FGDA and AdaFGDA converge linearly despite the significant heterogeneity, which is a significant improvement over Local-SGDA and optimized Momentum-Local-SGDA. Our methods also achieve a faster and more stable convergence rate than FEDNEST for varying heterogeneities $s=1$ and $s=10$, where a larger $s$ represents a larger heterogeneity. We can witness a mutation on FEDNEST in $||y-y^*||^2$ in the early training stage. Although the synthetic data simulation experiment is relatively simple and ideal compared to general federated minimax optimization, it provides a more detailed and specific comparison than simulation experiments on real datasets since the optimal solution is available.

\begin{figure}[ht]
\centering
  \subfigure{\includegraphics[width=0.235\textwidth]{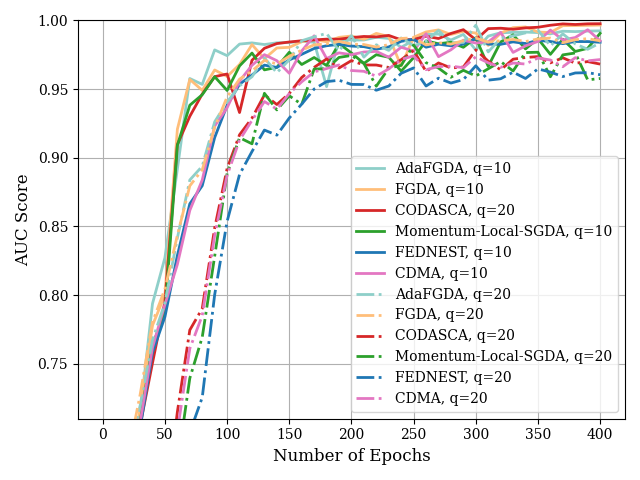}}
  \hfill
  \subfigure{\includegraphics[width=0.235\textwidth]{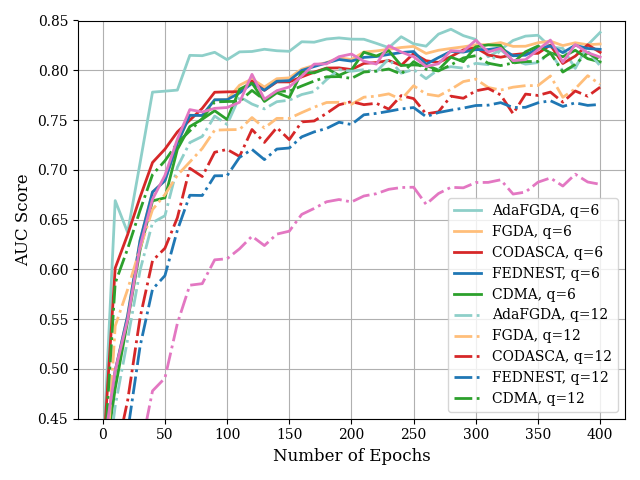}}
  \hfill
\caption{AUC Scores on \emph{MNIST} (left) and \emph{CIFAR10} (right).}
\label{auc1}
\end{figure}

\begin{figure*}[ht]
\subfigure{\includegraphics[width=0.3\textwidth]{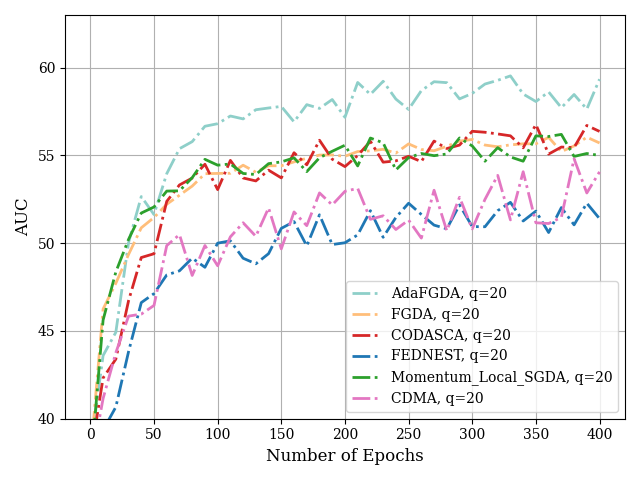}}
\subfigure{\includegraphics[width=0.3\textwidth]{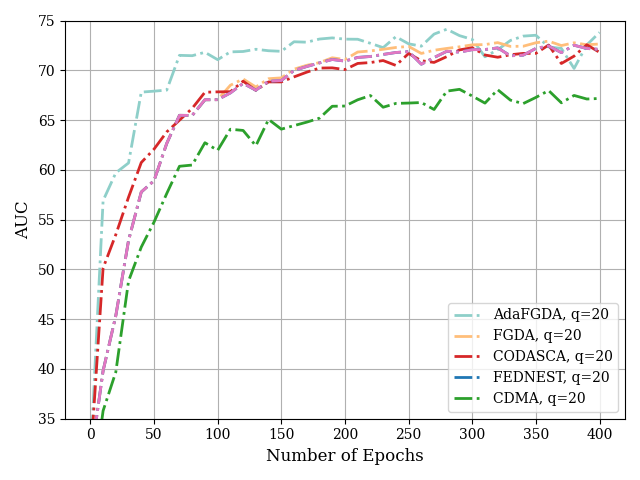}}
\subfigure{\includegraphics[width=0.3\textwidth]{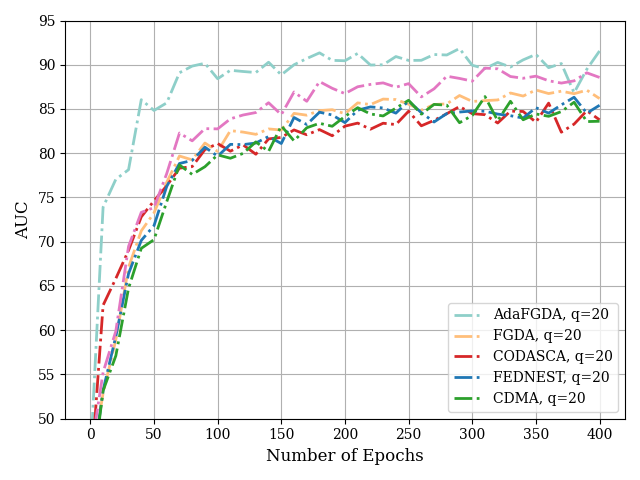}}
\caption{AUC Scores on \emph{ImageNet} (Left), \emph{CIFAR100} (Middle) and \emph{CheXpert} (Right). }
\label{auc2}
\end{figure*}

\subsection{Deep AUC Maximization}
 Data imbalance, where the number of samples from different classes is skewed, is a fundamental issue that can lead to model bias. Although Federated Learning (FL) offers an effective framework for leveraging multiple data sources, most existing FL methods still do not have the ability to address model bias caused by data imbalance, especially when such imbalance varies across clients. In this subsection, we adopt the definition in \cite{yuan2021federated} to maximize the AUC by using the minimax formulation for the distributed problem:
\begin{align}
    \min_{w \in \mathbb{R}^d, (a,b)\in \mathbb{R}^2}\max_{\alpha\in \mathbb{R}} &\frac{1}{K}\sum_{k=1}^K f_k(w,a,b,\alpha),
\end{align}
where
\begin{align}
& f_k(w,a,b,\alpha) \nonumber \\
& =\mathbb{E}_{z^k}\Big[(1-p){(h(w;x^k)-a)}^2\mathbb{I}_{[y^k=1]} \nonumber \\
    & \qquad +p{(h(w;x^k)-b)}^2\mathbb{I}_{[y^k=-1]} \nonumber \\
    &\qquad +2(1+\alpha)\Big(ph(w;x^k)\mathbb{I}_{[y^k=-1]} \nonumber \\
    &\qquad \quad -(1-p)h(w;x^k)\mathbb{I}_{[y^k=1]}\Big)-p(1-p)\alpha^2\Big], \nonumber
\end{align}
$z^k=(x^k,y^k)\sim\mathbb{P}_k$, and $\mathbb{P}_k$ is the data distribution on client $k$, and $p=\mbox{Pr}(y^k=1)$ is the ratio of positive data, and $h(w;x^k)$ denotes the prediction of the neural network on an input data $x^k$.

We evaluate the efficiency of our proposed FGDA and AdaFGDA on five benchmark datasets, i.e., MNIST, CIFAR10, CIFAR100, ImageNet, CheXpert, and compare them with state-of-the-art federated minimax optimization algorithms listed in Table~\ref{tab:1}, as well as a specially designed algorithm for AUC maximization, CODASCA~\cite{yuan2021federated}.

For the CIFAR10 experiment, we set the local iterations to $q=20$ for all methods and use a 7-layer CNN for training, while for MNIST, we set it to $q=12$ and use a LeNet5 for training. To construct the imbalanced heterogeneous dataset, we manually select 5 classes as positive and 5 classes as negative and split them further into different groups to increase data heterogeneity. Each split contains only samples from a unique class distribution that does not overlap with other groups. We refer to the proportion of positive samples in all samples as the imbalance ratio $p$, which is set to 5\% during training.
For the CIFAR100, CheXpert~\cite{irvin2019chexpert} and ImageNet datasets, we employed ResNet50 as the backbone model and set the local iterations to $q=20$ for all methods. To establish binary classification tasks, we manually selected two meta-classes (animal vs. machine) for ImageNet and CIFAR100 datasets, while the task of distinguishing between normal and abnormal samples naturally emerged in the case of CheXpert.

In the experiment, we use a grid search approach to determine the optimal hyper-parameters for all methods. Since each client's data distribution is heterogeneous, tuning a personalized model for each client can provide additional benefits. We typically set both the primal step size $\lambda$ and dual step size $\gamma$ in our proposed FGDA and AdaFGDA to 0.1 for selecting the learning rate. The learning rate $\eta$ is set to 0.001 when performing comparisons.

As shown in Figures~\ref{auc1} and~\ref{auc2}, our proposed FGDA and AdaFGDA algorithms achieve state-of-the-art performance and convergence rate compared to existing methods. Notably, we find that the Local-SGDA fails to converge in both datasets, and Momentum-Local-SGDA performs poorly on CIFAR10, so we do not report their results in Figure~\ref{auc1}. In more challenging settings such as ImageNet, our proposed methods display superior performance and convergence rates, which confirms the efficiency of our methods.

\begin{figure}[ht]
\centering  \subfigure{\includegraphics[width=0.235\textwidth]{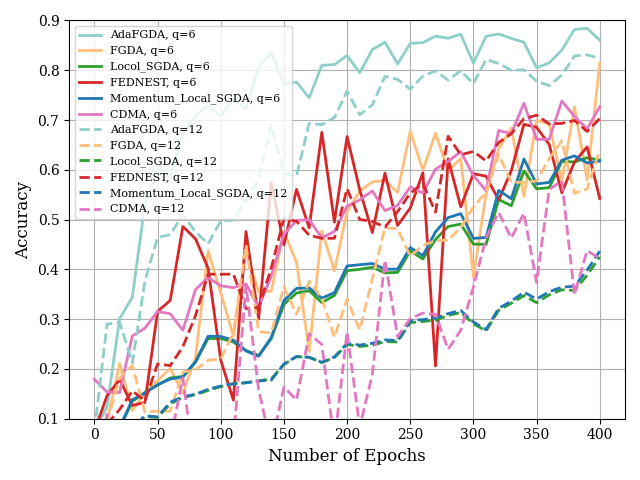}}
  \hfill
  \subfigure{\includegraphics[width=0.235\textwidth]{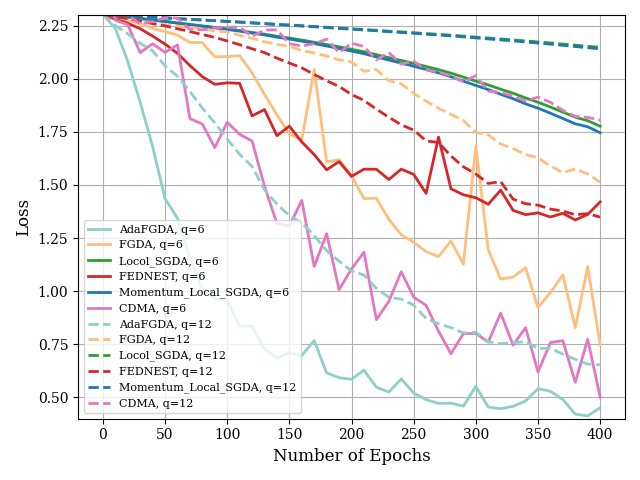}}
  \hfill
\caption{Test Accuracy for the robust NN training problem on the MNIST dataset, with 3-layer MLP. A comparison of different $q$ is also provided.}
\label{robustm}
\end{figure}

\begin{figure}[ht]
\centering
  \subfigure{\includegraphics[width=0.235\textwidth]{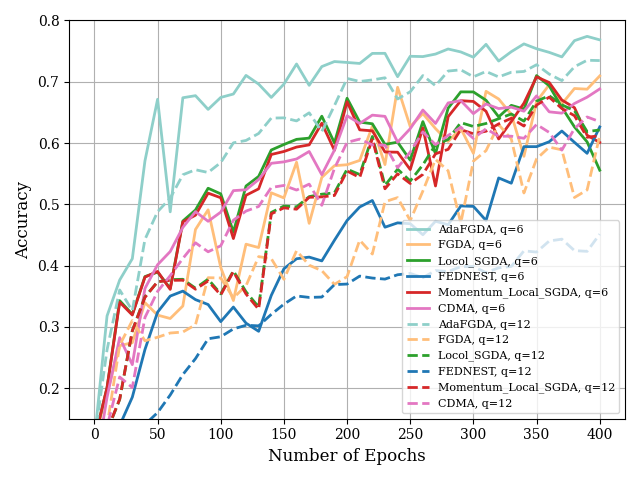}}
  \hfill  \subfigure{\includegraphics[width=0.235\textwidth]{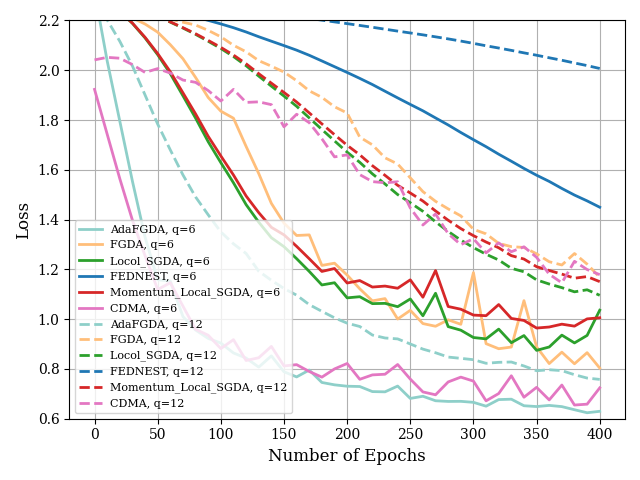}}
  \hfill
\caption{Test Accuracy for the robust NN training problem on the FashionMNIST dataset, with 3 layer MLP. A comparison of different $q$ is also provided.}
\label{robustf}
\end{figure}

\subsection{Robust Neural Network Training}
In this subsection, we will address the problem of training robust neural networks (NNs) in the presence of adversarial perturbations \cite{madry2017towards,sinha2017certifying}. We adopt a similar problem setting as in \cite{deng2021local}:
\begin{align}
    \min_{w \in \mathbb{R}^d}\max_{||\nu||^2\leq1}
    \frac{1}{K}\sum^K_{k=1}\sum^N_{i=1}\big(h(w;x^k_i+\nu),y^k_i\big)^2,
\end{align}
where $w$ denotes the parameters of the NNs, and $\varrho$ denotes the perturbations, and $(x^k_i,y^k_i)$ denotes the $i$-th data sample in the $k$-th client.

In the experiment, we use a 3-layer MLP network and use different values of $q\in\{6, 12\}$.
We compare our methods with state-of-the-art
federated minimax optimization algorithms listed in Table~\ref{tab:1}.
For both Local SGDA+ and Momentum Local SGDA+, we use $S=q^2$ as the algorithm of \cite{sharma2022federated}. For $\gamma$ and $\lambda$ in our methods, we use a grid search approach to determine the optimal hyper-parameters ranging from $\{0.001, 0.01, 0.02, 0.05, 0.1\}$. It can be seen from Figures~\ref{robustm} and~\ref{robustf}, our proposed AdaFDGA method provides a significant speedup over FDGA and achieves the best performance (superior test accuracy and faster convergence rates) among all the algorithms. For all three experiments, we select the momentum parameter from the range of $\{0.2,0.5,0.9,0.95\}$.

\section{Conclusion}
In the paper, we studied a class of distributed nonconvex minimax optimization under non-i.i.d. setting, and proposed a class of efficient adaptive federated minimax optimization methods (i.e., AdaFGDA and FGDA) based on momentum-based variance reduced and local-SGD techniques. Moreover, we provided a convergence analysis framework for our methods and proved that they obtain lower gradient and communication complexities simultaneously.

\subsubsection*{Acknowledgements}
We thank the anonymous reviewers for their valuable comments.
This work was partially supported by NSFC under Grant No.62376125, No.62076124. It was also partially supported by the Fundamental Research Funds for the Central Universities NO.NJ2023032.  Feihu Huang is the corresponding author (huangfeihu2018@gmail.com).

\small

 \bibliographystyle{abbrv}

\bibliography{AdaFGDA}

\section*{Checklist}

%

 \begin{enumerate}

 \item For all models and algorithms presented, check if you include:
 \begin{enumerate}
   \item A clear description of the mathematical setting, assumptions, algorithm, and/or model. [Yes]
   \item An analysis of the properties and complexity (time, space, sample size) of any algorithm. [Yes]
   \item (Optional) Anonymized source code, with specification of all dependencies, including external libraries. [No]
 \end{enumerate}

 \item For any theoretical claim, check if you include:
 \begin{enumerate}
   \item Statements of the full set of assumptions of all theoretical results. [Yes]
   \item Complete proofs of all theoretical results. [Yes]
   \item Clear explanations of any assumptions. [Yes]
 \end{enumerate}

 \item For all figures and tables that present empirical results, check if you include:
 \begin{enumerate}
   \item The code, data, and instructions needed to reproduce the main experimental results (either in the supplemental material or as a URL). [Yes]
   \item All the training details (e.g., data splits, hyperparameters, how they were chosen). [Yes]
         \item A clear definition of the specific measure or statistics and error bars (e.g., with respect to the random seed after running experiments multiple times). [Yes]
         \item A description of the computing infrastructure used. (e.g., type of GPUs, internal cluster, or cloud provider). [Yes]
 \end{enumerate}

 \item If you are using existing assets (e.g., code, data, models) or curating/releasing new assets, check if you include:
 \begin{enumerate}
   \item Citations of the creator If your work uses existing assets. [Yes]
   \item The license information of the assets, if applicable. [Not Applicable]
   \item New assets either in the supplemental material or as a URL, if applicable. [Not Applicable]
   \item Information about consent from data providers/curators. [Yes]
   \item Discussion of sensible content if applicable, e.g., personally identifiable information or offensive content. [Not Applicable]
 \end{enumerate}

 \item If you used crowdsourcing or conducted research with human subjects, check if you include:
 \begin{enumerate}
   \item The full text of instructions given to participants and screenshots. [Not Applicable]
   \item Descriptions of potential participant risks, with links to Institutional Review Board (IRB) approvals if applicable. [Not Applicable]
   \item The estimated hourly wage paid to participants and the total amount spent on participant compensation. [Not Applicable]
 \end{enumerate}

 \end{enumerate}

\appendix
\onecolumn

\section{Appendix}
In this section, we provide the detailed convergence analysis of our algorithms.

We first introduce some useful notations: $\bar{v}_t = \frac{1}{K} \sum_{k=1}^{K} v^k_t$, $\bar{w}_t = \frac{1}{K} \sum_{k=1}^{K} w^k_t$,
$\bar{x}_t = \frac{1}{K} \sum_{k=1}^{K} x^k_t$, $\hat{x}_t = \frac{1}{K}\sum_{k=1}^K\hat{x}^k_t$,
$\bar{y}_t = \frac{1}{K} \sum_{k=1}^{K} y^k_t$, $\hat{y}_t = \frac{1}{K}\sum_{k=1}^K\hat{y}^k_t$,
\begin{align}
& F(x) = \frac{1}{K}\sum_{k=1}^K f^k(x,y^*(x)), \quad \nabla_x f(x,y) = \frac{1}{K}\sum_{k=1}^{K}\nabla_x f^k(x,y), \quad \nabla_y f(x,y) = \frac{1}{K}\sum_{k=1}^{K}\nabla_y f^k(x,y), \nonumber \\
& \overline{\nabla_xf(x_t,y_t)} = \frac{1}{K}\sum_{k=1}^K\nabla_x f^k(x^k_t,y^k_t), \quad \overline{\nabla_y f(x_t,y_t)} = \frac{1}{K}\sum_{k=1}^K \nabla_y f^k(x^k_t,y^k_t), \quad \forall t\geq 1. \nonumber
\end{align}

Next, we review and provide some useful lemmas.

\begin{lemma} \label{lem:A2}
(\cite{karimi2016linear})
 Function $f(x): \mathbb{R}^d\rightarrow \mathbb{R}$ is $L$-smooth and satisfies PL condition with constant $\mu>0$, then it also satisfies error bound (EB) condition with $\mu$, i.e., for all $x \in \mathbb{R}^d$
\begin{align}
 \|\nabla f(x)\| \geq \mu\|x^*-x\|,
\end{align}
where $x^* \in \arg\min_{x} f(x)$. It also satisfies quadratic growth (QG) condition with $\mu$, i.e.,
\begin{align}
 f(x)-\min_x f(x) \geq \frac{\mu}{2}\|x^*-x\|^2.
\end{align}
\end{lemma}
From the above lemma~\ref{lem:A2}, when consider the problem $\max_x f(x)$ that is equivalent to the problem $-\min_x -f(x)$, we have
\begin{align}
 & \|\nabla f(x)\| \geq \mu\|x^*-x\|, \\
 & \max_x f(x) - f(x) \geq \frac{\mu}{2}\|x^*-x\|^2.
\end{align}

\begin{lemma} \label{lem:A3}
\cite{nesterov2018lectures}
Assume function $f(x)$ is convex and $\mathcal{X}$ is a convex set.
 $x^* \in \mathcal{X}$ is the solution of the
constrained problem $\min_{x\in \mathcal{X}}f(x)$, if
\begin{align}
 \langle \nabla f(x^*), x-x^*\rangle \geq 0, \ \forall x\in \mathcal{X}.
\end{align}
where $\nabla f(x^*)$ denote the (sub-)gradient of function $f(x)$ at $x^*$.
\end{lemma}

\begin{lemma} \label{lem:A4}
Given $K$ vectors $\{v^k\}_{k=1}^K$, the following inequalities satisfy: $||v^k + v^j||^2 \leq (1+\alpha)||v^k||^2 + (1+\frac{1}{\alpha})||v^j||^2$ for any $\alpha > 0$, and
$||\sum_{k=1}^K v^k||^2 \le K\sum_{k=1}^{K} ||v^k||^2$.
\end{lemma}

\begin{lemma} \label{lem:A5}
Given a finite sequence $\{w^{k}\}_{k=1}^K$, and $\bar{w} = \frac{1}{K}\sum_{k=1}^K w^{k}$,
the following inequality satisfies $\sum_{k=1}^K \|w^{k} - \bar{w}\|^2 \leq \sum_{k=1}^K \|w^{k}\|^2$.
\end{lemma}

\begin{lemma} \label{lem:C1}
Suppose the sequence $\{\bar{x}_t,\bar{y}_t\}_{t=1}^T$ be generated from Algorithm~\ref{alg:1}.
Under the Assumptions~\ref{ass:1},\ref{ass:2}, given $0<\gamma\leq \min\big(\frac{\lambda\mu}{16\rho_uL},\frac{\rho_l\mu}{16\rho_uL^2_f}\big)$ and $0<\lambda \leq \frac{1}{2\eta_t L_f\rho_u}$ for all $t\geq 1$, we have
\begin{align}
F(\bar{x}_{t+1}) - f(\bar{x}_{t+1},\bar{y}_{t+1})
& \leq (1-\frac{\eta_t\lambda\mu}{2\rho_u}) \big(F(\bar{x}_t) -f(\bar{x}_t,\bar{y}_t)\big) + \frac{\eta_t}{8\gamma}\|\hat{x}_{t+1}-\bar{x}_t\|^2  -\frac{\eta_t}{4\lambda\rho_u}\|\hat{y}_{t+1}-\bar{y}_t\|^2 \nonumber \\
& \quad + \frac{\eta_t\lambda}{\rho_l}\|\nabla_y f(\bar{x}_t,\bar{y}_t)-\bar{v}_t\|^2,
\end{align}
where $F(\bar{x}_t)=f(\bar{x}_t,y^*(\bar{x}_t))$ with $y^*(\bar{x}_t) \in \arg\max_{y}f(\bar{x}_t,y)$ for all $t\geq 1$.
\end{lemma}

\begin{proof}
Using $L_f$-smoothness of $f(x,\cdot)$, such that
\begin{align}
    f(\bar{x}_{t+1},\bar{y}_t) + \langle \nabla_y f(\bar{x}_{t+1},\bar{y}_t), \bar{y}_{t+1}-\bar{y}_t \rangle - \frac{L_f}{2}\|\bar{y}_{t+1}-\bar{y}_t\|^2 \leq f(\bar{x}_{t+1},\bar{y}_{t+1}),
\end{align}
then we have
\begin{align} \label{eq:B51}
    f(\bar{x}_{t+1},\bar{y}_t) & \leq f(\bar{x}_{t+1},\bar{y}_{t+1}) - \langle \nabla_y f(\bar{x}_{t+1},\bar{y}_t), \bar{y}_{t+1}-\bar{y}_t \rangle + \frac{L_f}{2}\|\bar{y}_{t+1}-\bar{y}_t\|^2 \nonumber \\
    & = f(\bar{x}_{t+1},\bar{y}_{t+1}) - \eta_t\langle \nabla_y f(\bar{x}_{t+1},\bar{y}_t), \hat{y}_{t+1}-\bar{y}_t \rangle + \frac{L_f\eta^2_t}{2}\|\hat{y}_{t+1}-\bar{y}_t\|^2.
\end{align}

Since $\rho_u I_{p} \succeq B_t \succeq \rho_l I_{p} \succ 0$ for any $t\geq 1$ is positive definite, we set $B_t=L_t(L_t)^T$, where $ \sqrt{\rho_u} I_{p} \succeq L_t \succeq \sqrt{\rho_l} I_{p} \succ 0$. Thus, we have $B^{-1}_t=(L^{-1}_t)^{T}L^{-1}_t$, where $ \frac{1}{\sqrt{\rho_l}}I_{p} \succeq L^{-1}_t \succeq \frac{1}{\sqrt{\rho_u}}I_{p} \succ 0$.

When $t=s_t=q\lfloor t/q\rfloor +1$, according to the line 7 of Algorithm \ref{alg:1}, we have
$\hat{y}_{t+1}=\bar{y}_t + \lambda B_t^{-1}\bar{v}_t$.
When $t\in (s_t,s_t+q)$, according to the line 11 of Algorithm \ref{alg:1}, we have for all $k\in [K]$, $\hat{y}^k_{t+1} = y^k_t + \lambda B_t^{-1}v^k_t$, and then we also have $\hat{y}_{t+1} = \frac{1}{K}\sum_{k=1}^K\hat{y}^k_{t+1} = \frac{1}{m}\sum_{k=1}^K\big(y^k_t + \lambda v^k_t\big)=\bar{y}_t + \lambda B_t^{-1}\bar{v}_t$.

Next, we bound the inner product in \eqref{eq:B51}. According to
$\hat{y}_{t+1} = \bar{y}_t + \lambda B_t^{-1}\bar{v}_t$, we have
\begin{align} \label{eq:B52}
    & - \eta_t\langle \nabla_y f(\bar{x}_{t+1},\bar{y}_t), \hat{y}_{t+1}-\bar{y}_t \rangle \nonumber \\
    & = - \eta_t\lambda\langle \nabla_y f(\bar{x}_{t+1},\bar{y}_t), B_t^{-1}\bar{v}_t \rangle \nonumber \\
    & = - \eta_t\lambda\langle L_t^{-1} \nabla_y f(\bar{x}_{t+1},\bar{y}_t), L_t^{-1}\bar{v}_t \rangle
    \nonumber \\
    & = -\frac{\eta_t\lambda}{2}\Big( \|L_t^{-1}\nabla_y f(\bar{x}_{t+1},\bar{y}_t)\|^2 + \|L_t^{-1}\bar{v}_t\|^2 - \|L_t^{-1}\nabla_y f(\bar{x}_{t+1},\bar{y}_t)-L_t^{-1}\nabla_y f(\bar{x}_t,\bar{y}_t) + L_t^{-1}\nabla_y f(\bar{x}_t,\bar{y}_t)-L_t^{-1}\bar{v}_t\|^2 \Big) \nonumber \\
    & \leq -\frac{\eta_t\lambda}{2\rho_u} \|\nabla_y f(\bar{x}_{t+1},\bar{y}_t)\|^2 -\frac{\eta_t}{2\lambda\rho_u} \|\hat{y}_{t+1}-\bar{y}_t\|^2 + \frac{\eta_t\lambda L^2_f}{\rho_l} \|\bar{x}_{t+1}-\bar{x}_t\|^2 + \frac{\eta_t\lambda}{\rho_l}\|\nabla_y f(\bar{x}_t,\bar{y}_t)-\bar{v}_t\|^2 \nonumber \\
    & \leq -\frac{\eta_t\lambda\mu}{\rho_u}\big(F(\bar{x}_{t+1})-f(\bar{x}_{t+1},\bar{y}_t)\big)-\frac{\eta_t}{2\lambda\rho_u} \|\hat{y}_{t+1}-\bar{y}_t\|^2 + \frac{\eta_t\lambda L^2_f}{\rho_l} \|\bar{x}_{t+1}-\bar{x}_t\|^2 + \frac{\eta_t\lambda}{\rho_l}\|\nabla_y f(\bar{x}_t,\bar{y}_t)-\bar{v}_t\|^2,
\end{align}
where the last inequality is due to the quadratic growth condition of $\mu$-PL functions, i.e.,
\begin{align}
    \|\nabla_y f(\bar{x}_{t+1},\bar{y}_t)\|^2 \geq 2\mu\big( \max_{y'}f(\bar{x}_{t+1},y')-f(\bar{x}_{t+1},\bar{y}_t)\big) = 2\mu\big( F(\bar{x}_{t+1})-f(\bar{x}_{t+1},\bar{y}_t)\big).
\end{align}
Substituting \eqref{eq:B52} in \eqref{eq:B51}, we have
\begin{align} \label{eq:B53}
    f(\bar{x}_{t+1},\bar{y}_t)
    & =f(\bar{x}_{t+1},\bar{y}_{t+1})-\frac{\eta_t\lambda\mu}{\rho_u}\big(F(\bar{x}_{t+1})-f(\bar{x}_{t+1},\bar{y}_t)\big)
    -\frac{\eta_t}{2\lambda\rho_u} \|\hat{y}_{t+1}-\bar{y}_t\|^2 + \frac{\eta_t\lambda L^2_f}{\rho_l} \|\bar{x}_{t+1}-\bar{x}_t\|^2 \nonumber \\
    & \quad + \frac{\eta_t\lambda}{\rho_l}\|\nabla_y f(\bar{x}_t,\bar{y}_t)-\bar{v}_t\|^2 + \frac{L_f\eta^2_t}{2}\|\hat{y}_{t+1}-\bar{y}_t\|^2,
\end{align}
then rearranging the terms, we can obtain
\begin{align} \label{eq:B54}
    F(\bar{x}_{t+1}) - f(\bar{x}_{t+1},\bar{y}_{t+1})
    & = (1-\frac{\eta_t\lambda\mu}{\rho_u})\big(F(\bar{x}_{t+1})-f(\bar{x}_{t+1},\bar{y}_t)\big)
    -\frac{\eta_t}{2\lambda\rho_u} \|\hat{y}_{t+1}-\bar{y}_t\|^2 + \frac{\eta_t\lambda L^2_f}{\rho_l} \|\bar{x}_{t+1}-\bar{x}_t\|^2 \nonumber \\
    & \quad + \frac{\eta_t\lambda}{\rho_l}\|\nabla_y f(\bar{x}_t,\bar{y}_t)-\bar{v}_t\|^2 + \frac{L_f\eta^2_t}{2}\|\hat{y}_{t+1}-\bar{y}_t\|^2.
\end{align}

Next, using $L_f$-smoothness of function $f(\cdot,\bar{y}_t)$, such that
\begin{align}
    f(\bar{x}_t,\bar{y}_t) + \langle \nabla_x f(\bar{x}_t,\bar{y}_t), \bar{x}_{t+1}-\bar{x}_t \rangle - \frac{L_f}{2}\|\bar{x}_{t+1}-\bar{x}_t\|^2 \leq f(\bar{x}_{t+1},\bar{y}_t),
\end{align}
then we have
\begin{align}
    & f(\bar{x}_t,\bar{y}_t) -f(\bar{x}_{t+1},\bar{y}_t)  \nonumber \\
    & \leq -\langle \nabla_x f(\bar{x}_t,\bar{y}_t), \bar{x}_{t+1}-\bar{x}_t \rangle + \frac{L_f}{2}\|\bar{x}_{t+1}-\bar{x}_t\|^2 \nonumber \\
    & = -\eta_t\langle \nabla_x f(\bar{x}_t,\bar{y}_t) - \nabla F(\bar{x}_t), \hat{x}_{t+1}-\bar{x}_t \rangle - \eta_t\langle \nabla F(\bar{x}_t), \hat{x}_{t+1}-\bar{x}_t \rangle + \frac{L_f\eta^2_t}{2}\|\hat{x}_{t+1}-\bar{x}_t\|^2 \nonumber \\
    & \leq \frac{\eta_t}{8\gamma}\|\hat{x}_{t+1}-\bar{x}_t\|^2 + 2\eta_t\gamma\|\nabla_x f(\bar{x}_t,\bar{y}_t) - \nabla F(\bar{x}_t)\|^2  - \eta_t\langle \nabla F(\bar{x}_t), \hat{x}_{t+1}-\bar{x}_t \rangle + \frac{L_f\eta^2_t}{2}\|\hat{x}_{t+1}-\bar{x}_t\|^2 \nonumber \\
    & \leq \frac{\eta_t}{8\gamma}\|\hat{x}_{t+1}-\bar{x}_t\|^2 + 2L^2_f\eta_t\gamma \|\bar{y}_t - y^*(\bar{x}_t)\|^2 + F(\bar{x}_t) - F(\bar{x}_{t+1}) \nonumber \\
    &  + \frac{\eta^2_tL}{2}\|\hat{x}_{t+1}-\bar{x}_t\|^2 + \frac{\eta^2_tL_f}{2}\|\hat{x}_{t+1}-\bar{x}_t\|^2 \nonumber \\
    & \leq \frac{4L^2_f\eta_t\gamma}{\mu} \big(F(\bar{x}_t) -f(\bar{x}_t,\bar{y}_t)\big) + F(\bar{x}_t) - F(\bar{x}_{t+1}) + \eta_t(\frac{1}{8\gamma}+\eta_tL)\|\hat{x}_{t+1}-\bar{x}_t\|^2,
\end{align}
where the second last inequality is due to Lemma~\ref{lem:1}, i.e.,
$L$-smoothness of function $F(x)$, and the last inequality holds by Lemma~\ref{lem:A2} and $L_f\leq L$.
Then we have
\begin{align} \label{eq:B55}
    F(\bar{x}_{t+1}) - f(\bar{x}_{t+1},\bar{y}_t) & = F(\bar{x}_{t+1}) - F(\bar{x}_t) + F(\bar{x}_t)- f(\bar{x}_t,\bar{y}_t) + f(\bar{x}_t,\bar{y}_t) -f(\bar{x}_{t+1},\bar{y}_t) \nonumber \\
    & \leq (1+\frac{4L^2_f\eta_t\gamma}{\mu}) \big(F(\bar{x}_t) -f(\bar{x}_t,\bar{y}_t)\big) + \eta_t(\frac{1}{8\gamma}+\eta_tL)\|\hat{x}_{t+1}-\bar{x}_t\|^2.
\end{align}

Substituting \eqref{eq:B55} in \eqref{eq:B54}, we get
\begin{align}
    & F(\bar{x}_{t+1}) - f(\bar{x}_{t+1},\bar{y}_{t+1}) \nonumber \\
    & \leq (1-\frac{\eta_t\lambda\mu}{\rho_u})(1+\frac{4L^2_f\eta_t\gamma}{\mu}) \big(F(\bar{x}_t) -f(\bar{x}_t,\bar{y}_t)\big) + \eta_t(\frac{1}{8\gamma}+\eta_tL)(1-\frac{\eta_t\lambda\mu}{\rho_u})\|\hat{x}_{t+1}-\bar{x}_t\|^2 \nonumber \\
    & \quad -\frac{\eta_t}{2\lambda\rho_u} \|\hat{y}_{t+1}-\bar{y}_t\|^2 + \frac{\eta_t\lambda L^2_f }{\rho_l} \|\bar{x}_{t+1}-\bar{x}_t\|^2 + \frac{\eta_t\lambda}{\rho_l}\|\nabla_y f(\bar{x}_t,\bar{y}_t)-\bar{v}_t\|^2 + \frac{L_f\eta^2_t}{2}\|\hat{y}_{t+1}-\bar{y}_t\|^2 \nonumber \\
    & = (1-\frac{\eta_t\lambda\mu}{\rho_u})(1+\frac{4L^2_f\eta_t\gamma}{\mu}) \big(F(\bar{x}_t) -f(\bar{x}_t,\bar{y}_t)\big) + \eta_t\big(\frac{1}{8\gamma}+\eta_tL-\frac{\eta_t\lambda\mu}{8\gamma\rho_u}-\frac{\eta^2_tL\lambda\mu}{\rho_u}
    +\frac{\eta^2_tL^2_f\lambda}{\rho_l}\big)\|\hat{x}_{t+1}-\bar{x}_t\|^2 \nonumber \\
    & \quad -\frac{\eta_t}{2}\big(\frac{1}{\lambda\rho_u}-L_f\eta_t\big) \|\hat{y}_{t+1}-\bar{y}_t\|^2 + \frac{ \eta_t\lambda}{\rho_l}\|\nabla_y f(\bar{x}_t,\bar{y}_t)-\bar{v}_t\|^2 \nonumber \\
    & \leq (1-\frac{\eta_t\lambda\mu}{2\rho_u}) \big(F(\bar{x}_t) -f(\bar{x}_t,\bar{y}_t)\big) + \frac{\eta_t}{8\gamma}\|\hat{x}_{t+1}-\bar{x}_t\|^2  -\frac{\eta_t}{4\lambda\rho_u}\|\hat{y}_{t+1}-\bar{y}_t\|^2 + \frac{\eta_t\lambda}{\rho_l}\|\nabla_y f(\bar{x}_t,\bar{y}_t)-\bar{v}_t\|^2,
\end{align}
where the last inequality holds by $\gamma\leq \min\big(\frac{\lambda\mu}{16\rho_uL},\frac{\rho_l\mu}{16\rho_uL^2_f}\big)$ and $\lambda \leq \frac{1}{2\eta_t L_f\rho_u}$ for all $t\geq 1$, i.e.,
\begin{align}
   & \gamma\leq \frac{\lambda\mu}{16\rho_uL} \Rightarrow \lambda\geq \frac{16\rho_uL\gamma}{\mu}=16\rho_u\gamma(\kappa+\frac{\kappa^2}{2})\geq 8\rho_u\kappa^2\gamma \Rightarrow  \frac{\eta_t\lambda\mu}{2\rho_u} \geq \frac{4L^2_f\eta_t\gamma}{\mu} \nonumber \\
   & \gamma\leq \min\big(\frac{\lambda\mu}{16\rho_uL},\frac{\rho_l\mu}{16\rho_uL^2_f}\big) \Rightarrow \frac{\eta_t\lambda\mu}{8\gamma\rho_u}\geq \eta_tL+ \frac{\eta^2_tL^2_f\lambda}{\rho_l},  \nonumber \\
   &\lambda \leq \frac{1}{2\eta_tL_f\rho_u}  \Rightarrow \frac{1}{2\lambda\rho_u} \geq \eta_t L_f, \ \forall t\geq 1.
\end{align}

\end{proof}

\begin{lemma} \label{lem:D1}
 Assume the sequences $\{\bar{x}_t,\bar{y}_t\}_{t=1}^T$ generated from Algorithm~\ref{alg:1}. Under the Assumptions~\ref{ass:1},\ref{ass:2}, given $0< \gamma \leq \frac{\rho}{2L\eta_t}$, we have
 \begin{align}
  F(\bar{x}_{t+1})
  & \leq F(\bar{x}_t) + \frac{4\gamma L^2_f\eta_t}{\rho\mu}\big(F(\bar{x}_t)-f(\bar{x}_t,\bar{y}_t)\big) + \frac{2\gamma\eta_t}{\rho}\|\nabla_x f(\bar{x}_t,\bar{x}_t)-\bar{w}_t\|^2 -\frac{\rho\eta_t}{2\gamma}\|\hat{x}_{t+1}-\bar{x}_t\|^2.
\end{align}
\end{lemma}

\begin{proof}
For notational simplicity, let $\bar{x}_t = \frac{1}{K}\sum_{k=1}^Kx^k_t$, $\bar{y}_t = \frac{1}{K}\sum_{k=1}^Ky^k_t$, $\bar{w}_t = \frac{1}{K}\sum_{k=1}^Kw^k_t$  and $\bar{v}_t = \frac{1}{K}\sum_{k=1}^Kv^k_t$. When $t=s_t=q\lfloor t/q\rfloor +1$, according to the lines 7 and 8 of Algorithm \ref{alg:1}, we have
$\hat{y}_{t+1}=\bar{y}_t + \lambda B_t^{-1}\bar{v}_t$, $\bar{y}_{t+1} = \bar{y}_t + \eta_t(\hat{y}_{t+1}-\bar{y}_t)$,
$\hat{x}_{t+1}=\bar{x}_t - \gamma A_t^{-1}\bar{w}_t$ and $\bar{x}_{t+1} = \bar{x}_t + \eta_t(\hat{x}_{t+1}-\bar{x}_t)$.

When $t\in (s_t,s_t+q)$, according to the lines 11 and 12 of Algorithm \ref{alg:1}, we have for all $k\in [K]$, $\hat{y}^k_{t+1} = y^k_t + \lambda B_t^{-1}v^k_t$, $y^k_{t+1} = y^k_t + \eta_t(\hat{y}^k_{t+1}-y^k_t)$, $\hat{x}^k_{t+1} = x^k_t - \gamma A_t^{-1} w^k_t$ and $x^k_{t+1} = x^k_t + \eta_t(\hat{x}^k_{t+1}-x^k_t)$. Then we also have $\hat{y}_{t+1} = \frac{1}{K}\sum_{k=1}^K\hat{y}^k_{t+1} = \frac{1}{m}\sum_{k=1}^K\big(y^k_t + \lambda v^k_t\big)=\bar{y}_t + \lambda B_t^{-1}\bar{v}_t$ and $\bar{y}_{t+1}= \frac{1}{K}\sum_{k=1}^Ky^k_{t+1} = \frac{1}{K}\sum_{k=1}^K\big(y^k_t + \eta_t(\hat{y}^k_{t+1}-y^k_t)\big)= \bar{y}_t + \eta_t(\hat{y}_{t+1}-\bar{y}_t)$. Similarly, we have $\hat{x}_{t+1}=\bar{x}_t - \gamma A_t^{-1}\bar{w}_t$ and $\bar{x}_{t+1} = \bar{x}_t + \eta_t(\hat{x}_{t+1}-\bar{x}_t)$.
Thus, we have for all $t\geq1$,
\begin{align} \label{eq:D1}
 \hat{x}_{t+1}=\bar{x}_t - \gamma A_t^{-1}\bar{w}_t = \arg\min_x \Big\{ \langle \bar{w}_t, x-\bar{x}_t\rangle + \frac{1}{2\gamma}(x-\bar{x}_t)^TA_t(x-\bar{x}_t) \Big\}.
\end{align}
By using the optimal condition of the above problem~(\ref{eq:D1}), we have, for all $x\in \mathbb{R}^d$
\begin{align}
 \langle \bar{w}_t + \frac{1}{\gamma}A_t(\hat{x}_{t+1}-\bar{x}_t), x-\hat{x}_{t+1}\rangle \geq 0.
\end{align}
Let $x=x_t$, we can obtain
\begin{align} \label{eq:D2}
  \langle \bar{w}_t, \hat{x}_{t+1}-\bar{x}_t\rangle \leq -\frac{1}{\gamma}(\hat{x}_{t+1}-\bar{x}_t)^TA_t(\hat{x}_{t+1}-\bar{x}_t) \leq -\frac{\rho}{\gamma}\|\hat{x}_{t+1}-\bar{x}_t\|^2.
\end{align}

According to Lemma~\ref{lem:1}, i.e., function $F(x)$ is $L$-smooth, we have
 \begin{align} \label{eq:D3}
  F(\bar{x}_{t+1}) & \leq F(\bar{x}_t) + \langle\nabla F(\bar{x}_t), \bar{x}_{t+1}-\bar{x}_t\rangle + \frac{L}{2}\|\bar{x}_{t+1}-\bar{x}_t\|^2  \nonumber \\
  & = F(\bar{x}_t)+ \langle \nabla F(\bar{x}_t),\eta_t(\hat{x}_{t+1}-\bar{x}_t)\rangle + \frac{L}{2}\|\eta_t(\hat{x}_{t+1}-\bar{x}_t)\|^2 \nonumber \\
  & = F(\bar{x}_t) + \eta_t\langle \bar{w}_t,\hat{x}_{t+1}-\bar{x}_t\rangle + \eta_t\langle \nabla F(\bar{x}_t)-\bar{w}_t,\hat{x}_{t+1}-\bar{x}_t\rangle + \frac{L\eta_t^2}{2}\|\hat{x}_{t+1}-\bar{x}_t\|^2 \nonumber \\
  & \leq  F(\bar{x}_t) - \frac{\rho\eta_t}{\gamma}\|\hat{x}_{t+1}-\bar{x}_t\|^2 + \eta_t\langle \nabla F(\bar{x}_t)-\bar{w}_t,\hat{x}_{t+1}-\bar{x}_t\rangle + \frac{L\eta_t^2}{2}\|\hat{x}_{t+1}-\bar{x}_t\|^2,
 \end{align}
where the second equality is due to $\bar{x}_{t+1}=\bar{x}_t + \eta_t(\hat{x}_{t+1}-\bar{x}_t)$,
and the last inequality holds by the above inequality~\eqref{eq:D2}.

Meanwhile, we have
\begin{align} \label{eq:D4}
  \langle \nabla F(\bar{x}_t)-\bar{w}_t,\hat{x}_{t+1}-\bar{x}_t\rangle
  & \leq \|\nabla F(\bar{x}_t)-\bar{w}_t\|\cdot\|\hat{x}_{t+1}-\bar{x}_t\| \nonumber \\
  & \leq \frac{\gamma}{\rho}\|\nabla F(\bar{x}_t)-\bar{w}_t\|^2+\frac{\rho}{4\gamma}\|\hat{x}_{t+1}-\bar{x}_t\|^2 \nonumber \\
  & =  \frac{\gamma}{\rho}\|\nabla_x f(\bar{x}_t,y^*(\bar{x}_t))- \nabla_x f(\bar{x}_t,\bar{y}_t)+\nabla_x f(\bar{x}_t,\bar{y}_t)- \bar{w}_t\|^2+\frac{\rho}{4\gamma}\|\hat{x}_{t+1}-\bar{x}_t\|^2 \nonumber \\
  & \leq \frac{2\gamma}{\rho}\|\nabla_x f(\bar{x}_t,y^*(\bar{x}_t))- \nabla_x f(\bar{x}_t,\bar{y}_t)\|^2 + \frac{2\gamma}{\rho}\|\nabla_x f(\bar{x}_t,\bar{x}_t)-\bar{w}_t\|^2
  + \frac{\rho}{4\gamma}\|\hat{x}_{t+1}-\bar{x}_t\|^2 \nonumber \\
  & \leq \frac{2\gamma L^2_f}{\rho}\|y^*(\bar{x}_t) - \bar{y}_t\|^2 + \frac{2\gamma}{\rho}\|\nabla_x f(\bar{x}_t,\bar{x}_t)-\bar{w}_t\|^2
  + \frac{\rho}{4\gamma}\|\hat{x}_{t+1}-\bar{x}_t\|^2,
\end{align}
where the first inequality is due to the Cauchy-Schwarz inequality and the second is due to Young's inequality.
By plugging the above inequalities~\eqref{eq:D4} into \eqref{eq:D3},
we obtain
\begin{align} \label{eq:D5}
  F(\bar{x}_{t+1}) &\leq F(\bar{x}_t) + \eta_t\langle \nabla F(\bar{x}_t)-\bar{w}_t,\hat{x}_{t+1}-\bar{x}_t\rangle + \eta_t\langle \bar{w}_t,\hat{x}_{t+1}-\bar{x}_t\rangle + \frac{L\eta_t^2}{2}\|\hat{x}_{t+1}-\bar{x}_t\|^2  \nonumber \\
  & \leq F(\bar{x}_t) + \frac{2\gamma L^2_f\eta_t}{\rho}|y^*(\bar{x}_t) - \bar{y}_t\|^2 + \frac{2\gamma\eta_t}{\rho}\|\nabla_x f(\bar{x}_t,\bar{x}_t)-\bar{w}_t\|^2
  + \frac{\rho\eta_t}{4\gamma}\|\hat{x}_{t+1}-\bar{x}_t\|^2   \nonumber \\
  & \quad -\frac{\rho\eta_t}{\gamma}\|\hat{x}_{t+1}-\bar{x}_t\|^2 + \frac{L\eta_t^2}{2}\|\hat{x}_{t+1}-\bar{x}_t\|^2 \nonumber \\
  & = F(\bar{x}_t) + \frac{2\gamma L^2_f\eta_t}{\rho}\|y^*(\bar{x}_t) - \bar{y}_t\|^2 + \frac{2\gamma\eta_t}{\rho}\|\nabla_x f(\bar{x}_t,\bar{x}_t)-\bar{w}_t\|^2 -\frac{\rho\eta_t}{2\gamma}\|\hat{x}_{t+1}-\bar{x}_t\|^2  \nonumber \\
  & \quad -\big(\frac{\rho\eta_t}{4\gamma}-\frac{L\eta_t^2}{2}\big)\|\hat{x}_{t+1}-\bar{x}_t\|^2 \nonumber \\
  & \leq F(\bar{x}_t) + \frac{2\gamma L^2_f\eta_t}{\rho}\|y^*(\bar{x}_t) - \bar{y}_t\|^2 + \frac{2\gamma\eta_t}{\rho}\|\nabla_x f(\bar{x}_t,\bar{x}_t)-\bar{w}_t\|^2 -\frac{\rho\eta_t}{2\gamma}\|\hat{x}_{t+1}-\bar{x}_t\|^2 \nonumber \\
  & \leq F(\bar{x}_t) + \frac{4\gamma L^2_f\eta_t}{\rho\mu}\big(F(\bar{x}_t)-f(\bar{x}_t,\bar{y}_t)\big) + \frac{2\gamma\eta_t}{\rho}\|\nabla_x f(\bar{x}_t,\bar{x}_t)-\bar{w}_t\|^2 -\frac{\rho\eta_t}{2\gamma}\|\hat{x}_{t+1}-\bar{x}_t\|^2,
\end{align}
where the second last inequality is due to $0< \gamma \leq \frac{\rho}{2L\eta_t}$, and the last inequality holds by the above Lemma~\ref{lem:A2} using in $F(\bar{x}_t)=f(\bar{x}_t,y^*(\bar{x}_t))$ with $y^*(\bar{x}_t)\in \arg\max f(\bar{x}_t,y)$.

\end{proof}

\begin{lemma} \label{lem:E1}
Under the above assumptions, and assume the stochastic gradient estimators $\big\{\bar{v}_t,\bar{w}_t\big\}_{t=1}^T$ be generated from Algorithm \ref{alg:1},
we have
 \begin{align}\label{eq:A49}
 \mathbb{E}\|\bar{v}_{t+1} - \overline{\nabla_y f(x_{t+1},y_{t+1})}\|^2
 & \leq (1-\alpha_{t+1})\mathbb{E} \|\bar{v}_t - \overline{\nabla_y f(x_t,y_t)}\|^2 + \frac{2\alpha_{t+1}^2\sigma^2 }{K} \nonumber \\
 & \quad + \frac{4L_f^2\eta_t^2}{K^2}\sum_{k=1}^K\mathbb{E}\big(\|\hat{x}^k_{t+1} - x^k_t\|^2 + \|\hat{y}^k_{t+1} - y^k_t\|^2 \big),
 \end{align}
 \begin{align} \label{eq:A50}
\mathbb{E}\|\bar{w}_{t+1} - \overline{\nabla_x f(x_{t+1},y_{t+1})} \|^2 & \leq (1-\beta_{t+1}) \mathbb{E}\|\bar{w}_t - \overline{\nabla_x f(x_t,y_t)}\|^2 + \frac{2\beta^2_{t+1}\sigma^2}{K} \nonumber \\
& \quad + \frac{4L^2_f\eta^2_t}{K^2}\sum_{k=1}^K\mathbb{E}\big(\|\hat{x}^k_{t+1}-x^k_t\|^2 + \|\hat{y}^k_{t+1}-y^k_t\|^2 \big).
\end{align}
\end{lemma}

\begin{proof}
Without loss of generality, we only prove the above inequality \eqref{eq:A50}, and it is similar to \eqref{eq:A49}.
Since $\bar{w}_{t+1} = \frac{1}{K}\sum_{k=1}^K\Big(\nabla_x f^k(x^k_{t+1},y^k_{t+1};\xi^k_{t+1}) + (1-\beta_{t+1})\big(w^k_t
 - \nabla_x f^k(x^k_t,y^k_t;\xi^k_{t+1})\big)\Big)$,
we have
\begin{align}
 &\mathbb{E}\|\bar{w}_{t+1} - \overline{\nabla_x f(x_{t+1},y_{t+1})}\|^2  \\
 & = \mathbb{E}\big\|\frac{1}{K}\sum_{k=1}^K\big(w^k_{t+1} - \nabla_x f^k(x^k_{t+1},y^k_{t+1}) \big)\big\|^2 \nonumber \\
 & = \mathbb{E}\big\|\frac{1}{K}\sum_{k=1}^K\Big(\nabla_x f^k(x^k_{t+1},y^k_{t+1};\xi^k_{t+1}) + (1-\beta_{t+1})\big(w^k_t
 - \nabla_x f^k(x^k_t,y^k_t;\xi^k_{t+1})\big) - \nabla_x f^k(x^k_{t+1},y^k_{t+1})\Big)\big\|^2 \nonumber \\
 & = \mathbb{E}\big\|\frac{1}{K}\sum_{k=1}^K\Big(\nabla_x f^k(x^k_{t+1},y^k_{t+1};\xi^k_{t+1}) - \nabla_x f^k(x^k_{t+1},y^k_{t+1}) - (1-\beta_{t+1})\big(
 \nabla_x f^k(x^k_t,y^k_t;\xi^k_{t+1}) - \nabla_x f^k(x^k_t,y^k_t) \big)\Big) \nonumber \\
 & \quad + (1-\beta_{t+1})\frac{1}{K}\sum_{k=1}^K\big( w^k_t - \nabla_x f^k(x^k_t,y^k_t) \big) \big\|^2 \nonumber \\
 & = \frac{1}{K^2}\sum_{k=1}^K\mathbb{E}\big\|\nabla_x f^k(x^k_{t+1},y^k_{t+1};\xi^k_{t+1}) - \nabla_xf^k(x^k_{t+1},y^k_{t+1})  - (1-\beta_{t+1})\big(
 \nabla_x f^k(x^k_t,y^k_t;\xi^k_{t+1}) - \nabla_x f^k(x^k_t,y^k_t) \big) \big\|^2 \nonumber \\
 & \quad + (1-\beta_{t+1})^2\|\bar{w}_t - \overline{\nabla_x f(x_t,y_t)}\|^2 \nonumber \\
 & \leq \frac{2(1-\beta_{t+1})^2}{K^2}\sum_{k=1}^K\mathbb{E}\big\|\nabla_x f^k(x^k_{t+1},y^k_{t+1};\xi^k_{t+1}) - \nabla_x f^k(x^k_t,y^k_t;\xi^k_{t+1}) -
 \big( \nabla_x f^k(x^k_{t+1},y^k_{t+1})- \nabla_x f^k(x^k_t,y^k_t)\big) \big\|^2 \nonumber \\
 & \quad + \frac{2\beta^2_{t+1}}{K^2}\sum_{k=1}^K\mathbb{E}\big\|\nabla_x f^k(x^k_{t+1},y^k_{t+1};\xi^k_{t+1}) -
 \nabla_x f^k(x^k_{t+1},y^k_{t+1}) \big\|^2 + (1-\beta_{t+1})^2\|\bar{w}_t - \overline{\nabla_x f(x_t,y_t)}\|^2 \nonumber \\
 & \leq \frac{2(1-\beta_{t+1})^2}{K^2}\sum_{k=1}^K\mathbb{E}\big\|\nabla_x f^k(x^k_{t+1},y^k_{t+1};\xi^k_{t+1}) - \nabla_x f^k(x^k_t,y^k_t;\xi^k_{t+1}) \big\|^2 + \frac{2\beta^2_{t+1}\sigma^2}{K} \nonumber \\
 & \quad  + (1-\beta_{t+1})^2\|\bar{w}_t - \overline{\nabla_x f(x_t,y_t)}\|^2 \nonumber \\
 & \leq (1-\beta_{t+1})^2 \mathbb{E}\|\bar{w}_t - \overline{\nabla_x f(x_t,y_t)}\|^2 + \frac{2\beta^2_{t+1}\sigma^2}{K} + \frac{4(1-\beta_{t+1})^2L^2_f}{K^2}\sum_{k=1}^K\mathbb{E}\big(
  \|x^k_{t+1}-x^k_t\|^2 + \|y^k_{t+1}-y^k_t\|^2 \big) \nonumber \\
 & \leq (1-\beta_{t+1}) \mathbb{E}\|\bar{w}_t - \overline{\nabla_x f(x_t,y_t)}\|^2 + \frac{2\beta^2_{t+1}\sigma^2}{K} + \frac{4L^2_f\eta^2_t}{K^2}\sum_{k=1}^K\mathbb{E}\big(
  \|\hat{x}^k_{t+1}-x^k_t\|^2 + \|\hat{y}^k_{t+1}-y^k_t\|^2 \big) \nonumber,
\end{align}
where the forth equality holds by, for any $k\in [K]$,
\begin{align}
 \mathbb{E}_{\xi^k_{t+1}} \big[\nabla_x f(x^k_{t+1},y^k_{t+1};\xi^k_{t+1})-\nabla_x f(x_{t+1},y_{t+1})\big] = 0, \
 \mathbb{E}_{\xi^k_{t+1}} \big[\nabla_x f(x_t,y_t;\xi^k_{t+1})) - \nabla_x f(x_t,y_t) \big] = 0, \nonumber
\end{align}
and for any $k\neq j\in [K]$, $\xi^k_{t+1}$ and $\xi^j_{t+1}$ are independent, i.e.,
\begin{align}
 \Big\langle \nabla_x f^k(x^k_{t+1},y^k_{t+1};\xi^k_{t+1}) - \nabla_x f^k(x^k_{t+1},y^k_{t+1})- (1-\beta_{t+1})\big(
 \nabla_x f^k(x^k_t,y^k_t;\xi^k_{t+1}) - \nabla_x f^k(x^k_t,y^k_t) \big) \nonumber \\
 ,\nabla_x f^k(x^j_{t+1},y^j_{t+1};\xi^j_{t+1})-\hat{\nabla} f^j(x^j_{t+1},y^j_{t+1}) - (1-\beta_{t+1})\big(
 \nabla_x f^j(x^j_t,y^j_t;\xi^j_{t+1}) - \nabla_x f^j(x^j_t,y^j_t) \big)\Big\rangle =0; \nonumber
\end{align}
the second inequality holds by the inequality $\mathbb{E}\|\zeta-\mathbb{E}[\zeta]\|^2 \leq \mathbb{E}\|\zeta\|^2$ and Assumption \ref{ass:3};
the second last inequality is due to Assumption \ref{ass:1}; the last inequality holds by $0<\beta_{t+1} \leq 1$ and $x^k_{t+1}=x^k_t+\eta_t(\hat{x}^k_{t+1}-x^k_t)$,
$y^k_{t+1}=y^k_t+\eta_t(\hat{y}^k_{t+1}-y^k_t)$.

\end{proof}

\begin{lemma} \label{lem:E2}
Based on the above Assumptions \ref{ass:1} and \ref{ass:4}, we have
\begin{align}
 & \sum_{k=1}^K\mathbb{E}\big\|\nabla_xf^k(x^k_t,y^k_t)-\frac{1}{K}\sum_{j=1}^K\nabla_xf^j(x^j_t,y^j_t)\big\|^2 \leq
 12L_f^2\sum_{k=1}^K\big(\mathbb{E}\|x^k_t-\bar{x}_t\|^2 + \mathbb{E}\|y^k_t-\bar{y}_t\|^2\big) + 3K\delta_x^2,  \label{eq:A52} \\
 & \sum_{k=1}^K\mathbb{E}\big\|\nabla_y f^k(x^k_t,y^k_t)-\frac{1}{K}\sum_{j=1}^K\nabla_yf^j(x^j_t,y^j_t)\big\|^2 \leq 12L^2_f\sum_{k=1}^K\big(\mathbb{E}\|x^k_t-\bar{x}_t\|^2
 + \mathbb{E}\|y^k_t-\bar{y}_t\|^2\big) + 3K\delta_y^2. \label{eq:A53}
\end{align}

\end{lemma}

\begin{proof}
Without loss of generality, we only prove the above inequality \eqref{eq:A52}, and it is similar to \eqref{eq:A53}.
Consider the term $\sum_{k=1}^K\mathbb{E}\big\|\nabla_xf^k(x^k_t,y^k_t)-\frac{1}{K}\sum_{j=1}^K\nabla_xf^j(x^j_t,y^j_t)\big\|^2$, we have
\begin{align}
 & \sum_{k=1}^K\mathbb{E}\big\|\nabla_xf^k(x^k_t,y^k_t)-\frac{1}{K}\sum_{j=1}^K\nabla_xf^j(x^j_t,y^j_t)\big\|^2   \nonumber \\
 & = \sum_{k=1}^K\mathbb{E}\big\|\nabla_xf^k(x^k_t,y^k_t)-\nabla_xf^k(\bar{x}_t,\bar{y}_t) +\nabla_xf^k(\bar{x}_t,\bar{y}_t) - \frac{1}{K}\sum_{j=1}^K\nabla_xf^j(\bar{x}_t,\bar{y}_t) \nonumber \\
 & \quad + \frac{1}{K}\sum_{j=1}^K\nabla_xf^j(\bar{x}_t,\bar{y}_t)-\frac{1}{K}\sum_{j=1}^K\nabla_xf^j(x^j_t,y^j_t)\big\|^2   \nonumber \\
 & \leq \sum_{k=1}^K 3\mathbb{E}\big\|\nabla_xf^k(x^k_t,y^k_t)-\nabla_xf^k(\bar{x}_t,\bar{y}_t)\big\|^2 + \sum_{k=1}^K 3\mathbb{E}\big\|\nabla_xf^k(\bar{x}_t,\bar{y}_t) - \frac{1}{K}\sum_{j=1}^K\nabla_xf^j(\bar{x}_t,\bar{y}_t)\big\|^2 \nonumber \\
 & \quad + \sum_{k=1}^K 3\mathbb{E}\big\|\frac{1}{K}\sum_{j=1}^K\nabla_xf^j(\bar{x}_t,\bar{y}_t)-\frac{1}{K}\sum_{j=1}^K\nabla_xf^j(x^j_t,y^j_t)\big\|^2   \nonumber \\
 & \leq 6L_f^2\sum_{k=1}^K\big(\mathbb{E}\|x^k_t-\bar{x}_t\|^2 + \mathbb{E}\|y^k_t-\bar{y}_t\|^2\big) + 3\sum_{k=1}^K \frac{1}{K}\sum_{j=1}^K \mathbb{E}\|\nabla_xf^k(\bar{x}_t,\bar{y}_t) - \nabla_xf^j(\bar{x}_t,\bar{y}_t)\|^2 \nonumber \\
 & \quad + 3\sum_{k=1}^K\frac{1}{K}\sum_{j=1}^K\big\|\nabla_xf^j(\bar{x}_t,\bar{y}_t)-\nabla_xf^j(x^j_t,y^j_t)\big\|^2 \nonumber \\
 & \leq 12L_f^2\sum_{k=1}^K\big(\mathbb{E}\|x^k_t-\bar{x}_t\|^2 + \mathbb{E}\|y^k_t-\bar{y}_t\|^2\big) + 3K\delta_x^2,
\end{align}
where the last inequality holds by the above Assumptions \ref{ass:1} and \ref{ass:4}.

\end{proof}

\begin{lemma} \label{lem:A8}
Suppose the iterates $\{x^k_t,y^k_t\}_{t=1}^T$, for all $k \in [K]$ generated from Algorithm \ref{alg:1} satisfy:
\begin{align}
& \sum_{k=1}^K \mathbb{E}\|x^k_t- \bar{x}_t \|^2 \leq (q-1)\sum_{l = s_t+1}^{t-1} \gamma^2\eta_l^2 \sum_{k = 1}^K \mathbb{E}\|A^{-1}_l(w^k_l - \bar{w}_l)\|^2, \label{eq:A55} \\
& \sum_{k=1}^K \mathbb{E}\|y^k_t- \bar{y}_t \|^2 \leq (q-1)\sum_{l = s_t+1}^{t-1} \lambda^2\eta_l^2 \sum_{k = 1}^K \mathbb{E}\|B^{-1}_l(v^k_l - \bar{v}_l)\|^2. \label{eq:A56}
\end{align}
\end{lemma}

\begin{proof}
From Algorithm \ref{alg:1}, when $s_t=q\lfloor t/q \rfloor$, we have $t = s_t+1$ and $x^k_t = \bar{x}_t$, the above inequality holds trivially. When $t\in (s_t+1,s_t+q]$, we have
\begin{align*}
    x^k_t = x^k_{s_t+1} - \sum_{l=s_t+1}^{t-1}\gamma\eta_lA_l^{-1}w^k_l, \quad \text{and} \quad \bar{x}_{t}  = \bar{x}_{s_t+1}  - \sum_{l=s_t+1}^{t-1}\gamma\eta_l A_l^{-1}\bar{w}_l.
\end{align*}
Thus we have
\begin{align*}
  \sum_{k=1}^K\mathbb{E}\|x^k_t - \bar{x}_t\|^2 & = \sum_{k =1}^K \mathbb{E}\Big\|x^k_{s_t+1} - \bar{x}_{s_t+1}
  - \Big(\sum_{l=s_t+1}^{t-1}\gamma\eta_lA_l^{-1}w^k_l - \sum_{l=s_t+1}^{t-1} \gamma\eta_l A_l^{-1} \bar{w}_l\Big) \Big\|^2 \\
  & = \sum_{k=1}^K \mathbb{E}\Big\|\sum_{l=s_t+1}^{t-1}\big(\gamma\eta_lA_l^{-1}w^k_l - \gamma\eta_l A_l^{-1} \bar{w}_l\big) \Big\|^2
  \leq {(q -1)} \sum_{l = s_t+1}^{t-1} \gamma^2\eta_l^2 \sum_{k = 1}^K \mathbb{E}\|A^{-1}_l(w^k_l - \bar{w}_l)\|^2,
\end{align*}
where the above inequality is due to $t-s_t-1\leq q-1$.
Similarly, we can obtain the above inequality \eqref{eq:A56}.
\end{proof}

\begin{lemma} \label{lem:A9}
Let $\eta_t \leq \frac{\rho}{12\sqrt{2}\lambda qL_f}$ for all $t\geq 1$, $\gamma=\tau\lambda \ (0<\tau \leq 1)$, $\alpha_{t+1}=c_1\eta^2_t \in (0,1]$,
$\beta_{t+1}=c_2\eta^2_t\in (0,1]$ and
$c^2_1+c^2_2 \leq \frac{12^4\lambda^4q^2L^2_f}{\rho^4}$.
Let $s_t=\lfloor t/q \rfloor$ and $t\in [s_t,s_t+q-1]$, we have
\begin{align}
  & \sum_{t=s_t}^{s_t+q-1}\eta_t\sum_{k=1}^K\mathbb{E}\big( \|A^{-1}_t(w^k_t - \bar{w}_t)\|^2 + \|B^{-1}_t(v^k_t - \bar{v}_t)\|^2 \big)  \nonumber \\
  & \leq \frac{8K}{15}\sum_{t=s_t}^{s_t+q-1}\eta_t\mathbb{E}\big(\tau^2\|A^{-1}_t\bar{w}_t\|^2+\|B^{-1}_t\bar{v}_t\|^2\big)
   + \frac{2K\Delta }{15\lambda^2L^2_f}\sum_{t=s_t}^{s_t+q-1} \eta^3_t,  \nonumber
\end{align}
where $\Delta = c^2_2\sigma^2 + c^2_1\sigma^2 + 3c^2_2\delta_x^2 + 3c^2_1\delta_y^2$.
\end{lemma}

\begin{proof}
When $t=s_t=q\lfloor t/q\rfloor$, we have $w^k_{t+1} = \bar{w}_{t+1}$ for all $k\in [K]$, and then we have $\sum_{k=1}^K\mathbb{E}\|A^{-1}_{t+1}(w^k_{t+1} - \bar{w}_{t+1})\|=0$.
When $t\in (s_t,s_t+q)$, we have
\begin{align} \label{eq:A58}
  & \sum_{k=1}^K\mathbb{E}\|A^{-1}_{t+1}(w^k_{t+1} - \bar{w}_{t+1})\|^2  \nonumber \\
  & = \sum_{k=1}^K\mathbb{E}\big\|A^{-1}_{t+1}\bigg(\nabla_x f^k(x^k_{t+1},y^k_{t+1};\xi^k_{t+1}) + (1-\beta_{t+1})\big(w^k_t
 - \nabla_x f^k(x^k_t,y^k_t;\xi^k_{t+1}) \nonumber \\
  & \quad - \frac{1}{K}\sum_{k=1}^K\big(\nabla_x f^k(x^k_{t+1},y^k_{t+1};\xi^k_{t+1}) + (1-\beta_{t+1})\big(w^k_t
 - \nabla_x f^k(x^k_t,y^k_t;\xi^k_{t+1})\big)\big)\bigg)\big\|^2 \nonumber \\
  & = \sum_{k=1}^K\mathbb{E}\big\|A^{-1}_{t+1}\bigg( (1-\beta_{t+1})(w^k_t - \bar{w}_t) + \big(\nabla_x f^k(x^k_{t+1},y^k_{t+1};\xi^k_{t+1}) - \frac{1}{K}\sum_{k=1}^K\nabla_x f^k(x^k_{t+1},y^k_{t+1};\xi^k_{t+1})\big) \nonumber \\
  & \quad - (1-\beta_{t+1})\big(\nabla_x f^k(x^k_t,y^k_t;\xi^k_{t+1}) - \frac{1}{K}\sum_{k=1}^K\nabla_x f^k(x^k_t,y^k_t;\xi^k_{t+1}) \big) \bigg)\big\|^2 \nonumber \\
  & \leq (1+\nu)(1-\beta_{t+1})^2\sum_{k=1}^K\mathbb{E}\|A^{-1}_t(w^k_t - \bar{w}_t)\|^2 + (1+\frac{1}{\nu})\frac{1}{\rho^2}\sum_{k=1}^K\mathbb{E}\big\|\nabla_x f^k(x^k_{t+1},y^k_{t+1};\xi^k_{t+1})  \nonumber \\
  & \quad - \frac{1}{K}\sum_{k=1}^K\nabla_x f^k(x^k_{t+1},y^k_{t+1};\xi^k_{t+1}) - (1-\beta_{t+1})\big(\nabla_x f^k(x^k_t,y^k_t;\xi^k_{t+1}) - \frac{1}{K}\sum_{k=1}^K\nabla_x f^k(x^k_t,y^k_t;\xi^k_{t+1}) \big)\big\|^2
\end{align}
where the last inequality holds by $A_{t+1}=A_t$ for any $t\in [s_t,s_t+q-1)$ and $A_t\succeq \rho I_d$ for any $t\geq 1$.

Next, we have
\begin{align} \label{eq:A59}
  & \sum_{k=1}^K \mathbb{E}\big\|\nabla_x f^k(x^k_{t+1},y^k_{t+1};\xi^k_{t+1})
  - \frac{1}{K}\sum_{k=1}^K\nabla_x f^k(x^k_{t+1},y^k_{t+1};\xi^k_{t+1})  \nonumber \\
  & \quad - (1-\beta_{t+1})\big(\nabla_x f^k(x^k_t,y^k_t;\xi^k_{t+1}) - \frac{1}{K}\sum_{k=1}^K\nabla_x f^k(x^k_t,y^k_t;\xi^k_{t+1}) \big)\big\|^2 \nonumber \\
  & = \sum_{k=1}^K \mathbb{E}\big\| \nabla_x f^k(x^k_{t+1},y^k_{t+1};\xi^k_{t+1})-\nabla_x f^k(x^k_t,y^k_t;\xi^k_{t+1})
  - \frac{1}{K}\sum_{k=1}^K\big(\nabla_x f^k(x^k_{t+1},y^k_{t+1};\xi^k_{t+1})-\nabla_x f^k(x^k_t,y^k_t;\xi^k_{t+1})\big)  \nonumber \\
  & \quad +\beta_{t+1}\big(\nabla_x f^k(x^k_t,y^k_t;\xi^k_{t+1}) - \frac{1}{K}\sum_{k=1}^K\nabla_x f^k(x^k_t,y^k_t;\xi^k_{t+1}) \big)\big\|^2 \nonumber \\
  & \leq 2\sum_{k=1}^K\mathbb{E}\big\|\nabla_x f^k(x^k_{t+1},y^k_{t+1};\xi^k_{t+1})-\nabla_x f^k(x^k_t,y^k_t;\xi^k_{t+1})\|^2 \nonumber \\
  & \quad + 2\beta^2_{t+1}\sum_{k=1}^K \mathbb{E}\big\|\nabla_x f^k(x^k_t,y^k_t;\xi^k_{t+1}) - \frac{1}{K}\sum_{k=1}^K\nabla_x f^k(x^k_t,y^k_t;\xi^k_{t+1}) \big\|^2 \nonumber \\
  & \leq 4L^2_f\sum_{k=1}^K\mathbb{E}\big(\|x^k_{t+1}-x^k_t\|^2+\|y^k_{t+1}-y^k_t\|^2\big)+ 2\beta^2_{t+1}\sum_{k=1}^K\mathbb{E}\big\|\nabla_x f^k(x^k_t,y^k_t;\xi^k_{t+1}) - \frac{1}{K}\sum_{k=1}^K\nabla_x f^k(x^k_t,y^k_t;\xi^k_{t+1}) \big\|^2,
\end{align}
where the second last inequality is due to Young inequality and the above Lemma \ref{lem:A5}, and
the last inequality holds by Assumption \ref{ass:2}.

Consider the term $\sum_{k=1}^K\big\|\nabla_x f^k(x^k_t,y^k_t;\xi^k_{t+1}) - \frac{1}{K}\sum_{k=1}^K\nabla_x f^k(x^k_t,y^k_t;\xi^k_{t+1}) \big\|$, we have
\begin{align} \label{eq:A60}
 & \sum_{k=1}^K \big\|\nabla_x f^k(x^k_t,y^k_t;\xi^k_{t+1}) - \frac{1}{K}\sum_{k=1}^K\nabla_x f^k(x^k_t,y^k_t;\xi^k_{t+1}) \big\|^2 \nonumber \\
 & = \sum_{k=1}^K \big\|\nabla_x f^k(x^k_t,y^k_t;\xi^k_{t+1}) - \nabla_x f^k(x^k_t,y^k_t) - \frac{1}{K}\sum_{k=1}^K\big(\nabla_x f^k(x^k_t,y^k_t;\xi^k_{t+1}) - \nabla_x f^k(x^k_t,y^k_t)\big) \nonumber \\
 & \quad \quad + \nabla_x f^k(x^k_t,y^k_t) - \frac{1}{K}\sum_{k=1}^K\nabla_x f^k(x^k_t,y^k_t) \big\|^2 \nonumber \\
 & \leq 2\sum_{k=1}^K\big\|\nabla_x f^k(x^k_t,y^k_t;\xi^k_{t+1}) - \nabla_x f^k(x^k_t,y^k_t) - \frac{1}{K}\sum_{k=1}^K\big(\nabla_x f^k(x^k_t,y^k_t;\xi^k_{t+1}) - \nabla_x f^k(x^k_t,y^k_t)\big) \big\| \nonumber \\
 & \quad \quad + 2\sum_{k=1}^K\big\|\nabla_x f^k(x^k_t,y^k_t) - \frac{1}{K}\sum_{k=1}^K\nabla_x f^k(x^k_t,y^k_t) \big\|^2 \nonumber \\
 & \leq 2\sum_{k=1}^K\big\|\nabla_x f^k(x^k_t,y^k_t;\xi^k_{t+1}) - \nabla_x f^k(x^k_t,y^k_t)\big\| + 2\sum_{k=1}^K\big\|\nabla_x f^k(x^k_t,y^k_t) - \frac{1}{K}\sum_{k=1}^K\nabla_x f^k(x^k_t,y^k_t) \big\|^2 \nonumber \\
 & \leq 2K\sigma^2 + 24L_f^2\sum_{k=1}^K\big(\mathbb{E}\|x^k_t-\bar{x}_t\|^2 + \mathbb{E}\|y^k_t-\bar{y}_t\|^2\big) + 6K\delta_x^2,
\end{align}
where the last inequality holds by Assumption \ref{ass:1} and the above Lemma \ref{lem:E2}.

By combining the above inequalities \eqref{eq:A58}, \eqref{eq:A59} and \eqref{eq:A60}, we have
\begin{align} \label{eq:A61}
 & \sum_{k=1}^K\mathbb{E}\|A^{-1}_{t+1}(w^k_{t+1} - \bar{w}_{t+1})\|^2  \\
 & \leq (1+\nu)(1-\beta_{t+1})^2\sum_{k=1}^K\mathbb{E}\|A^{-1}_t(w^k_t - \bar{w}_t)\|^2 + (1+\frac{1}{\nu})\frac{1}{\rho^2}
 \bigg(4L^2_f\sum_{k=1}^K\mathbb{E}\big(\|x^k_{t+1}-x^k_t\|^2+\|y^k_{t+1}-y^k_t\|^2\big)  \nonumber \\
 & \quad + 4\beta^2_{t+1}K\sigma^2 + 48\beta^2_{t+1}L_f^2\sum_{k=1}^K\big(\mathbb{E}\|x^k_t-\bar{x}_t\|^2
  + \mathbb{E}\|y^k_t-\bar{y}_t\|^2\big) + 12\beta^2_{t+1}K\delta_x^2 \bigg)  \nonumber \\
 & \leq (1+\nu)(1-\beta_{t+1})^2\sum_{k=1}^K\mathbb{E}\|A^{-1}_t(w^k_t - \bar{w}_t)\|^2 + (1+\frac{1}{\nu})\frac{1}{\rho^2}
 \bigg(4L^2_f\sum_{k=1}^K\mathbb{E}\big(\gamma^2\eta^2_t\|A^{-1}_tw^k_t\|^2+\lambda^2\eta^2_t\|B^{-1}_tv^k_t\|^2\big)  \nonumber \\
 & \quad + 4\beta^2_{t+1}K\sigma^2 + 48\beta^2_{t+1}L_f^2\Big((q-1)\sum_{l = s_t+1}^{t-1} \gamma^2\eta_l^2 \sum_{k = 1}^K \mathbb{E}\|A^{-1}_l(w^k_l - \bar{w}_l)\|^2 \nonumber \\
 & \quad + (q-1)\sum_{l = s_t+1}^{t-1} \lambda^2\eta_l^2 \sum_{k = 1}^K \mathbb{E}\|B^{-1}_l(v^k_l - \bar{v}_l)\|^2\Big) + 12\beta^2_{t+1}K\delta_x^2 \bigg)  \nonumber \\
 & \leq (1+\nu)(1-\beta_{t+1})^2\sum_{k=1}^K\mathbb{E}\|A^{-1}_t(w^k_t - \bar{w}_t)\|^2 + (1+\frac{1}{\nu})\frac{1}{\rho^2}
 \bigg(8L^2_f\sum_{k=1}^K\mathbb{E}\big(\gamma^2\eta^2_t\|A^{-1}_t(w^k_t-\bar{w}_t)\|^2+\lambda^2\eta^2_t\|B^{-1}_t(v^k_t-\bar{v}_t)\|^2\big)  \nonumber \\
 & \quad + 8L^2_f\sum_{k=1}^K\mathbb{E}\big(\gamma^2\eta^2_t\|A^{-1}_t\bar{w}_t\|^2+\lambda^2\eta^2_t\|B^{-1}_t\bar{v}_t\|^2\big)
 + 4\beta^2_{t+1}K\sigma^2 \nonumber \\
 & \quad + 48\beta^2_{t+1}L_f^2\big((q-1)\sum_{l = s_t+1}^{t-1} \gamma^2\eta_l^2 \sum_{k = 1}^K \mathbb{E}\|A^{-1}_l(w^k_l - \bar{w}_l)\|^2
 + (q-1)\sum_{l = s_t+1}^{t-1} \lambda^2\eta_l^2 \sum_{k = 1}^K \mathbb{E}\|B^{-1}_l(v^k_l - \bar{v}_l)\|^2\big) + 12\beta^2_{t+1}K\delta_x^2 \bigg),  \nonumber
\end{align}
where the second inequality holds by the above Lemma \ref{lem:A8}.

Similarly, we can also obtain
\begin{align} \label{eq:A62}
 & \sum_{k=1}^K\mathbb{E}\|B^{-1}_{t+1}(v^k_{t+1} - \bar{v}_{t+1})\|^2  \\
 & \leq (1+\nu)(1-\alpha_{t+1})^2\sum_{k=1}^K\mathbb{E}\|B^{-1}_t(v^k_t - \bar{v}_t)\|^2 + (1+\frac{1}{\nu})\frac{1}{\rho^2}
 \bigg(4L^2_f\sum_{k=1}^K\mathbb{E}\big(\|x^k_{t+1}-x^k_t\|^2+\|y^k_{t+1}-y^k_t\|^2\big)  \nonumber \\
 & \quad + 4K\alpha^2_{t+1}\sigma^2 + 48\alpha^2_{t+1}L^2_f\sum_{k=1}^K\big(\mathbb{E}\|x^k_t-\bar{x}_t\|^2
  + \mathbb{E}\|y^k_t-\bar{y}_t\|^2\big) + 12\alpha^2_{t+1}K\delta_y^2 \bigg) \nonumber \\
 & \leq (1+\nu)(1-\alpha_{t+1})^2\sum_{k=1}^K\mathbb{E}\|B^{-1}_t(v^k_t - \bar{v}_t)\|^2 + (1+\frac{1}{\nu})\frac{1}{\rho^2}
 \bigg(8L^2_f\sum_{k=1}^K\mathbb{E}\big(\gamma^2\eta^2_t\|A^{-1}_t(w^k_t-\bar{w}_t)\|^2+\lambda^2\eta^2_t\|B^{-1}_t(v^k_t-\bar{v}_t)\|^2\big)  \nonumber \\
 & \quad + 8L^2_f\sum_{k=1}^K\mathbb{E}\big(\gamma^2\eta^2_t\|A^{-1}_t\bar{w}_t\|^2+\lambda^2\eta^2_t\|B^{-1}_t\bar{v}_t\|^2\big) + 4K\alpha^2_{t+1}\sigma^2 \nonumber\\
 & \quad + 48\alpha^2_{t+1}L^2_f\big( (q-1)\sum_{l = s_t+1}^{t-1} \gamma^2\eta_l^2 \sum_{k = 1}^K \mathbb{E}\|A^{-1}_l(w^k_l - \bar{w}_l)\|^2
 + (q-1)\sum_{l = s_t+1}^{t-1} \lambda^2\eta_l^2 \sum_{k = 1}^K \mathbb{E}\|B^{-1}_l(v^k_l - \bar{v}_l)\|^2 \big) + 12\alpha^2_{t+1}K\delta_y^2 \bigg). \nonumber
\end{align}

By combining the above inequalities \eqref{eq:A61} with \eqref{eq:A62}, we have
\begin{align} \label{eq:A63}
 & \sum_{k=1}^K\mathbb{E}\big( \|A^{-1}_{t+1}(w^k_{t+1} - \bar{w}_{t+1})\|^2 + \|B^{-1}_{t+1}(v^k_{t+1} - \bar{v}_{t+1})\|^2 \big) \nonumber \\
 & \leq (1+\nu)(1-\beta_{t+1})^2\sum_{k=1}^K\mathbb{E}\|A^{-1}_t(w^k_t - \bar{w}_t)\|^2 + (1+\frac{1}{\nu})\frac{1}{\rho^2}
 \bigg(8L^2_f\sum_{k=1}^K\mathbb{E}\big(\gamma^2\eta^2_t\|A^{-1}_t(w^k_t-\bar{w}_t)\|^2+\lambda^2\eta^2_t\|B^{-1}_t(v^k_t-\bar{v}_t)\|^2\big)  \nonumber \\
 & \quad + 8L^2_f\sum_{k=1}^K\mathbb{E}\big(\gamma^2\eta^2_t\|A^{-1}_t\bar{w}_t\|^2+\lambda^2\eta^2_t\|B^{-1}_t\bar{v}_t\|^2\big)
 + 4\beta^2_{t+1}K\sigma^2 \nonumber \\
 & \quad + 48\beta^2_{t+1}L_f^2\big((q-1)\sum_{l = s_t+1}^{t-1} \gamma^2\eta_l^2 \sum_{k = 1}^K \mathbb{E}\|A^{-1}_l(w^k_l - \bar{w}_l)\|^2
 + (q-1)\sum_{l = s_t+1}^{t-1} \lambda^2\eta_l^2 \sum_{k = 1}^K \mathbb{E}\|B^{-1}_l(v^k_l - \bar{v}_l)\|^2\big) + 12\beta^2_{t+1}K\delta_x^2 \bigg)  \nonumber \\
 & \quad + (1+\nu)(1-\alpha_{t+1})^2\sum_{k=1}^K\mathbb{E}\|B^{-1}_t(v^k_t - \bar{v}_t)\|^2 + (1+\frac{1}{\nu})\frac{1}{\rho^2}
 \bigg(8L^2_f\sum_{k=1}^K\mathbb{E}\big(\gamma^2\eta^2_t\|A^{-1}_t(w^k_t-\bar{w}_t)\|^2+\lambda^2\eta^2_t\|B^{-1}_t(v^k_t-\bar{v}_t)\|^2\big)  \nonumber \\
 & \quad + 8L^2_f\sum_{k=1}^K\mathbb{E}\big(\gamma^2\eta^2_t\|A^{-1}_t\bar{w}_t\|^2+\lambda^2\eta^2_t\|B^{-1}_t\bar{v}_t\|^2\big) + 4K\alpha^2_{t+1}\sigma^2 \nonumber\\
 & \quad + 48\alpha^2_{t+1}L^2_f\big( (q-1)\sum_{l = s_t+1}^{t-1} \gamma^2\eta_l^2 \sum_{k = 1}^K \mathbb{E}\|A^{-1}_l(w^k_l - \bar{w}_l)\|^2
 + (q-1)\sum_{l = s_t+1}^{t-1} \lambda^2\eta_l^2 \sum_{k = 1}^K \mathbb{E}\|B^{-1}_l(v^k_l - \bar{v}_l)\|^2 \big) + 12\alpha^2_{t+1}K\delta_y^2 \bigg) \nonumber \\
 & \leq \max\bigg((1+\nu)(1-\beta_{t+1})^2 + 16\gamma^2\eta^2_t(1+\frac{1}{\nu})\frac{1}{\rho^2}L^2_f, (1+\nu)(1-\alpha_{t+1})^2 + 16\lambda^2\eta^2_t(1+\frac{1}{\nu})\frac{1}{\rho^2}L^2_f \bigg) \nonumber \\
 & \quad \cdot\sum_{k=1}^K\mathbb{E}\big( \|A^{-1}_t(w^k_t - \bar{w}_t)\|^2 + \|B^{-1}_t(v^k_t - \bar{v}_t)\|^2 \big) \nonumber \\
 & \quad + 16\eta^2_tL^2_f(1+\frac{1}{\nu})\frac{1}{\rho^2}\sum_{k=1}^K\mathbb{E}\big(\gamma^2\|A^{-1}_t\bar{w}_t\|^2+\lambda^2\|B^{-1}_t\bar{v}_t\|^2\big) \nonumber \\
 & \quad + (1+\frac{1}{\nu})\frac{1}{\rho^2}\bigg(4K\beta^2_{t+1}\sigma^2
 + 4K\alpha^2_{t+1}\sigma^2 + 12\beta^2_{t+1}K\delta_x^2 + 12\alpha^2_{t+1}K\delta_y^2 \bigg) \nonumber \\
 & \quad + 48(q-1)(1+\frac{1}{\nu})\frac{1}{\rho^2}(\beta^2_{t+1}+\alpha^2_{t+1})L^2_f\max(\gamma^2,\lambda^2)\sum_{l = s_t+1}^{t-1}\eta_l^2 \sum_{k = 1}^K
 \Big( \mathbb{E}\|A^{-1}_l(w^k_l - \bar{w}_l)\|^2 + \mathbb{E}\|B^{-1}_l(v^k_l - \bar{v}_l)\|^2 \Big).
\end{align}

Let $\gamma = \tau\lambda \ (0<\tau \leq 1)$, $\nu=\frac{1}{q}$ and $\eta_t \leq \frac{\rho}{12\sqrt{2}\lambda qL_f}$ for all $t\geq 1$.
Since $\alpha_{t+1}\in (0,1)$ and $\beta_{t+1}\in (0,1)$ for all $t\geq 0$, we have
\begin{align}
 & (1+\nu)(1-\beta_{t+1})^2 + 16\gamma^2\eta^2_t(1+\frac{1}{\nu})\frac{1}{\rho^2}L^2_f \nonumber \\
 & \leq 1+ \frac{1}{q} + 16(1+q)\frac{\gamma^2}{\rho^2}L^2_f\frac{\rho^2}{288\lambda^2q^2L^2_f} \nonumber \\
 & \leq 1+ \frac{1}{q} + \frac{\gamma^2}{\lambda^2}\frac{1+q}{18q^2} \leq 1+ \frac{10}{9q}.
\end{align}
Similarly, we can also obtain $(1+\nu)(1-\alpha_{t+1})^2+16\lambda^2\eta^2_t(1+\frac{1}{\nu})\frac{1}{\rho^2}L^2_f \leq 1+ \frac{10}{9q}$.
Thus, we have
\begin{align} \label{eq:A64}
 & \sum_{k=1}^K\mathbb{E}\big( \|A^{-1}_{t+1}(w^k_{t+1} - \bar{w}_{t+1})\|^2 + \|B^{-1}_{t+1}(v^k_{t+1} - \bar{v}_{t+1})\|^2 \big) \nonumber  \\
 & \leq \big(1+ \frac{10}{9q} \big)\sum_{k=1}^K\mathbb{E}\big( \|A^{-1}_t(w^k_t - \bar{w}_t)\|^2 + \|B^{-1}_t(v^k_t - \bar{v}_t)\|^2 \big) \nonumber \\
 & \quad + 16\eta^2_tL^2_f(1+q)\frac{\lambda^2}{\rho^2}\sum_{k=1}^K\mathbb{E}\big(\tau^2\|A^{-1}_t\bar{w}_t\|^2+\|B^{-1}_t\bar{v}_t\|^2\big) \nonumber \\
 & \quad + (1+q)\frac{1}{\rho^2}\big(4K\beta^2_{t+1}\sigma^2 + 4K\alpha^2_{t+1}\sigma^2 + 12\beta^2_{t+1}K\delta_x^2 + 12\alpha^2_{t+1}K\delta_y^2 \big) \nonumber \\
 & \quad + 48q^2\frac{\lambda^2}{\rho^2}(\beta^2_{t+1}+\alpha^2_{t+1})L^2_f\sum_{l = s_t+1}^{t-1}\eta_l^2 \sum_{k = 1}^K
 \big( \mathbb{E}\|A^{-1}_l(w^k_l - \bar{w}_l)\|^2 + \mathbb{E}\|B^{-1}_l(v^k_l - \bar{v}_l)\|^2 \big) \nonumber \\
 & \leq \big(1+ \frac{10}{9q} \big)\sum_{k=1}^K\mathbb{E}\big( \|A^{-1}_t(w^k_t - \bar{w}_t)\|^2 + \|B^{-1}_t(v^k_t - \bar{v}_t)\|^2 \big) + \frac{1}{9q}\sum_{k=1}^K\mathbb{E}\big(\tau^2\|A^{-1}_t\bar{w}_t\|^2+\|B^{-1}_t\bar{v}_t\|^2\big) \nonumber \\
 & \quad + \frac{\sqrt{2}K\eta^3_t}{3\rho\lambda L_f}\big( c^2_2\sigma^2 + c^2_1\sigma^2 + 3c^2_2\delta_x^2 + 3c^2_1\delta_y^2 \big) \nonumber \\
 & \quad + (c^2_2 + c^2_1)\frac{\eta^2_tL^2_f}{6}\sum_{l = s_t+1}^{t-1}\eta_l^2 \sum_{k = 1}^K
 \big( \mathbb{E}\|A^{-1}_l(w^k_l - \bar{w}_l)\|^2 + \mathbb{E}\|B^{-1}_l(v^k_l - \bar{v}_l)\|^2 \big),
\end{align}
where the first inequality is due to $\alpha_{t+1}=c_1\eta^2_t$ and $\beta_{t+1}=c_2\eta^2_t$, and
the last inequality holds by $16\lambda^2\eta^2_t(1+\frac{1}{\nu})\frac{1}{\rho^2}L^2_f \leq \frac{1}{9q}$.

When $t=s_t=q\lfloor t/q\rfloor$, we have $w^k_{t+1} = \bar{w}_{t+1}$ and $v^k_{t+1} = \bar{v}_{t+1}$ for all $k\in [K]$,
and then we have $\sum_{k=1}^K\mathbb{E}\|A^{-1}_{t+1}(w^k_{t+1} - \bar{w}_{t+1})\|=0$
and $\sum_{k=1}^K\mathbb{E}\|B^{-1}_{t+1}(v^k_{t+1} - \bar{v}_{t+1})\|=0$.
When $t\in (s_t,s_t+q)$, we have
\begin{align} \label{eq:A65}
 & \sum_{k=1}^K\mathbb{E}\big( \|A^{-1}_{t+1}(w^k_{t+1} - \bar{w}_{t+1})\|^2 + \|B^{-1}_{t+1}(v^k_{t+1} - \bar{v}_{t+1})\|^2 \big) \nonumber \\
 & \leq \frac{1}{9q}\sum_{s=s_t}^{t}\big(1+ \frac{10}{9q} \big)^{t-s_t}\sum_{k=1}^K\mathbb{E}\big(\tau^2\|A^{-1}_s\bar{w}_s\|^2+\|B^{-1}_s\bar{v}_s\|^2\big) \nonumber \\
 & \quad + \frac{\sqrt{2}K}{3\rho\lambda L_f}\big( c^2_2\sigma^2 + c^2_1\sigma^2 + 3c^2_2\delta_x^2 + 3c^2_1\delta_y^2 \big)
 \sum_{s=s_t}^t\big(1+ \frac{10}{9q} \big)^{t-s_t} \eta^3_s\nonumber \\
 & \quad + \frac{(c^2_2 + c^2_1)L^2_f}{6}\sum_{s=s_t}^t\big(1+ \frac{10}{9q} \big)^{t-s_t}\eta^2_s\sum_{l=s_t}^{s}\eta_l^2 \sum_{k = 1}^K
 \big( \mathbb{E}\|A^{-1}_l(w^k_l - \bar{w}_l)\|^2 + \mathbb{E}\|B^{-1}_l(v^k_l - \bar{v}_l)\|^2 \big) \nonumber \\
 & \leq \frac{1}{9q}\sum_{s=s_t}^{t}\big(1+ \frac{10}{9q} \big)^{q}\sum_{k=1}^K\mathbb{E}\big(\tau^2\|A^{-1}_s\bar{w}_s\|^2+\|B^{-1}_s\bar{v}_s\|^2\big) \nonumber \\
 & \quad + \frac{\sqrt{2}K}{3\rho\lambda L_f}\big( c^2_2\sigma^2 + c^2_1\sigma^2 + 3c^2_2\delta_x^2 + 3c^2_1\delta_y^2 \big)
 \sum_{s=s_t}^t\big(1+ \frac{10}{9q} \big)^{q} \eta^3_s\nonumber \\
 & \quad + \frac{(c^2_2 + c^2_1)L^2_f}{6}\sum_{s=s_t}^t\big(1+ \frac{10}{9q} \big)^{q}\eta^2_s\sum_{l=s_t}^{s}\eta_l^2 \sum_{k = 1}^K
 \big( \mathbb{E}\|A^{-1}_l(w^k_l - \bar{w}_l)\|^2 + \mathbb{E}\|B^{-1}_l(v^k_l - \bar{v}_l)\|^2 \big) \nonumber \\
 & \leq \frac{4K}{9q}\sum_{s=s_t}^{t+1} \mathbb{E}\big(\tau^2\|A^{-1}_s\bar{w}_s\|^2+\|B^{-1}_s\bar{v}_s\|^2\big) \nonumber \\
 & \quad + \frac{4\sqrt{2}K}{3\rho\lambda L_f}\big( c^2_2\sigma^2 + c^2_1\sigma^2 + 3c^2_2\delta_x^2 + 3c^2_1\delta_y^2 \big)
 \sum_{s=s_t}^{t+1} \eta^3_s\nonumber \\
 & \quad + \frac{\rho^3(c^2_2 + c^2_1)}{36*(12)^2\sqrt{2}\lambda^3q^2L_f}\sum_{s=s_t}^{t+1} \eta_s\sum_{k = 1}^K
 \big( \mathbb{E}\|A^{-1}_s(w^k_s - \bar{w}_s)\|^2 + \mathbb{E}\|B^{-1}_s(v^k_s - \bar{v}_s)\|^2 \big),
\end{align}
where the last inequality holds by $\big(1+ \frac{10}{9q} \big)^{q} \leq e^{10/9}\leq 4$.

By multiplying both sides of \eqref{eq:A65} by $\eta_{t+1}$ and summing over $t=s_t-1$ to $s_t+q-2$, we have
\begin{align} \label{eq:A66}
 & \sum_{t=s_t}^{s_t+q-1}\eta_t\sum_{k=1}^K\mathbb{E}\big( \|A^{-1}_t(w^k_t - \bar{w}_t)\|^2 + \|B^{-1}_t(v^k_t - \bar{v}_t)\|^2 \big) \nonumber  \\
 & \leq \frac{4K}{9}\sum_{t=s_t}^{s_t+q-1}\eta_t\mathbb{E}\big(\tau^2\|A^{-1}_t\bar{w}_t\|^2+\|B^{-1}_t\bar{v}_t\|^2\big) \nonumber \\
 & \quad + \frac{K}{9\lambda^2L^2_f}\big( c^2_2\sigma^2 + c^2_1\sigma^2 + 3c^2_2\delta_x^2 + 3c^2_1\delta_y^2 \big)
 \sum_{t=s_t}^{s_t+q-1} \eta^3_t\nonumber \\
 & \quad + \frac{\rho^4(c^2_2 + c^2_1)}{72*(12)^3\lambda^4q^2L^2_f}\sum_{t=s_t}^{s_t+q-1} \eta_t\sum_{k = 1}^K
 \big( \mathbb{E}\|A^{-1}_t(w^k_t - \bar{w}_t)\|^2 + \mathbb{E}\|B^{-1}_t(v^k_t - \bar{v}_t)\|^2 \big),
\end{align}

Let $c^2_1+c^2_2 \leq \frac{12^4\lambda^4q^2L^2_f}{\rho^4}$, we have
$\frac{60}{72}\leq 1-\frac{\rho^4(c^2_2 + c^2_1)}{72*(12)^3\lambda^4q^2L^2_f}$,
we have
\begin{align}
 & \sum_{t=s_t}^{s_t+q-1}\eta_t\sum_{k=1}^K\mathbb{E}\big( \|A^{-1}_t(w^k_t - \bar{w}_t)\|^2 + \|B^{-1}_t(v^k_t - \bar{v}_t)\|^2 \big) \nonumber  \\
 & \leq \frac{8K}{15}\sum_{t=s_t}^{s_t+q-1}\eta_t\mathbb{E}\big(\tau^2\|A^{-1}_t\bar{w}_t\|^2+\|B^{-1}_t\bar{v}_t\|^2\big) \nonumber \\
 & \quad + \frac{2K}{15\lambda^2L^2_f}\big( c^2_2\sigma^2 + c^2_1\sigma^2 + 3c^2_2\delta_x^2 + 3c^2_1\delta_y^2 \big)
 \sum_{t=s_t}^{s_t+q-1} \eta^3_t.
\end{align}

\end{proof}

\begin{theorem}  \label{th:A1}
(Restatement of Theorem \ref{th:1})
Assume the sequence $\{\bar{x}_t,\bar{y}_t\}_{t=1}^T$ be generated from Algorithm \ref{alg:1}.
 Under the above Assumptions~\ref{ass:1},\ref{ass:2}-\ref{ass:6}, and let $\eta_t=\frac{nK^{1/3}}{(m+t)^{1/3}}$ for all $t\geq 0$, $\alpha_{t+1}=c_1\eta_t^2$, $\beta_{t+1}=c_2\eta_t^2$, $m \geq \max\Big(2,n^3, (c_1n)^3K, (c_2n)^3K, \frac{K\big(12\sqrt{2}n\lambda qL_f\big)^3}{\rho^3}\Big)$, $n>0$, $c^2_1+c^2_2 \leq \frac{12^4\lambda^4q^2L^2_f}{\rho^4}$,
 $ c_1 \geq \frac{2}{3n^3K} + \frac{9\rho_uL_f^2}{2\mu^2\rho}$, $c_2 \geq \frac{2}{3n^3K} + \frac{9}{2}$, $\gamma=\tau\lambda$,
 $\tau \leq \min\big(\frac{\sqrt{5K}}{4\sqrt{2\Lambda}},1\big)$, $\gamma \leq  \min\big( \frac{m^{1/3}\rho}{4Ln}, \frac{\lambda\mu}{16\rho_uL},\frac{\rho_l\mu}{16\rho_uL^2_f}, \frac{2\lambda\mu^2\rho}{27L^2_f\rho_u}, \frac{\sqrt{K}\rho}{8\sqrt{3}L_f}\big)$,  $\lambda \leq \min\big( \frac{m^{1/3}}{4L_fn\rho_u},\frac{3\sqrt{5K}}{32\sqrt{2}\mu}\big)$, $0<\rho\leq 1$ and $0<\rho_u\leq \frac{135}{64\rho^2}$, we have
\begin{align}
 \frac{1}{T}\sum_{t=1}^T\mathbb{E}\|\nabla F(\bar{x}_t)\| \leq \sqrt{\frac{1}{T}\sum_{t=1}^T\mathbb{E}\|A_t\|^2}\Big( \frac{\sqrt{3G}m^{1/6}}{K^{1/6}T^{1/2}} +
 \frac{\sqrt{3G}}{K^{1/6}T^{1/3}}\Big),
\end{align}
where $G = \frac{4(F(\bar{x}_1) - F^*)}{\rho\gamma n} + \frac{36\rho_u L^2_f}{\rho\lambda\mu^2n}\big(F(\bar{x}_1) -f(\bar{x}_1,\bar{y}_1) + \frac{8m^{1/3}\sigma^2}{qK^{4/3}n^2\rho} +  8Kn^2\Big( \frac{(c_1^2+c_2^2)\sigma^2}{\rho^2K} + \frac{\Lambda\Delta }{15K\lambda^2L^2_f}\Big)\ln(m+t)$,
 $\Delta = c^2_2\sigma^2 + c^2_1\sigma^2 + 3c^2_2\delta_x^2 + 3c^2_1\delta_y^2$ and $\Lambda= \frac{1}{16} + \frac{L^2_f\rho_u}{4\mu^2}+ \frac{16\lambda^2L^2_f}{K\rho^2}$.
\end{theorem}

\begin{proof}
According to Lemma~\ref{lem:D1}, we have
\begin{align} \label{eq:W1}
  F(\bar{x}_{t+1})
  & \leq F(\bar{x}_t) + \frac{4\gamma L^2_f\eta_t}{\rho\mu}\big(F(\bar{x}_t)-f(\bar{x}_t,\bar{y}_t)\big) + \frac{2\gamma\eta_t}{\rho}\|\nabla_x f(\bar{x}_t,\bar{x}_t)-\bar{w}_t\|^2 -\frac{\rho\eta_t}{2\gamma}\|\hat{x}_{t+1}-\bar{x}_t\|^2.
\end{align}

Since $\nabla_x f(\bar{x}_t,\bar{y}_t)=\frac{1}{K}\sum_{k=1}^K\nabla_x f^k(\bar{x}_t,\bar{y}_t)$
and $\overline{\nabla_x f(x_t,y_t)}=\frac{1}{K}\sum_{k=1}^K\nabla_x f^k(x^k_t,y^k_t)$, we have
\begin{align} \label{eq:W2}
& \|\bar{w}_t-\nabla_x f(\bar{x}_t,\bar{x}_t)\|^2 \nonumber \\
& = \|\bar{w}_t-\overline{\nabla_x f(x_t,y_t)}+\overline{\nabla_x f(x_t,y_t)}-\nabla_x f(\bar{x}_t,\bar{x}_t)\|^2 \nonumber \\
& \leq 2\|\bar{w}_t-\overline{\nabla_x f(x_t,y_t)}\|^2 + 2\|\overline{\nabla_x f(x_t,y_t)}-\nabla_x f(\bar{x}_t,\bar{x}_t)\|^2 \nonumber \\
& \leq 2\|\bar{w}_t-\overline{\nabla_x f(x_t,y_t)}\|^2 + 2\|\frac{1}{K}\sum_{k=1}^K\nabla_x f^k(x^k_t,y^k_t)-\frac{1}{K}\sum_{k=1}^K\nabla_x f^k(\bar{x}_t,\bar{y}_t)\|^2 \nonumber \\
& \leq 2\|\bar{w}_t-\overline{\nabla_x f(x_t,y_t)}\|^2 + \frac{4L^2_f}{K}\sum_{k=1}^K \big(\|x^k_t-\bar{x}_t\|^2 +\|y^k_t-\bar{y}_t)\|^2\big) \nonumber \\
& \leq 2\|\bar{w}_t-\overline{\nabla_x f(x_t,y_t)}\|^2 + \frac{4(q-1)L^2_f}{K}\Big(\sum_{l = s_t+1}^{t-1} \gamma^2\eta_l^2 \sum_{k = 1}^K \mathbb{E}\|A^{-1}_l(w^k_l - \bar{w}_l)\|^2 + \sum_{l = s_t+1}^{t-1} \lambda^2\eta_l^2 \sum_{k = 1}^K \mathbb{E}\|B^{-1}_l(v^k_l - \bar{v}_l)\|^2\Big).
\end{align}

By combining the above inequalities~(\ref{eq:W1}) with~(\ref{eq:W2}), we have
\begin{align} \label{eq:W3}
  F(\bar{x}_{t+1})
  & \leq F(\bar{x}_t) + \frac{4\gamma L^2_f\eta_t}{\rho\mu}\big(F(\bar{x}_t)-f(\bar{x}_t,\bar{y}_t)\big) + \frac{2\gamma\eta_t}{\rho}\|\nabla_x f(\bar{x}_t,\bar{x}_t)-\bar{w}_t\|^2 -\frac{\rho\eta_t}{2\gamma}\|\hat{x}_{t+1}-\bar{x}_t\|^2 \nonumber \\
  & \leq F(\bar{x}_t) + \frac{4\gamma L^2_f\eta_t}{\rho\mu}\big(F(\bar{x}_t)-f(\bar{x}_t,\bar{y}_t)\big) + \frac{4\gamma\eta_t}{\rho}\|\bar{w}_t-\overline{\nabla_x f(x_t,y_t)}\|^2  -\frac{\rho\gamma\eta_t}{2}\|A_t^{-1}\bar{w}_t\|^2\nonumber \\
  & \quad + \frac{8\gamma\eta_t(q-1)L^2_f}{\rho K}\Big(\sum_{l = s_t+1}^{t-1} \gamma^2\eta_l^2 \sum_{k = 1}^K \mathbb{E}\|A^{-1}_l(w^k_l - \bar{w}_l)\|^2 + \sum_{l = s_t+1}^{t-1} \lambda^2\eta_l^2 \sum_{k = 1}^K \mathbb{E}\|B^{-1}_l(v^k_l - \bar{v}_l)\|^2\Big).
\end{align}

Since $\eta_t=\frac{nK^{1/3}}{(m+t)^{1/3}}$ on $t$ is decreasing and $m\geq Kn^3$, we have $\eta_t \leq \eta_0 = \frac{nK^{1/3}}{m^{1/3}} \leq 1$ and $\gamma \leq \frac{m^{1/3}\rho}{4LnK^{1/3}}= \frac{\rho}{2L\eta_0} \leq \frac{\rho}{2L\eta_t}$ for any $t\geq 0$. Similarly, we have $0<\lambda \leq \frac{m^{1/3}}{4L_fnK^{1/3}\rho_u}= \frac{1}{2L_f\eta_0\rho_u} \leq \frac{1}{2L_f\eta_t\rho_u}$.  Since $\eta_t \leq \frac{\rho}{12\sqrt{2}q\lambda L_f} $ for all $t\geq 0$, we have
$\frac{nK^{1/3}}{m^{1/3}} =\eta_0\leq \eta_t \leq \frac{\rho}{12\sqrt{2}\lambda qL_f}$, then we have $m \geq \frac{K\big(12\sqrt{2}n\lambda qL_f\big)^3}{\rho^3}$.
 Due to $0 < \eta_t \leq 1$ and $m\geq (c_1n)^3K$, we have $\alpha_{t+1} = c_1\eta_t^2 \leq c_1\eta_t \leq c_1\eta_0 \leq \frac{c_1nK^{1/3}}{m^{1/3}}\leq 1$.
Similarly, due to $m \geq (c_2n)^3K$, we have $\beta_{t+1}\leq 1$.

 According to Lemma \ref{lem:E1}, we have
 \begin{align}
  & \frac{1}{\eta_t}\mathbb{E} \|\bar{v}_{t+1} - \overline{\nabla_y f(x_{t+1},y_{t+1})}\|^2 - \frac{1}{\eta_{t-1}}\mathbb{E} \|\bar{v}_t - \overline{\nabla_y f(x_t,y_t)}\|^2  \\
  & \leq \big(\frac{1-\alpha_{t+1}}{\eta_t} - \frac{1}{\eta_{t-1}}\big)\mathbb{E} \|\bar{v}_t - \overline{\nabla_y f(x_t,y_t)}\|^2 + \frac{4L^2_f}{K^2}\eta_t\sum_{k=1}^K\big(\|\hat{x}^k_{t+1}-x^k_t\|^2 + \|\hat{y}^k_{t+1}-y^k_t\|^2\big) + \frac{2\alpha_{t+1}^2\sigma^2}{K\eta_t}\nonumber \\
  & = \big(\frac{1}{\eta_t} - \frac{1}{\eta_{t-1}} - c_1\eta_t\big)\mathbb{E} \|\bar{v}_t - \overline{\nabla_y f(x_t,y_t)}\|^2  +\frac{ 4L^2_f}{K^2}\eta_t\sum_{k=1}^K\big(\|\hat{x}^k_{t+1}-x^k_t\|^2 +\|\hat{y}^k_{t+1}-y^k_t\|^2\big) + \frac{2c_1^2\eta^3_t\sigma^2}{K}, \nonumber
 \end{align}
 where the second equality is due to $\alpha_{t+1}=c_1\eta^2_t$.
 Similarly, we have
 \begin{align}
 & \frac{1}{\eta_t}\mathbb{E}\|\bar{w}_{t+1} -\overline{\nabla_x f(x_{t+1},y_{t+1})}\|^2 - \frac{1}{\eta_{t-1}}\mathbb{E}\|\bar{w}_t - \overline{\nabla_x f(x_t,y_t)}\|^2 \\
 & \leq \big(\frac{1-\beta_{t+1}}{\eta_t} -\frac{1}{\eta_{t-1}}\big) \mathbb{E}\|\bar{w}_t - \overline{\nabla_x f(x_t,y_t)}\|^2
 + \frac{4L^2_f}{K^2}\eta_t\sum_{k=1}^K\big( \|\hat{x}^k_{t+1}-x^k_t\|^2 + \|\hat{y}^k_{t+1}-y^k_t\|^2 \big) + \frac{2\beta^2_{t+1}\sigma^2}{K\eta_t} \nonumber \\
 & =  \big(\frac{1}{\eta_t} -\frac{1}{\eta_{t-1}} -c_2\eta_t\big) \mathbb{E}\|\bar{w}_t - \overline{\nabla_xf(x_t,y_t)}\|^2
 + \frac{4L^2_f}{K^2}\eta_t\sum_{k=1}^K\big( \|\hat{x}_{t+1}^k-x^k_t\|^2 + \|\hat{y}^k_{t+1}-y^k_t\|^2 \big) + \frac{2c^2_2\eta^3_t\sigma^2}{K}. \nonumber
 \end{align}
By $\eta_t = \frac{nK^{1/3}}{(m+t)^{1/3}}$, we have
 \begin{align}
  \frac{1}{\eta_t} - \frac{1}{\eta_{t-1}} &= \frac{1}{nK^{1/3}}\big( (m+t)^{\frac{1}{3}} - (m+t-1)^{\frac{1}{3}}\big) \leq \frac{1}{3nK^{1/3}(m+t-1)^{2/3}} \leq \frac{1}{3nK^{1/3}\big(m/2+t\big)^{2/3}} \nonumber \\
  & \leq \frac{2^{2/3}}{3nK^{1/3}(m+t)^{2/3}} = \frac{2^{2/3}}{3n^3K}\frac{n^2K^{2/3}}{(m+t)^{2/3}} = \frac{2^{2/3}}{3n^3K}\eta_t^2 \leq \frac{2}{3n^3K}\eta_t,
 \end{align}
 where the first inequality holds by the concavity of function $f(x)=x^{1/3}$, \emph{i.e.}, $(x+y)^{1/3}\leq x^{1/3} + \frac{y}{3x^{2/3}}$; the second inequality is due to $m\geq 2$,  and
 the last inequality is due to $0<\eta_t\leq 1$.

Let $c_1 \geq \frac{2}{3n^3K} + \frac{9\rho_lL_f^2}{2\rho_u\mu^2}$, we have
 \begin{align} \label{eq:W4}
  & \frac{1}{\eta_t}\mathbb{E} \|\bar{v}_{t+1} - \overline{\nabla_y f(x_{t+1},y_{t+1})}\|^2 - \frac{1}{\eta_{t-1}}\mathbb{E} \|\bar{v}_t - \overline{\nabla_y f(x_t,y_t)}\|^2  \nonumber \\
  & \leq -\frac{9\rho_lL_f^2}{2\rho_u\mu^2} \mathbb{E} \|\bar{v}_t - \overline{\nabla_y f(x_t,y_t)}\|^2 + \frac{4L^2_f}{K^2}\eta_t\sum_{k=1}^K\big(\|\hat{x}^k_{t+1}-x^k_t\|^2 +\|\hat{y}^k_{t+1}-y^k_t\|^2\big)
  + \frac{2c_1^2\eta^3_t\sigma^2}{K} \nonumber \\
  & = -\frac{9\rho_lL_f^2}{2\rho_u\mu^2} \mathbb{E} \|\bar{v}_t - \overline{\nabla_y f(x_t,y_t)}\|^2 + \frac{4L^2_f}{K^2}\eta_t\sum_{k=1}^K\big(\gamma^2\|A^{-1}_t(w^k_t-\bar{w}_t+\bar{w}_t)\|^2 + \lambda^2\|B^{-1}_t(v^k_t-\bar{v}_t+\bar{v}_t)\|^2\big) + \frac{2c_1^2\eta^3_t\sigma^2}{K} \nonumber \\
  & \leq -\frac{9\rho_lL_f^2}{2\rho_u\mu^2} \mathbb{E} \|\bar{v}_t - \overline{\nabla_y f(x_t,y_t)}\|^2 + \frac{8L^2_f}{K^2}\eta_t\sum_{k=1}^K\big(\gamma^2\|A^{-1}_t(w^k_t-\bar{w}_t)\|^2 + \gamma^2\|A^{-1}_t\bar{w}_t\|^2 \nonumber \\
  & \quad + \lambda^2\|B^{-1}_t(v^k_t-\bar{v}_t)\|^2 + \lambda^2\|B^{-1}_t\bar{v}_t\|^2\big) + \frac{2c_1^2\eta^3_t\sigma^2}{K}.
 \end{align}
Let $c_2 \geq \frac{2}{3n^3K} + \frac{9}{2}$, we have
 \begin{align} \label{eq:W5}
  & \frac{1}{\eta_t}\mathbb{E}\|\bar{w}_{t+1} - \overline{\nabla_x f(x_{t+1},y_{t+1})}\|^2 - \frac{1}{\eta_{t-1}}\mathbb{E}\|\bar{w}_t - \overline{\nabla_x f(x_t,y_t)}\|^2 \nonumber \\
  & \leq -\frac{9\eta_t}{2} \mathbb{E}\|\bar{w}_t - \overline{\nabla_x f(x_t,y_t)}\|^2 +
   \frac{4L^2_f}{K^2}\eta_t\sum_{k=1}^K\big(\|\hat{x}^k_{t+1}-x^k_t\|^2 +\|\hat{y}^k_{t+1}-y^k_t\|^2\big) + \frac{2c_2^2\eta_t^3\sigma^2}{K} \nonumber \\
  & \leq -\frac{9\eta_t}{2} \mathbb{E}\|\bar{w}_t - \overline{\nabla_x f(x_t,y_t)}\|^2 +
   \frac{4L^2_f}{K^2}\eta_t\sum_{k=1}^K\big(\gamma^2\|w^k_t-\bar{w}_t+\bar{w}_t\|^2 + \lambda^2\|v^k_t-\bar{v}_t+\bar{v}_t\|^2\big) + \frac{2c_2^2\eta_t^3\sigma^2}{K} \nonumber \\
  & \leq -\frac{9\eta_t}{2} \mathbb{E}\|\bar{w}_t - \overline{\nabla_x f(x_t,y_t)}\|^2 + \frac{8L^2_f}{K^2}\eta_t\sum_{k=1}^K\big(\gamma^2\|A^{-1}_t(w^k_t-\bar{w}_t)\|^2 + \gamma^2\|A^{-1}_t\bar{w}_t\|^2 \nonumber \\
  & \quad + \lambda^2\|B^{-1}_t(v^k_t-\bar{v}_t)\|^2 + \lambda^2\|B^{-1}_t\bar{v}_t\|^2\big) + \frac{2c_2^2\eta_t^3\sigma^2}{K}.
 \end{align}

According to Lemma \ref{lem:C1}, we have
\begin{align} \label{eq:W6}
F(\bar{x}_{t+1}) - f(\bar{x}_{t+1},\bar{y}_{t+1})
& \leq (1-\frac{\eta_t\lambda\mu}{2\rho_u}) \big(F(\bar{x}_t) -f(\bar{x}_t,\bar{y}_t)\big) + \frac{\eta_t}{8\gamma}\|\hat{x}_{t+1}-\bar{x}_t\|^2  -\frac{\eta_t}{4\lambda\rho_u}\|\hat{y}_{t+1}-\bar{y}_t\|^2 \nonumber \\
& \quad + \frac{\eta_t\lambda}{\rho_l}\|\nabla_y f(\bar{x}_t,\bar{y}_t)-\bar{v}_t\|^2 \nonumber \\
& = (1-\frac{\eta_t\lambda\mu}{2\rho_u}) \big(F(\bar{x}_t) -f(\bar{x}_t,\bar{y}_t)\big) + \frac{\gamma\eta_t}{8}\|A^{-1}_t\bar{w}_t\|^2  -\frac{\lambda\eta_t}{4\rho_u}\|B^{-1}_t\bar{v}_t\|^2 \nonumber \\
& \quad + \frac{\eta_t\lambda}{\rho_l}\|\nabla_y f(\bar{x}_t,\bar{y}_t)-\overline{\nabla_y f(x_t,y_t)}+\overline{\nabla_y f(x_t,y_t)}-\bar{v}_t\|^2 \nonumber \\
& \leq (1-\frac{\eta_t\lambda\mu}{2\rho_u}) \big(F(\bar{x}_t) -f(\bar{x}_t,\bar{y}_t)\big) + \frac{\gamma\eta_t}{8}\|A^{-1}_t\bar{w}_t\|^2  -\frac{\lambda\eta_t}{4\rho_u}\|B^{-1}_t\bar{v}_t\|^2 \nonumber \\
& \quad + \frac{2\eta_t\lambda}{\rho_l}\|\nabla_y f(\bar{x}_t,\bar{y}_t)-\overline{\nabla_y f(x_t,y_t)}\|^2 \|+\frac{2\eta_t\lambda}{\rho_l}\|\overline{\nabla_y f(x_t,y_t)}-\bar{v}_t\|^2 \nonumber \\
& \leq (1-\frac{\eta_t\lambda\mu}{2\rho_u}) \big(F(\bar{x}_t) -f(\bar{x}_t,\bar{y}_t)\big) + \frac{\gamma\eta_t}{8}\|A^{-1}_t\bar{w}_t\|^2  -\frac{\lambda\eta_t}{4\rho_u}\|B^{-1}_t\bar{v}_t\|^2 \nonumber \\
& \quad + \frac{4\eta_t\lambda L^2_f}{\rho_lK}\sum_{k=1}^K\big(\|x^k_t-\bar{x}_t\|^2 + \|y^k_t-\bar{y}_t)\|^2\big) + \frac{2\eta_t\lambda}{\rho_l}\|\overline{\nabla_y f(x_t,y_t)}-\bar{v}_t\|^2 \nonumber \\
& \leq (1-\frac{\eta_t\lambda\mu}{2\rho_u}) \big(F(\bar{x}_t) -f(\bar{x}_t,\bar{y}_t)\big) + \frac{\gamma\eta_t}{8}\|A^{-1}_t\bar{w}_t\|^2  -\frac{\lambda\eta_t}{4\rho_u}\|B^{-1}_t\bar{v}_t\|^2 \nonumber \\
& \quad + \frac{4(q-1)\eta_t\lambda L^2_f}{\rho_lK}\Big(\sum_{l = s_t+1}^{t-1} \gamma^2\eta_l^2 \sum_{k = 1}^K \mathbb{E}\|A^{-1}_l(w^k_l - \bar{w}_l)\|^2 + \sum_{l = s_t+1}^{t-1} \lambda^2\eta_l^2 \sum_{k = 1}^K \mathbb{E}\|B^{-1}_l(v^k_l - \bar{v}_l)\|^2\Big) \nonumber\\
& \quad + \frac{2\eta_t\lambda}{\rho_l}\|\overline{\nabla_y f(x_t,y_t)}-\bar{v}_t\|^2,
\end{align}

Next, we define a potential function, for any $t\geq 1$
\begin{align}
 \Omega_t & = \mathbb{E}\Big [F(\bar{x}_t) + \frac{9\rho_u\gamma L^2_f}{\rho\lambda\mu^2}\big(F(\bar{x}_t) -f(\bar{x}_t,\bar{y}_t)\big) + \frac{\gamma}{\rho\eta_{t-1}} \big(\|\bar{v}_t - \overline{\nabla_y f(x_t,y_t)}\|^2
 + \|\bar{w}_t - \overline{\nabla_xf(x_t,y_t)}\|^2 \big) \Big]. \nonumber
\end{align}
Then we have
 \begin{align} \label{eq:W7}
 & \Omega_{t+1} - \Omega_t \nonumber \\
 & = F(\bar{x}_{t+1}) - F(\bar{x}_t) + \frac{9\rho_u\gamma L^2_f}{\rho\lambda\mu^2}\Big(F(\bar{x}_{t+1}) - f(\bar{x}_{t+1},\bar{y}_{t+1}) - \big(F(\bar{x}_t) -f(\bar{x}_t,\bar{y}_t)\big) \Big)
 + \frac{\gamma}{\rho} \bigg( \frac{1}{\eta_t}\mathbb{E}\|\bar{v}_{t+1} - \overline{\nabla_y f(x_{t+1},y_{t+1})}\|^2 \nonumber \\
 & \quad - \frac{1}{\eta_{t-1}}\mathbb{E}\|\bar{v}_t - \overline{\nabla_y f(x_t,y_t)}\|^2 + \frac{1}{\eta_t}\mathbb{E}\|\bar{w}_{t+1} - \overline{\nabla_x f(x_{t+1},y_{t+1})}\|^2
 - \frac{1}{\eta_{t-1}}\mathbb{E}\|\bar{w}_t - \overline{\nabla_x f(x_t,y_t)}\|^2 \bigg) \nonumber \\
 & \leq F(\bar{x}_t) + \frac{4\gamma L^2_f\eta_t}{\rho\mu}\big(F(\bar{x}_t)-f(\bar{x}_t,\bar{y}_t)\big) + \frac{4\gamma\eta_t}{\rho}\|\bar{w}_t-\overline{\nabla_x f(x_t,y_t)}\|^2  -\frac{\rho\gamma\eta_t}{2}\|A_t^{-1}\bar{w}_t\|^2\nonumber \\
  & \quad + \frac{8\gamma\eta_t(q-1)L^2_f}{\rho K}\Big(\sum_{l = s_t+1}^{t-1} \gamma^2\eta_l^2 \sum_{k = 1}^K \mathbb{E}\|A^{-1}_l(w^k_l - \bar{w}_l)\|^2 + \sum_{l = s_t+1}^{t-1} \lambda^2\eta_l^2 \sum_{k = 1}^K \mathbb{E}\|B^{-1}_l(v^k_l - \bar{v}_l)\|^2\Big) \nonumber \\
 & \quad + \frac{9\rho_u\gamma L^2_f}{\rho\lambda\mu^2} \Bigg( -\frac{\eta_t\lambda\mu}{2\rho_u}\big(F(\bar{x}_t) -f(\bar{x}_t,\bar{y}_t)\big) + \frac{\gamma\eta_t}{8}\|A^{-1}_t\bar{w}_t\|^2  -\frac{\lambda\eta_t}{4\rho_u}\|B^{-1}_t\bar{v}_t\|^2 \nonumber \\
& \qquad + \frac{4(q-1)\eta_t\lambda L^2_f}{\rho_lK}\Big(\sum_{l = s_t+1}^{t-1} \gamma^2\eta_l^2 \sum_{k = 1}^K \mathbb{E}\|A^{-1}_l(w^k_l - \bar{w}_l)\|^2 + \sum_{l = s_t+1}^{t-1} \lambda^2\eta_l^2 \sum_{k = 1}^K \mathbb{E}\|B^{-1}_l(v^k_l - \bar{v}_l)\|^2\Big) + \frac{2\eta_t\lambda}{\rho_l}\|\bar{v}_t - \overline{\nabla_y f(x_t,y_t)}\|^2 \Bigg)  \nonumber \\
 & \quad + \frac{\gamma}{\rho} \bigg( -\frac{9\rho_lL_f^2}{2\rho_u\mu^2} \mathbb{E} \|\bar{v}_t - \overline{\nabla_y f(x_t,y_t)}\|^2 + \frac{8L^2_f}{K^2}\eta_t\sum_{k=1}^K\big(\gamma^2\|A^{-1}_t(w^k_t-\bar{w}_t)\|^2 + \gamma^2\|A^{-1}_t\bar{w}_t\|^2 \nonumber \\
  & \qquad + \lambda^2\|B^{-1}_t(v^k_t-\bar{v}_t)\|^2 + \lambda^2\|B^{-1}_t\bar{v}_t\|^2\big) + \frac{2c_1^2\eta^3_t\sigma^2}{K} \nonumber \\
 & \qquad -\frac{9\eta_t}{2} \mathbb{E}\|\bar{w}_t - \overline{\nabla_x f(x_t,y_t)}\|^2 + \frac{8L^2_f}{K^2}\eta_t\sum_{k=1}^K\big(\gamma^2\|A^{-1}_t(w^k_t-\bar{w}_t)\|^2 + \gamma^2\|A^{-1}_t\bar{w}_t\|^2 \nonumber \\
  & \qquad + \lambda^2\|B^{-1}_t(v^k_t-\bar{v}_t)\|^2 + \lambda^2\|B^{-1}_t\bar{v}_t\|^2\big) + \frac{2c_2^2\eta_t^3\sigma^2}{K} \bigg) \nonumber \\
 & = - \frac{\gamma L^2_f\eta_t}{2\mu\rho}\big(F(\bar{x}_t)-f(\bar{x}_t,\bar{y}_t)\big) - \frac{\gamma\eta_t}{2\rho}\|\bar{w}_t-\overline{\nabla_x f(x_t,y_t)}\|^2 -\Big(\frac{\eta_t\gamma\rho}{2}-\frac{9\rho_u\gamma^2 L^2_f\eta_t}{8\rho\lambda\mu^2}-\frac{16\gamma^3L^2_f\eta_t}{\rho K}\Big)\|A^{-1}_t\bar{w}_t\|^2 \nonumber \\
 & \quad + \Big(\frac{8\gamma\eta_t(q-1)L^2_f}{\rho K} + \frac{36(q-1)\eta_t L^4_f\gamma\rho_u}{\mu^2K\rho}\Big)\sum_{l = s_t+1}^{t-1} \big(\gamma^2\eta_l^2 \sum_{k = 1}^K \mathbb{E}\|A^{-1}_l(w^k_l - \bar{w}_l)\|^2 + \sum_{l = s_t+1}^{t-1} \lambda^2\eta_l^2 \sum_{k = 1}^K \mathbb{E}\|B^{-1}_l(v^k_l - \bar{v}_l)\|^2\big) \nonumber \\
 & \quad -\big(\frac{9\gamma L^2_f\eta_t}{4\mu^2\rho}-\frac{16\gamma\lambda^2L^2_f\eta_t}{K\rho}\big)\|B^{-1}_t\bar{v}_t\|^2+ \frac{2(c_1^2+c_2^2)\gamma\eta_t^3\sigma^2}{K\rho} \nonumber \\
 & \quad + \frac{16\gamma^3L^2_f\eta_t}{K^2\rho}\sum_{k=1}^K\|A^{-1}_t(w^k_t-\bar{w}_t)\|^2 + \frac{16\gamma\lambda^2L^2_f\eta_t}{K^2\rho}\sum_{k=1}^K\|B^{-1}_t(v^k_t-\bar{v}_t)\|^2,
 \end{align}
where the above inequality holds by the above inequalities \eqref{eq:W3}, \eqref{eq:W4}, \eqref{eq:W5} and \eqref{eq:W6}.

Here considering the term $\|\bar{w}_t - \nabla_xf(\bar{x}_t,\bar{y}_t)\|^2$, we have
\begin{align}
 & \|\bar{w}_t-\nabla_x f(\bar{x}_t,\bar{x}_t)\|^2 \nonumber \\
& = \|\bar{w}_t-\overline{\nabla_x f(x_t,y_t)}+\overline{\nabla_x f(x_t,y_t)}-\nabla_x f(\bar{x}_t,\bar{x}_t)\|^2 \nonumber \\
& \leq 2\|\bar{w}_t-\overline{\nabla_x f(x_t,y_t)}\|^2 + 2\|\overline{\nabla_x f(x_t,y_t)}-\nabla_x f(\bar{x}_t,\bar{x}_t)\|^2 \nonumber \\
& \leq 2\|\bar{w}_t-\overline{\nabla_x f(x_t,y_t)}\|^2 + 2\|\frac{1}{K}\sum_{k=1}^K\nabla_x f^k(x^k_t,y^k_t)-\frac{1}{K}\sum_{k=1}^K\nabla_x f^k(\bar{x}_t,\bar{y}_t)\|^2 \nonumber \\
& \leq 2\|\bar{w}_t-\overline{\nabla_x f(x_t,y_t)}\|^2 + \frac{4L^2_f}{K}\sum_{k=1}^K \big(\|x^k_t-\bar{x}_t\|^2 +\|y^k_t-\bar{y}_t)\|^2\big) \nonumber \\
& \leq 2\|\bar{w}_t-\overline{\nabla_x f(x_t,y_t)}\|^2 + \frac{4(q-1)L^2_f}{K}\Big(\sum_{l = s_t+1}^{t-1} \gamma^2\eta_l^2 \sum_{k = 1}^K \mathbb{E}\|A^{-1}_l(w^k_l - \bar{w}_l)\|^2 + \sum_{l = s_t+1}^{t-1} \lambda^2\eta_l^2 \sum_{k = 1}^K \mathbb{E}\|B^{-1}_l(v^k_l - \bar{v}_l)\|^2\Big),
\end{align}
then we obtain
\begin{align} \label{eq:W8}
 & -\|\bar{w}_t-\overline{\nabla_x f(x_t,y_t)}\|^2 \nonumber \\
& \leq -\frac{1}{2}\|\bar{w}_t-\nabla_x f(\bar{x}_t,\bar{x}_t)\|^2+ \frac{2(q-1)L^2_f}{K}\Big(\sum_{l = s_t+1}^{t-1} \gamma^2\eta_l^2 \sum_{k = 1}^K \mathbb{E}\|A^{-1}_l(w^k_l - \bar{w}_l)\|^2 + \sum_{l = s_t+1}^{t-1} \lambda^2\eta_l^2 \sum_{k = 1}^K \mathbb{E}\|B^{-1}_l(v^k_l - \bar{v}_l)\|^2\Big),
\end{align}

By combining the above inequalities~\ref{eq:W7} with~\ref{eq:W8}, we can obtain
\begin{align} \label{eq:W9}
 & \Omega_{t+1} - \Omega_t \nonumber \\
 & \leq - \frac{\gamma L^2_f\eta_t}{2\mu\rho}\big(F(\bar{x}_t)-f(\bar{x}_t,\bar{y}_t)\big) - \frac{\gamma\eta_t}{4\rho}\|\bar{w}_t-\nabla_x f(\bar{x}_t,\bar{x}_t)\|^2 -\Big(\frac{\eta_t\gamma\rho}{2}-\frac{9\rho_u\gamma^2 L^2_f\eta_t}{8\rho\lambda\mu^2}-\frac{16\gamma^3L^2_f\eta_t}{\rho K}\Big)\|A^{-1}_t\bar{w}_t\|^2 \nonumber \\
 & \quad + \Big(\frac{9\gamma\eta_t(q-1)L^2_f}{\rho K} + \frac{36(q-1)\eta_t L^4_f\gamma\rho_u}{\mu^2K\rho}\Big)\sum_{l = s_t+1}^{t-1} \big(\gamma^2\eta_l^2 \sum_{k = 1}^K \mathbb{E}\|A^{-1}_l(w^k_l - \bar{w}_l)\|^2 + \sum_{l = s_t+1}^{t-1} \lambda^2\eta_l^2 \sum_{k = 1}^K \mathbb{E}\|B^{-1}_l(v^k_l - \bar{v}_l)\|^2\big) \nonumber \\
 & \quad -\big(\frac{9\gamma L^2_f\eta_t}{4\mu^2\rho}-\frac{16\gamma\lambda^2L^2_f\eta_t}{K\rho}\big)\|B^{-1}_t\bar{v}_t\|^2+ \frac{2(c_1^2+c_2^2)\gamma\eta_t^3\sigma^2}{K\rho} \nonumber \\
 & \quad + \frac{16\gamma^3L^2_f\eta_t}{K^2\rho}\sum_{k=1}^K\|A^{-1}_t(w^k_t-\bar{w}_t)\|^2 + \frac{16\gamma\lambda^2L^2_f\eta_t}{K^2\rho}\sum_{k=1}^K\|B^{-1}_t(v^k_t-\bar{v}_t)\|^2 .
 \end{align}

Let $s_t=q\lfloor t/q\rfloor$, summing the above inequality \eqref{eq:W9} over $t=s_t$ to $s_t+q-1$,
we have
 \begin{align} \label{eq:W10}
 & \sum_{t=s_t}^{s_t+q-1}\big( \Omega_{t+1} - \Omega_t \big) \nonumber \\
 & \leq \sum_{t=s_t}^{s_t+q-1}\Bigg( - \frac{\gamma L^2_f\eta_t}{2\mu\rho}\big(F(\bar{x}_t)-f(\bar{x}_t,\bar{y}_t)\big) - \frac{\gamma\eta_t}{4\rho}\|\bar{w}_t-\nabla_x f(\bar{x}_t,\bar{x}_t)\|^2 -\Big(\frac{\eta_t\gamma\rho}{2}-\frac{9\rho_u\gamma^2 L^2_f\eta_t}{8\rho\lambda\mu^2}-\frac{16\gamma^3L^2_f\eta_t}{\rho K}\Big)\|A^{-1}_t\bar{w}_t\|^2 \nonumber \\
 & \quad + \Big(\frac{9\gamma\eta_t(q-1)L^2_f}{\rho K} + \frac{36(q-1)\eta_t L^4_f\gamma\rho_u}{\mu^2K\rho}\Big)\sum_{l = s_t+1}^{t-1} \big(\gamma^2\eta_l^2 \sum_{k = 1}^K \mathbb{E}\|A^{-1}_l(w^k_l - \bar{w}_l)\|^2 + \sum_{l = s_t+1}^{t-1} \lambda^2\eta_l^2 \sum_{k = 1}^K \mathbb{E}\|B^{-1}_l(v^k_l - \bar{v}_l)\|^2\big) \nonumber \\
 & \quad -\big(\frac{9\gamma L^2_f\eta_t}{4\mu^2\rho}-\frac{16\gamma\lambda^2L^2_f\eta_t}{K\rho}\big)\|B^{-1}_t\bar{v}_t\|^2+ \frac{2(c_1^2+c_2^2)\gamma\eta_t^3\sigma^2}{K\rho} \nonumber \\
 & \quad + \frac{16\gamma^3L^2_f\eta_t}{K^2\rho}\sum_{k=1}^K\|A^{-1}_t(w^k_t-\bar{w}_t)\|^2 + \frac{16\gamma\lambda^2L^2_f\eta_t}{K^2\rho}\sum_{k=1}^K\|B^{-1}_t(v^k_t-\bar{v}_t)\|^2 \Bigg) \nonumber \\
 & \leq \sum_{t=s_t}^{s_t+q-1}\Bigg( - \frac{\gamma L^2_f\eta_t}{2\mu\rho}\big(F(\bar{x}_t)-f(\bar{x}_t,\bar{y}_t)\big) - \frac{\gamma\eta_t}{4\rho}\|\bar{w}_t-\nabla_x f(\bar{x}_t,\bar{x}_t)\|^2 -\Big(\frac{\eta_t\gamma\rho}{2}-\frac{9\rho_u\gamma^2 L^2_f\eta_t}{8\rho\lambda\mu^2}-\frac{16\gamma^3L^2_f\eta_t}{\rho K}\Big)\|A^{-1}_t\bar{w}_t\|^2 \nonumber \\
 & \quad + \Big(\frac{\rho\gamma\eta_t}{16K} + \frac{L^2_f\gamma\eta_t\rho\rho_u}{4\mu^2K}\Big)\sum_{k = 1}^K \big(\mathbb{E}\|A^{-1}_t(w^k_t - \bar{w}_t)\|^2 + \mathbb{E}\|B^{-1}_t(v^k_t - \bar{v}_t)\|^2\big) \nonumber \\
 & \quad -\big(\frac{9\gamma L^2_f\eta_t}{4\mu^2\rho}-\frac{16\gamma\lambda^2L^2_f\eta_t}{K\rho}\big)\|B^{-1}_t\bar{v}_t\|^2+ \frac{2(c_1^2+c_2^2)\gamma\eta_t^3\sigma^2}{K\rho} \nonumber \\
 & \quad + \frac{16\gamma^3L^2_f\eta_t}{K^2\rho}\sum_{k=1}^K\|A^{-1}_t(w^k_t-\bar{w}_t)\|^2 + \frac{16\gamma\lambda^2L^2_f\eta_t}{K^2\rho}\sum_{k=1}^K\|B^{-1}_t(v^k_t-\bar{v}_t)\|^2 \Bigg) \nonumber \\
& =\sum_{t=s_t}^{s_t+q-1}\Bigg( - \frac{\gamma L^2_f\eta_t}{2\mu\rho}\big(F(\bar{x}_t)-f(\bar{x}_t,\bar{y}_t)\big) - \frac{\gamma\eta_t}{4\rho}\|\bar{w}_t-\nabla_x f(\bar{x}_t,\bar{x}_t)\|^2 -\Big(\frac{\eta_t\gamma\rho}{2}-\frac{9\rho_u\gamma^2 L^2_f\eta_t}{8\rho\lambda\mu^2}-\frac{16\gamma^3L^2_f\eta_t}{\rho K}\Big)\|A^{-1}_t\bar{w}_t\|^2 \nonumber \\
& \quad -\big(\frac{9\gamma L^2_f\eta_t}{4\mu^2\rho}-\frac{16\gamma\lambda^2L^2_f\eta_t}{K\rho}\big)\|B^{-1}_t\bar{v}_t\|^2+ \frac{2(c_1^2+c_2^2)\gamma\eta_t^3\sigma^2}{K\rho} \nonumber \\
& \quad  + \Big(\frac{\rho\gamma\eta_t}{16K} + \frac{L^2_f\gamma\eta_t\rho\rho_u}{4\mu^2K}+ \frac{16\gamma^3L^2_f\eta_t}{K^2\rho}\Big)\sum_{k = 1}^K \mathbb{E}\|A^{-1}_t(w^k_t - \bar{w}_t)\|^2 \nonumber \\
& \quad + \Big(\frac{\rho\gamma\eta_t}{16K} + \frac{L^2_f\gamma\eta_t\rho\rho_u}{4\mu^2K}+ \frac{16\gamma\lambda^2L^2_f\eta_t}{K^2\rho}\Big)\sum_{k = 1}^K \mathbb{E}\|B^{-1}_t(v^k_t - \bar{v}_t)\|^2 \Bigg) ,
 \end{align}
where the second inequality is due to $\lambda\geq \gamma > 0$ and $\eta_t \leq \frac{\rho}{12\lambda qL_f} $ for all $t\geq 1$.

According to the above inequality \eqref{eq:W10}, we have
\begin{align} \label{eq:W11}
 & \sum_{t=s_t}^{s_t+q-1}\big( \Omega_{t+1} - \Omega_t \big) \nonumber \\
 & \leq \sum_{t=s_t}^{s_t+q-1}\Bigg( - \frac{\gamma L^2_f\eta_t}{2\mu\rho}\big(F(\bar{x}_t)-f(\bar{x}_t,\bar{y}_t)\big) - \frac{\gamma\eta_t}{4\rho}\|\bar{w}_t-\nabla_x f(\bar{x}_t,\bar{x}_t)\|^2 -\Big(\frac{\eta_t\gamma\rho}{2}-\frac{9\rho_u\gamma^2 L^2_f\eta_t}{8\rho\lambda\mu^2}-\frac{16\gamma^3L^2_f\eta_t}{\rho K}\Big)\|A^{-1}_t\bar{w}_t\|^2 \nonumber \\
& \quad -\big(\frac{9\gamma L^2_f\eta_t}{4\mu^2\rho}-\frac{16\gamma\lambda^2L^2_f\eta_t}{K\rho}\big)\|B^{-1}_t\bar{v}_t\|^2+ \frac{2(c_1^2+c_2^2)\gamma\eta_t^3\sigma^2}{K\rho} \nonumber \\
& \quad  + \Big(\frac{\rho}{16} + \frac{L^2_f\rho\rho_u}{4\mu^2}+ \frac{16\lambda^2L^2_f}{K\rho}\Big) \frac{\gamma\eta_t}{K}\sum_{k = 1}^K \big(\mathbb{E}\|A^{-1}_t(w^k_t - \bar{w}_t)\|^2 + \mathbb{E}\|B^{-1}_t(v^k_t - \bar{v}_t)\|^2\big) \Bigg) \nonumber \\
 & \leq \sum_{t=s_t}^{s_t+q-1}\Bigg( - \frac{\gamma L^2_f\eta_t}{2\mu\rho}\big(F(\bar{x}_t)-f(\bar{x}_t,\bar{y}_t)\big) - \frac{\gamma\eta_t}{4\rho}\|\bar{w}_t-\nabla_x f(\bar{x}_t,\bar{x}_t)\|^2 -\Big(\frac{\eta_t\gamma\rho}{2}-\frac{9\rho_u\gamma^2 L^2_f\eta_t}{8\rho\lambda\mu^2}-\frac{16\gamma^3L^2_f\eta_t}{\rho K}\Big)\|A^{-1}_t\bar{w}_t\|^2 \nonumber \\
& \quad -\big(\frac{9\gamma L^2_f\eta_t}{4\mu^2\rho}-\frac{16\gamma\lambda^2L^2_f\eta_t}{K\rho}\big)\|B^{-1}_t\bar{v}_t\|^2+ \frac{2(c_1^2+c_2^2)\gamma\eta_t^3\sigma^2}{K\rho} \nonumber \\
& \quad  + \Big(\frac{\rho}{16} + \frac{L^2_f\rho\rho_u}{4\mu^2}+ \frac{16\lambda^2L^2_f}{K\rho}\Big) \Big(  \frac{8\gamma\eta_t}{15}\mathbb{E}\big(\tau^2\|A^{-1}_t\bar{w}_t\|^2+\|B^{-1}_t\bar{v}_t\|^2\big)
   + \frac{2\gamma\Delta }{15\lambda^2L^2_f}\eta^3_t\Big) \Bigg) \nonumber \\
 & = \sum_{t=s_t}^{s_t+q-1}\Bigg( - \frac{\gamma L^2_f\eta_t}{2\mu\rho}\big(F(\bar{x}_t)-f(\bar{x}_t,\bar{y}_t)\big) - \frac{\gamma\eta_t}{4\rho}\mathbb{E}\|\bar{w}_t-\nabla_x f(\bar{x}_t,\bar{x}_t)\|^2 \nonumber \\
& \quad + \frac{2(c_1^2+c_2^2)\gamma\sigma^2}{K\rho}\eta_t^3 + \Big(\frac{\rho}{16} + \frac{L^2_f\rho\rho_u}{4\mu^2}+ \frac{16\lambda^2L^2_f}{K\rho}\Big)\frac{2\gamma\Delta }{15\lambda^2L^2_f}\eta^3_t\nonumber \\
& \quad -\Big(\frac{9\gamma L^2_f\eta_t}{4\mu^2\rho}-\frac{16\gamma\lambda^2L^2_f\eta_t}{K\rho} - \Big(\frac{\rho}{16} + \frac{L^2_f\rho\rho_u}{4\mu^2}+ \frac{16\lambda^2L^2_f}{K\rho}\Big)\frac{8\gamma\eta_t}{15}\Big)\mathbb{E}\|B^{-1}_t\bar{v}_t\|^2 \nonumber \\
& \quad  -\Big(\frac{\eta_t\gamma\rho}{2}-\frac{9\rho_u\gamma^2 L^2_f\eta_t}{8\rho\lambda\mu^2}-\frac{16\gamma^3L^2_f\eta_t}{\rho K} - \Big(\frac{\rho}{16} + \frac{L^2_f\rho\rho_u}{4\mu^2}+ \frac{16\lambda^2L^2_f}{K\rho}\Big)\frac{8\gamma\tau^2\eta_t}{15} \Big)\mathbb{E}\|A^{-1}_t\bar{w}_t\|^2 \Bigg),
 \end{align}
where the second inequality holds by Lemma \ref{lem:A9}.

Let $\Lambda= \frac{1}{16} + \frac{L^2_f\rho_u}{4\mu^2}+ \frac{16\lambda^2L^2_f}{K\rho^2}$. Since $\tau \leq \min\big(\frac{\sqrt{5K}}{4\sqrt{2\Lambda}},1\big)$, we have
$\frac{\gamma\eta_t\rho}{12} \geq \frac{8\rho\gamma \Lambda\tau^2}{15K}\eta_t$. Since $\gamma \leq \min\big( \frac{2\lambda\mu^2\rho^2}{27L^2_f\rho_u}, \frac{\sqrt{K}\rho}{8\sqrt{3}L_f}\big)$, we have $\frac{\gamma\eta_t\rho}{12} \geq \frac{9\gamma^2 L^2_f\eta_t\rho_u}{8\lambda\mu^2\rho}$ and $\frac{\gamma\eta_t\rho}{12}\geq \frac{16\gamma^3L^2_f\eta_t}{K\rho}$. Thus, we have
\begin{align}
 \frac{\gamma\eta_t\rho}{2}-\frac{9\gamma^2 L^2_f\eta_t\rho_u}{8\lambda\mu^2\rho}-\frac{16\gamma^3L^2_f\eta_t}{K\rho} - \frac{8\Lambda\rho \gamma\tau^2\eta_t}{15}  \geq \frac{\gamma\eta_t\rho}{4}.
\end{align}
Due to $\lambda\leq \frac{3\sqrt{K}}{8\sqrt{2}\mu}$, we have
$\frac{9\gamma L^2_f\eta_t}{8\mu^2\rho}\geq \frac{16\gamma\lambda^2L^2_f\eta_t}{K\rho} $.
Since $\frac{L_f}{\mu}\geq 1$ and $\eta_t>0$ for all $t\geq1$, let $0<\rho\leq 1$ and $0<\rho_u\leq \frac{135}{64\rho^2}$, we have $\frac{9 L^2_f\gamma\eta_t}{16\mu^2\rho} \geq \big(\frac{\rho}{16}+ \frac{L^2_f\rho\rho_u}{4\mu^2}\big)\frac{8\gamma\eta_t}{15}$.
Due to $\lambda \leq \frac{3\sqrt{5K}}{32\sqrt{2}\mu}$, we have
$\frac{9\gamma L^2_f\eta_t}{16\mu^2\rho}\geq \frac{16L_f^2\lambda^2}{K\rho}\frac{8\gamma\eta_t}{15}$.
Thus, we have
\begin{align}
 \frac{9\gamma L^2_f\eta_t}{4\mu^2\rho}-\frac{16\gamma\lambda^2L^2_f\eta_t}{K\rho} - \Big(\frac{\rho}{16} + \frac{L^2_f\rho\rho_u}{4\mu^2}+ \frac{16\lambda^2L^2_f}{K\rho}\Big)\frac{8\gamma\eta_t}{15} \geq 0.
\end{align}

Let $\tau \leq \min\big(\frac{\sqrt{5K}}{4\sqrt{2\Lambda}},1\big)$, $\gamma \leq  \min\big( \frac{2\lambda\mu^2\rho}{27L^2_f\rho_u}, \frac{\sqrt{K}\rho}{8\sqrt{3}L_f}\big)$ and $\lambda \leq \min\big( \frac{3\sqrt{K}}{8\sqrt{2}\mu},\frac{3\sqrt{5K}}{32\sqrt{2}\mu}\big)=
\frac{3\sqrt{5K}}{32\sqrt{2}\mu}$, thus we can obtain
\begin{align} \label{eq:W12}
 & \sum_{t=s_t}^{s_t+q-1}\big( \Omega_{t+1} - \Omega_t \big) \nonumber \\
 & \leq \sum_{t=s_t}^{s_t+q-1}\Bigg( - \frac{\gamma L^2_f\eta_t}{2\mu\rho}\big(F(\bar{x}_t)-f(\bar{x}_t,\bar{y}_t)\big) - \frac{\gamma\eta_t}{4\rho}\mathbb{E}\|\bar{w}_t-\nabla_x f(\bar{x}_t,\bar{x}_t)\|^2 - \frac{\gamma\eta_t\rho}{4}\mathbb{E}\|A^{-1}_t\bar{w}_t\|^2\nonumber \\
& \quad + \frac{2(c_1^2+c_2^2)\gamma\sigma^2}{K\rho}\eta_t^3 + \Big(\frac{\rho}{16} + \frac{L^2_f\rho\rho_u}{4\mu^2}+ \frac{16\lambda^2L^2_f}{K\rho}\Big)\frac{2\gamma\Delta }{15\lambda^2L^2_f}\eta^3_t \Bigg),
 \end{align}

Summing the above inequality~\eqref{eq:W12} from $t=1$ to $T$, then we have
\begin{align} \label{eq:W13}
 & \sum_{t=1}^{T}\big( \Omega_{t+1} - \Omega_t \big) \nonumber \\
 & \leq  - \frac{\gamma L^2_f}{2\mu\rho}\sum_{t=1}^{T}\eta_t\big(F(\bar{x}_t)-f(\bar{x}_t,\bar{y}_t)\big) - \frac{\gamma}{4\rho}\sum_{t=1}^{T}\eta_t\mathbb{E}\|\bar{w}_t-\nabla_x f(\bar{x}_t,\bar{x}_t)\|^2 - \frac{\gamma\rho}{4}\sum_{t=1}^{T}\eta_t\mathbb{E}\|A^{-1}_t\bar{w}_t\|^2\nonumber \\
& \quad + \frac{2(c_1^2+c_2^2)\gamma\sigma^2}{K\rho}\sum_{t=1}^{T}\eta_t^3 + \frac{2\Lambda\rho\gamma\Delta }{15K\lambda^2L^2_f}\sum_{t=1}^{T}\eta^3_t .
 \end{align}

Since $v^k_1 = \frac{1}{q}\sum_{j=1}^q\nabla_y f^k(x^k_1,y^k_1;\xi^k_{1,j})$, and $w^k_1 = \frac{1}{q}\sum_{j=1}^q\nabla_xf^k(x^k_1,y^k_1;\xi^k_{1,j})$, we have
\begin{align} \label{eq:W14}
 \Omega_1 &= \mathbb{E}\Big[F(\bar{x}_1) + \frac{9\rho_u\gamma L^2_f}{\rho\lambda\mu^2}\big(F(\bar{x}_1) -f(\bar{x}_1,\bar{y}_1)\big) + \frac{\gamma}{\rho\eta_0} \big(\|\bar{v}_t - \overline{\nabla_y f(x_t,y_t)}\|^2
 + \|\bar{w}_t - \overline{\nabla_xf(x_t,y_t)}\|^2 \big)  \Big] \nonumber \\
 & \leq F(\bar{x}_1) + \frac{9\rho_u\gamma L^2_f}{\rho\lambda\mu^2}\big(F(\bar{x}_1) -f(\bar{x}_1,\bar{y}_1)\big) + \frac{2\gamma\sigma^2}{qK\rho\eta_0},
\end{align}
where the last inequality holds by Assumption \ref{ass:3}.

Since $\eta_t=\frac{nK^{1/3}}{(m+t)^{1/3}}$ is decreasing, i.e., $\eta_T^{-1} \geq \eta_t^{-1}$ for any $0\leq t\leq T$, we have
 \begin{align}  \label{eq:W15}
 & \frac{1}{T} \sum_{t=1}^T \mathbb{E}\Big[ \frac{2L^2_f}{\rho^2\mu}\big(F(\bar{x}_t)-f(\bar{x}_t,\bar{y}_t)\big) + \frac{1}{\rho^2}\|\bar{w}_t-\nabla_x f(\bar{x}_t,\bar{x}_t)\|^2 + \|A^{-1}_t\bar{w}_t\|^2 \Big]  \\
 & \leq  \frac{4}{T\rho\gamma\eta_T} \sum_{t=1}^T\big(\Omega_t - \Omega_{t+1}\big)
 + \Big( \frac{(c_1^2+c_2^2)\sigma^2}{\rho^2K} + \frac{\Lambda\Delta }{15K\lambda^2L^2_f}\Big)\frac{8}{T\eta_T}\sum_{t=1}^{T}\eta^3_t \nonumber \\
 & \leq \frac{4}{T\rho\gamma\eta_T} \Big(F(\bar{x}_1) -F^* + \frac{9\rho_u\gamma L^2_f}{\rho\lambda\mu^2}\big(F(\bar{x}_1) -f(\bar{x}_1,\bar{y}_1)+ \frac{2\gamma\sigma^2}{qK\rho\eta_0}\Big)
 + \Big( \frac{(c_1^2+c_2^2)\sigma^2}{\rho^2K} + \frac{\Lambda\Delta }{15K\lambda^2L^2_f}\Big)\frac{8}{T\eta_T}\sum_{t=1}^{T}\eta^3_t  \nonumber \\
 & \leq \frac{4}{T\rho\gamma\eta_T} \Big(F(\bar{x}_1) -F^* + \frac{9\rho_u\gamma L^2_f}{\rho\lambda\mu^2}\big(F(\bar{x}_1) -f(\bar{x}_1,\bar{y}_1)+ \frac{2\gamma\sigma^2}{qK\rho\eta_0}\Big)
 + \Big( \frac{(c_1^2+c_2^2)\sigma^2}{\rho^2K} + \frac{\Lambda\Delta }{15K\lambda^2L^2_f}\Big)\frac{8}{T\eta_T}\int^T_1\frac{Kn^3}{m+t} dt \nonumber \\
 & \leq\frac{4}{T\rho\gamma\eta_T} \Big(F(\bar{x}_1) -F^* + \frac{9\rho_u\gamma L^2_f}{\rho\lambda\mu^2}\big(F(\bar{x}_1) -f(\bar{x}_1,\bar{y}_1)+ \frac{2\gamma\sigma^2}{qK\rho\eta_0}\Big)
 + \Big( \frac{(c_1^2+c_2^2)\sigma^2}{\rho^2K} + \frac{\Lambda\Delta }{15K\lambda^2L^2_f}\Big)\frac{8Kn^3}{T\eta_T}\ln(m+t) \nonumber \\
 & = \bigg( \frac{4(F(\bar{x}_1) - F^*)}{\rho\gamma n} + \frac{36\rho_u L^2_f}{\rho\lambda\mu^2n}\big(F(\bar{x}_1) -f(\bar{x}_1,\bar{y}_1) + \frac{8m^{1/3}\sigma^2}{qK^{4/3}n^2\rho} +  8Kn^2\Big( \frac{(c_1^2+c_2^2)\sigma^2}{\rho^2K} + \frac{\Lambda\Delta }{15K\lambda^2L^2_f}\Big)\ln(m+t) \bigg) \nonumber \\
 & \quad \cdot \frac{(m+T)^{1/3}}{K^{1/3}T}, \nonumber
\end{align}
where the second inequality holds by the above inequality \eqref{eq:W14}.

Let $G = \frac{4(F(\bar{x}_1) - F^*)}{\rho\gamma n} + \frac{36\rho_u L^2_f}{\rho\lambda\mu^2n}\big(F(\bar{x}_1) -f(\bar{x}_1,\bar{y}_1) + \frac{8m^{1/3}\sigma^2}{qK^{4/3}n^2\rho} +  8Kn^2\Big( \frac{(c_1^2+c_2^2)\sigma^2}{\rho^2K} + \frac{\Lambda\Delta }{15K\lambda^2L^2_f}\Big)\ln(m+t)$,
we have
\begin{align} \label{eq:W16}
 \frac{1}{T} \sum_{t=1}^T \mathbb{E}\Big[ \frac{2L^2_f}{\rho^2\mu}\big(F(\bar{x}_t)-f(\bar{x}_t,\bar{y}_t)\big) + \frac{1}{\rho^2}\|\bar{w}_t-\nabla_x f(\bar{x}_t,\bar{x}_t)\|^2 + \|A^{-1}_t\bar{w}_t\|^2 \Big]  \leq \frac{G}{K^{1/3}T}(m+T)^{1/3}.
\end{align}

We define a useful metric
\begin{align}
\mathcal{M}_t  = \frac{1}{\rho}\big( \frac{\sqrt{2}L_f}{\sqrt{\mu}} \sqrt{F(\bar{x}_t)-f(\bar{x}_t,\bar{y}_t)}  + \|\nabla_x f(\bar{x}_t,\bar{y}_t)- \bar{w}_t\|\big) + \|A^{-1}_t\bar{w}_t\|
.
\end{align}
According to the above inequality~\ref{eq:W16}, we have
\begin{align} \label{eq:W17}
 \frac{1}{T}\sum_{t=1}^T\mathbb{E}\big[\mathcal{M}^2_t\big] & \leq \frac{1}{T} \sum_{t=1}^T \mathbb{E}\big[ \frac{6L^2_f}{\rho^2\mu}\big(F(\bar{x}_t)-f(\bar{x}_t,\bar{y}_t)\big) + \frac{3}{\rho^2}\|\bar{w}_t-\nabla_x f(\bar{x}_t,\bar{x}_t)\|^2 + 3\|A^{-1}_t\bar{w}_t\|^2 \big] \nonumber \\
 & \leq \frac{3G}{K^{1/3}T}(m+T)^{1/3}.
\end{align}
Let $F(\bar{x}_t) = f(\bar{x}_t,y^*(\bar{x}_t)) = \max_{y} f(\bar{x}_t,y)$. According to the Lemma~\ref{lem:A1}, i.e., $\nabla F(\bar{x}_t)=\nabla_x f(\bar{x}_t,y^*(\bar{x}_t))$, we have
\begin{align}
\| \nabla F(\bar{x}_t) - \bar{w}_t\| & = \|\nabla_x f(\bar{x}_t,y^*(\bar{x}_t)) - w_t\| = \|\nabla_x f(\bar{x}_t,y^*(\bar{x}_t)) - \nabla_x f(\bar{x}_t,y_t)+ \nabla_x f(\bar{x}_t,y_t)- w_t\| \nonumber \\
& \leq \|\nabla_x f(\bar{x}_t,y^*(\bar{x}_t)) - \nabla_x f(\bar{x}_t,y_t)\| + \|\nabla_x f(\bar{x}_t,y_t)- w_t\| \nonumber \\
& \leq L_f \|y^*(\bar{x}_t) - y_t\| + \|\nabla_x f(\bar{x}_t,y_t)- w_t\|.
\end{align}
Meanwhile, according to the Lemma~\ref{lem:A2}, we have
$$F(\bar{x}_t)-f(\bar{x}_t,y_t)=f(\bar{x}_t,y^*(x))-f(\bar{x}_t,y_t)=\max_yf(\bar{x}_t,y)-f(\bar{x}_t,y_t) \geq \frac{\mu}{2}\|y^*(\bar{x}_t)-y_t\|^2,$$
then we can obtain
\begin{align}
 \frac{\sqrt{2}}{\sqrt{\mu}}\sqrt{F(\bar{x}_t)-f(\bar{x}_t,y_t)} \geq \|y^*(\bar{x}_t)-y_t\|.
\end{align}

Thus we have
\begin{align}  \label{eq:W18}
\mathcal{M}_t & = \|A_t^{-1}\bar{w}_t\|
+ \frac{1}{\rho}\big( \frac{\sqrt{2}L_f}{\sqrt{\mu}} \sqrt{F(\bar{x}_t)-f(\bar{x}_t,y_t)}  + \|\nabla_x f(\bar{x}_t,y_t)- \bar{w}_t\|\big) \nonumber \\
& \geq \|A_t^{-1}\bar{w}_t\|
+ \frac{1}{\rho}\big( L_f \|y^*(\bar{x}_t) - y_t\|  + \|\nabla_x f(\bar{x}_t,y_t)- \bar{w}_t\|\big) \nonumber \\
& \geq  \|A_t^{-1}\bar{w}_t\|
+ \frac{1}{\rho}\|\nabla  F(\bar{x}_t)-\bar{w}_t\| \nonumber \\
& = \frac{1}{\|A_t\|}\|A_t\|\|A_t^{-1}\bar{w}_t\|
+ \frac{1}{\rho}\|\nabla  F(\bar{x}_t)-\bar{w}_t\| \nonumber \\
& \geq \frac{1}{\|A_t\|}\|\bar{w}_t\|
+ \frac{1}{\rho}\|\nabla  F(\bar{x}_t)-\bar{w}_t\| \nonumber \\
& \mathop{\geq}^{(i)} \frac{1}{\|A_t\|}\|\bar{w}_t\|
+ \frac{1}{\|A_t\|}\|\nabla  F(\bar{x}_t)-\bar{w}_t\| \nonumber \\
& \geq \frac{1}{\|A_t\|}\|\nabla  F(\bar{x}_t)\|,
\end{align}
where the above inequality $(i)$ holds by $\|A_t\| \geq \rho$ for all $t\geq1$ due to Assumption \ref{ass:6}.
Then we have
 \begin{align}
  \|\nabla F(\bar{x}_t)\| \leq \mathcal{M}_t \|A_t\|.
 \end{align}
According to Cauchy-Schwarz inequality, we have
 \begin{align} \label{eq:W19}
  \frac{1}{T}\sum_{t=1}^T\mathbb{E}\|\nabla F(\bar{x}_t)\| \leq \frac{1}{T}\sum_{t=1}^T\mathbb{E}\big[\mathcal{M}_t \|A_t\|\big] \leq \sqrt{\frac{1}{T}\sum_{t=1}^T\mathbb{E}[\mathcal{M}_t^2]} \sqrt{\frac{1}{T}\sum_{t=1}^T\mathbb{E}\|A_t\|^2}.
 \end{align}

By plugging the above inequalities \eqref{eq:W17} into \eqref{eq:W19}, we can obtain
\begin{align}
 \frac{1}{T}\sum_{t=1}^T\mathbb{E}\|\nabla F(\bar{x}_t)\| \leq \sqrt{\frac{1}{T}\sum_{t=1}^T\mathbb{E}\|A_t\|^2}\frac{\sqrt{3G}}{K^{1/6}T^{1/2}}(m+T)^{1/6}\leq \sqrt{\frac{1}{T}\sum_{t=1}^T\mathbb{E}\|A_t\|^2}\Big( \frac{\sqrt{3G}m^{1/6}}{K^{1/6}T^{1/2}} +
 \frac{\sqrt{3G}}{K^{1/6}T^{1/3}}\Big).
\end{align}

\end{proof}

\end{document}